\documentclass[10pt,journal,compsoc]{IEEEtran}

\usepackage[utf8]{inputenc} 
\usepackage[T1]{fontenc}    
\usepackage{hyperref}       
\usepackage{url}            
\usepackage{booktabs}       
\usepackage{amsfonts}       
\usepackage{nicefrac}       
\usepackage{microtype}      
\usepackage{microtype}
\usepackage{graphicx}
\usepackage{subfig}
\usepackage{booktabs} 
\usepackage{multirow}
\usepackage{times}
\usepackage{epsfig}
\usepackage{graphicx}
\usepackage{amsmath}
\usepackage{amssymb}

\usepackage[nobiblatex,linkcolor = blue]{xurl}
\hypersetup{
    colorlinks=true,
    linkcolor=blue,
    filecolor=magenta,      
    urlcolor=cyan,
    pdftitle={Overleaf Example},
    pdfpagemode=FullScreen,
    }
\usepackage{wrapfig}
\usepackage{bm}

\usepackage[ruled,linesnumbered]{algorithm2e}
 \usepackage{bbm}
\usepackage{float}
\usepackage[utf8]{inputenc}
\usepackage[english]{babel}
 
\usepackage{amsthm}
\usepackage{amsmath}
\DeclareMathOperator*{\argmax}{arg\,max}
\DeclareMathOperator*{\argmin}{arg\,min}
\usepackage[utf8]{inputenc}
\usepackage[english]{babel}
 \usepackage{MnSymbol}
\newtheorem{theorem}{Theorem}
\newtheorem{corollary}{Corollary}
\newtheorem{proposition}{Proposition}
\newtheorem{example}{Example}
\newtheorem{assumption}{Assumption}

\newtheorem{lemma}{Lemma}
\newtheorem{definition}{Definition}
\newtheorem{remark}{Remark}
\newcommand{\ie}{\emph{i.e.}}
\newcommand{\eg}{\emph{e.g.}}
\usepackage{hyperref}
\hypersetup{hidelinks}
\usepackage{color} 
 
  \def\cao{\textcolor{black}}
\usepackage{ragged2e}

\begin{document}
\title{Data-Efficient Learning via Minimizing Hyperspherical Energy}

%

\author{Xiaofeng Cao, Weiyang Liu, and 
       Ivor W.  Tsang, \IEEEmembership{Fellow IEEE} 
\IEEEcompsocitemizethanks{\IEEEcompsocthanksitem \emph{X. Cao is with  the School of Artificial Intelligence, Jilin University, Changchun, 130012, China, and the  Australian Artificial Intelligence Institute,  University of Technology Sydney, NSW 2008, Australia.  E-mail: xiaofeng.cao.uts@gmail.com. }}\protect 

\IEEEcompsocitemizethanks{\IEEEcompsocthanksitem \emph{W. Liu is with the Department of Engineering, University
of Cambridge, United Kingdom, and the Max Planck Institute for Intelligent Systems, T\"ubingen, Germany.  E-mail: wl396@cam.ac.uk. }}\protect

\IEEEcompsocitemizethanks{\IEEEcompsocthanksitem \emph{I. W. Tsang is with the  
Australian Artificial Intelligence Institute,  University of Technology Sydney, NSW 2008, Australia. 
E-mail: ivor.tsang@uts.edu.au.     }}\protect


\thanks{Manuscript received xx xx, xxxx; revised xx xx, xxxx. }}

%
%

\markboth{Journal of \LaTeX\ Class Files,~Vol.~14, No.~8, August~2015}%
{Shell \MakeLowercase{\textit{et al.}}: Bare Demo of IEEEtran.cls for Computer Society Journals}
%

\IEEEtitleabstractindextext{%
\begin{abstract}\justifying
\cao{Deep learning on large-scale data is dominant nowadays. The unprecedented scale of data has been arguably one of the most important driving forces for the success of deep learning. However, there still exist scenarios where collecting data or labels could be extremely expensive, \eg, medical imaging and robotics. To fill up this gap, this paper considers the problem of data-efficient  learning from scratch using a small amount of  representative data. First, we characterize this problem by active learning on homeomorphic tubes of spherical manifolds. This naturally generates feasible hypothesis class. With homologous topological properties, we identify an important connection -- finding tube manifolds is equivalent to minimizing hyperspherical energy (MHE) in physical geometry. Inspired by this connection, we propose a MHE-based active learning (MHEAL) algorithm, and  provide comprehensive theoretical guarantees for MHEAL, covering     convergence  and generalization analysis. Finally, we demonstrate the empirical performance of MHEAL in a wide range of applications on 
data-efficient learning, including deep clustering, distribution matching, version space sampling and deep active learning.}
 \end{abstract}
}

\maketitle

\section{Introduction}
\IEEEPARstart{R}{ecent} \cao{years have witnessed the success of deep learning~\cite{lecun2015deep} in a wide range of applications in computer vision~\cite{krizhevsky2012imagenet}, natural language processing~\cite{vaswani2017attention}, and speech processing~\cite{amodei2016deep}. The fuel that drives the progress made by deep learning is the unprecedented scale of available datasets~\cite{o2013artificial,labrinidis2012challenges}. 
Taking advantage of a huge amount of data and annotations, some powerful artificial intelligence (AI) systems are built, \eg, Deep Blue \cite{campbell2002deep}, AlphaGo \cite{wang2016does}, etc. Despite their superior performance, training and annotating large-scale data are quite expensive. In contrast, humans can easily learn to distinguish a dog and a wolf by glancing one picture. This motivates us to study the scenario where only a small amount of data is available for    AI systems.}

\cao{There are several related techniques that study how to learn   efficiently from a small amount of data or annotations, such as few-shot learning~\cite{wang2020generalizing}, active learning~\cite{gal2017deep}, and unsupervised representation learning~\cite{xie2016unsupervised}. Besides, data augmentation~\cite{zhong2020random} and generative models \cite{liu2019generative} can also be used to help data-efficient learning. This leads to a broad interest for our study. To narrow our scope, we focus on  the generalization performance  of a small amount of efficient data distributed over the version space.  Given a finite hypothesis class, data distributed around the decision boundary of version space usually generate effective hypotheses, and also produce significant updates for the model. We thus adopt a novel perspective for data-efficient learning  -- characterizing data representation via the topology of decision boundary. }

 \begin{figure} 
 \centering
\includegraphics[scale=0.235]{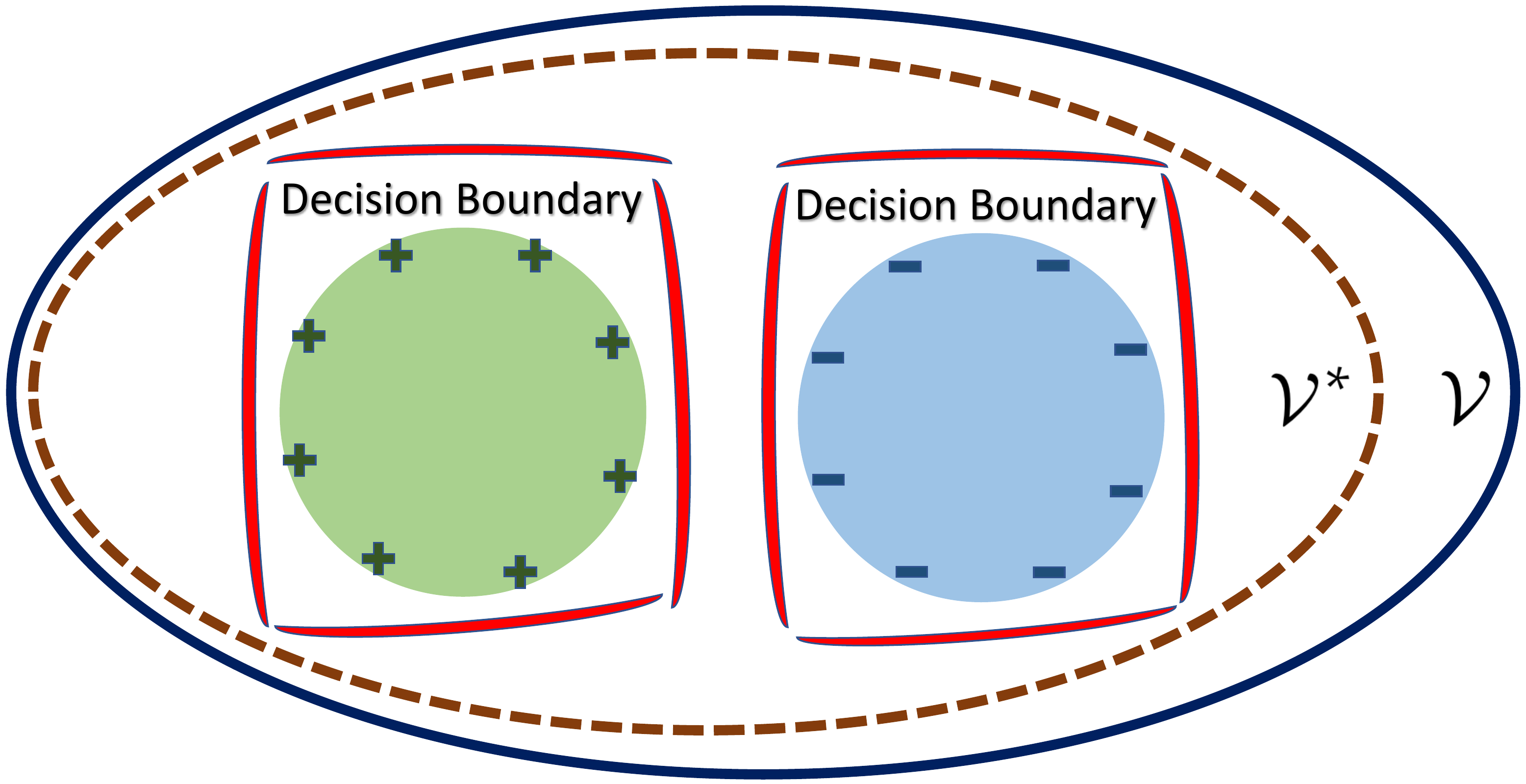}
\caption{\cao{Active learning shrinks the version space $\mathcal{V}$ into its optimal expression $\mathcal{V}^*$ which tightly covers the decision boundaries of the two spherical classes, deriving effective hypotheses. There are a small amount of data distributed around the decision boundaries, which characterize their topological structures.}}
\label{figure_boundary}
\end{figure}

\cao{Active learning  (AL), which initializes with any hypothesis (even a null hypothesis), 
  is a promising tool to learn the topology of decision boundary via iteratively shrinking the version space (see Figure~\ref{figure_boundary}).}
 Specifically,  the highly informative data which derive  the hypothesis updates are distributed in version space regions (called in-version-space), which maintain    homeomorphic manifolds \cite{li2020finding} over the topological properties of  decision boundaries. Hereafter, an AL algorithm that uses in-version-space sampling is termed version space-based AL~(VSAL)~\cite{beygelzimer2010agnostic}. 
Usually, a typical sampling criterion of VSAL  is to select those   data   which disagree with the current   hypothesis  the most, where   error disagreement  \cite{hanneke2007bound} is one effective indicator to control the sample selection.

To identify a small amount of efficient data,  typical VSAL algorithms estimate the error disagreements   by repeatedly visiting the topology of  decision boundaries. This strategy leads to  one major drawback of computational intractability \cite{beygelzimer2011efficient}. Moreover, computational estimations/approximations on generalization error     may mislead the data selection, resulting  in   incorrect or biased sampling. The reason behind lies in two folds:  1) poor initialization of model parameters, and  2)  incremental estimations/approximations. This motivates us to  learn from a small amount of data without error estimation.   Instead, we propose to utilize the topological properties of the version space for data-efficient learning. 

\begin{figure} 
 \centering
\includegraphics[scale=0.47]{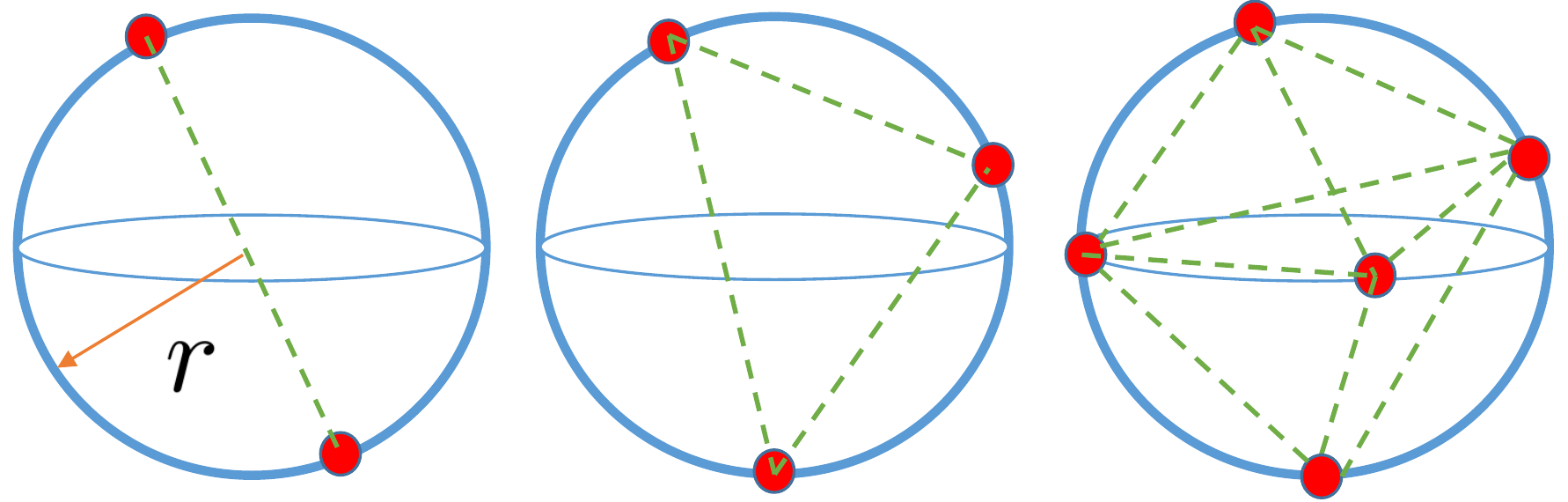}
\caption{ MHE on a sphere with a radius $r$. Left to right: $N=\{2,3,5\}$, where $N$ denotes the number of the electrons. Red points denote electrons. Dash lines denote the level of potential energy. MHE performs small yet effective data sampling from decision boundaries of version space. }  
\label{Example_of_MHE}
\end{figure}
    \begin{figure*} 
 \centering
\includegraphics[scale=0.3]{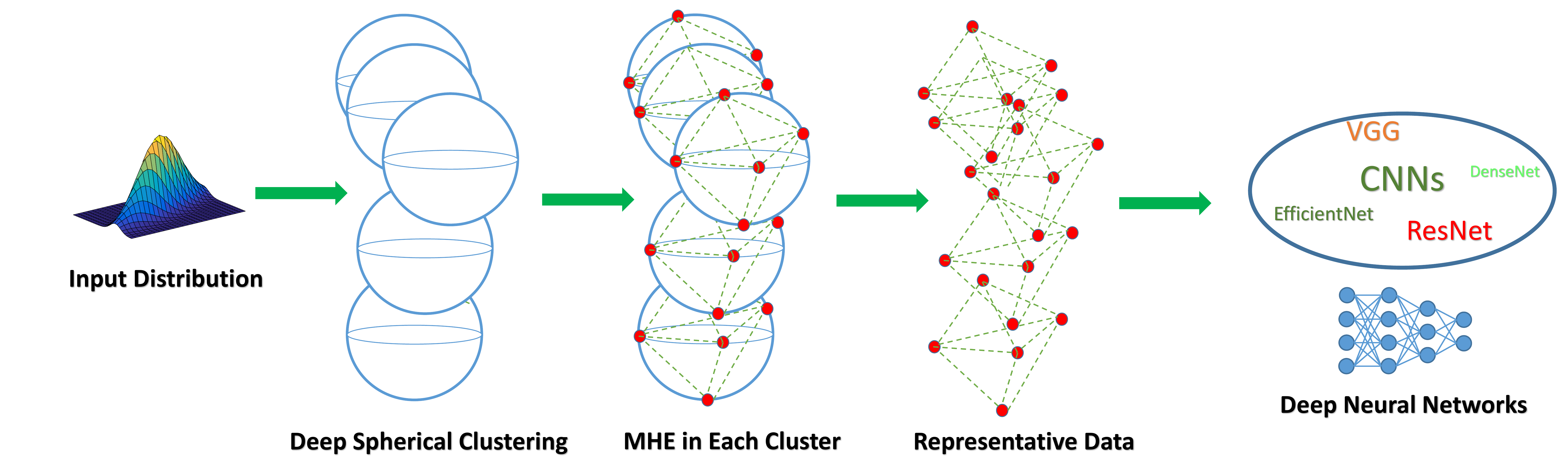}
\caption{Pipeline of MHEAL.   Input distribution is clustered into a set of spheres which perform  MHE. MHE then characterizes decision boundaries of each cluster by a small amount of representative data points (labeled in red).  }  
\label{Pipeline}
\end{figure*}
\par  In physical geometry,   there is a parallel  problem, namely 
minimizing hyperspherical energy (MHE) \cite{liu2018learning} for $N$ electrons (see Figure~\ref{Example_of_MHE}), which preserves   homologous topological properties over a spherical version space. The standard MHE includes its $\ell_0$, $\ell_1$, and $\ell_2$ expressions \cao{using a natural generalization of the 0,1,2-norm for finite-dimensional vector spaces, respectively.}
Geometrically, MHE is equivalent  to characterizing  decision boundaries  of a version space via   homeomorphic tube manifold \cite{ben2008relating},  which  naturally generates feasible hypothesis class. This paper aims at learning with a small amount of data from decision boundaries via the proposed MHE-based AL (MHEAL) algorithm. To characterize these representative data, we maximize the lower bound of the optimal $\ell_0$ expression of MHE as a feasible and efficient approximation over  each pre-estimated   spherical cluster  (see Figure~\ref{Pipeline}). Our theoretical insights show that this approximation enjoys the properties of geometric preservation and controllable approximation accuracy.  In the importance weighting setting of AL, we further prove that MHEAL yields a faster hypothesis-pruning speed, which then leads to tighter bounds on its generalization error and label complexity\footnote{Label scale of achieving a desired error threshold.}. A high-level summarization of this paper is given as follows:
\begin{itemize}
  
\item \textbf{Goal:}  Learning efficient-representative data from homeomorphic tubes of spherical manifolds via AL.

\item \textbf{Motivation:} Exploring  data representation from decision boundary by typical VSAL has computational intractability. To address this, we characterize the topological properties of version space by MHE.
 
\item \textbf{Solution:}  {We propose the MHEAL algorithm with theoretical guarantees on generalization error and label complexity.}


\end{itemize}

{\color{black}Our major contributions are summarized as follows:

\begin{itemize}
\item \cao{ We study data-efficient learning as an  important step toward training AI systems with a small amount of  data. In a nutshell, we propose to characterize the decision boundaries of version space by minimizing the hyperspherical energy of a small amount of representative data.}

\item  	We derive a lower bound maximization scheme to approximate the greedy sequential solution of $\ell_0$ expression of MHE, yielding a much lower computation complexity.

\item 	We give theoretical analyses for our approximation optimization scheme which guarantees both geometric preservation and approximation accuracy.

\item We propose a novel active learning algorithm called MHEAL, which performs MHE over each pre-estimated spherical cluster  in an unsupervised manner and effectively characterizes their decision boundaries.

\item  We prove the theoretical convergence of MHEAL by deriving its generalization error and label complexity bounds. Moreover, we conduct an extensive empirical study on the generalization performance to demonstrate the effectiveness of our  bounds.

\end{itemize}}


\section{Related Work}
VSAL presents a theoretical interpretation for learning from representative data via AL, and MHE implements an advanced VSAL algorithm from physical geometry. We briefly introduce related work from these two aspects. 

\textbf{VSAL}   The VSAL algorithm \cite{beygelzimer2010agnostic,beygelzimer2011efficient} prunes the hypothesis class by    annotating  the unlabeled data via the hypothesis disagreement maximization, which requires the optimal hypothesis always to be included. Specifically, the hypothesis-pruning~\cite{cao2020shattering} always needs to maintain a correct update, where error disagreement \cite{hanneke2014theory} is an important indicator to control these updates. With the estimation on errors, the error disagreement indicator can be specified as the best-in-class error \cite{cortes2020adaptive}, average-in-class  error \cite{cortes2019active}, entropy of error \cite{roy2001toward}, etc.  A series of confidence label complexity bounds were derived under noise-free settings, such as the agnostic PAC bound \cite{balcan2009agnostic,zhang2014beyond,huang2015efficient,wang2009sufficient}. With a desired error threshold, the  annotating budget is also bounded. By employing importance weighting, importance weighted AL (IWAL) \cite{beygelzimer2009importance}   utilizes  the on-line sampling theory to tighten the label complexity bounds,  before converging into the optimal hypothesis. To further improve it,  \cite{cortes2019active} gives more refined analysis on the current hypothesis class to reduce the best-in-class error, resulting in a tighter bound on label complexity \cite{hanneke2007bound}. Guarantees under different noise settings are also studied, \eg, adversarial noise \cite{awasthi2014power}, malicious noise \cite{kearns1993learning}, random classification
noise \cite{angluin1988learning}, and bounded noise \cite{massart2006risk}.

Few-shot learning \cite{wang2020generalizing} also studies how to adapt to a new task with a very small amount of data, but it typically requires extra information about the base tasks. Moreover, few-shot learning mostly considers the passive setting where the label distribution is explicitly controlled by one specific sampling scenario in the pre-defined training set. In contrast, AL that learns from scratch using  representative data is not as limited as few-shot learning, since its algorithms can stop their iterative sampling either when they achieve desired accuracy or when the annotation budgets are exhausted.

\textbf{MHE}   Thomson problem \cite{bowick2002crystalline} in physics describes the ground state energy and configuration of  a set of  electrons on a unit hypersphere. Given $N$ electrons, Liu~\emph{et al.}~\cite{liu2018learning} seek to achieve their minimum  potential
energy by considering their interactions on a unit hypersphere. This also can be viewed as maximizing hyperspherical uniformity~\cite{liu2021learning}. Recent studies~\cite{liu2017sphereface,liu2017deep,liu2018decoupled,chen2020angular} have shown that hyperspherical similarity preserves the
most abundant and discriminative information.
Modeling with DNNs,  geodesic distances \cite{lou2020differentiating} between neurons  
that  discriminate the features,  are projected and optimized on the hypersphere. However, naively minimizing hyperspherical energy from geometry suffers from
some difficulties  due to the underlying non-linearity and non-convexity. To alleviate these difficulties, Lin \emph{et al.} \cite{lin2020regularizing} propose the
compressive MHE (CoMHE) as a more effective regularization to minimize hyperspherical energy for neural networks. Following \cite{liu2018learning,lin2020regularizing}, Perez-Lapillo~\emph{et al.}~\cite{perez2020improving}  and  Shah~\emph{et al.} \cite{shah2020impact} improve voice separation by applying MHE to Wave-U-Net and  time-frequency domain networks, respectively. MHE has wide applications in image recognition~\cite{chen2020angular,li2020oslnet,liu2021orthogonal}, face recognition~\cite{liu2017sphereface,liu2018learning,deng2019arcface}, speaker verification~\cite{liu2019large}, adversarial robustness~\cite{pang2019rethinking}, few-shot learning~\cite{mettes2019hyperspherical,liu2019neural}, etc.

\begin{figure}
 \centering
\includegraphics[scale=0.39]{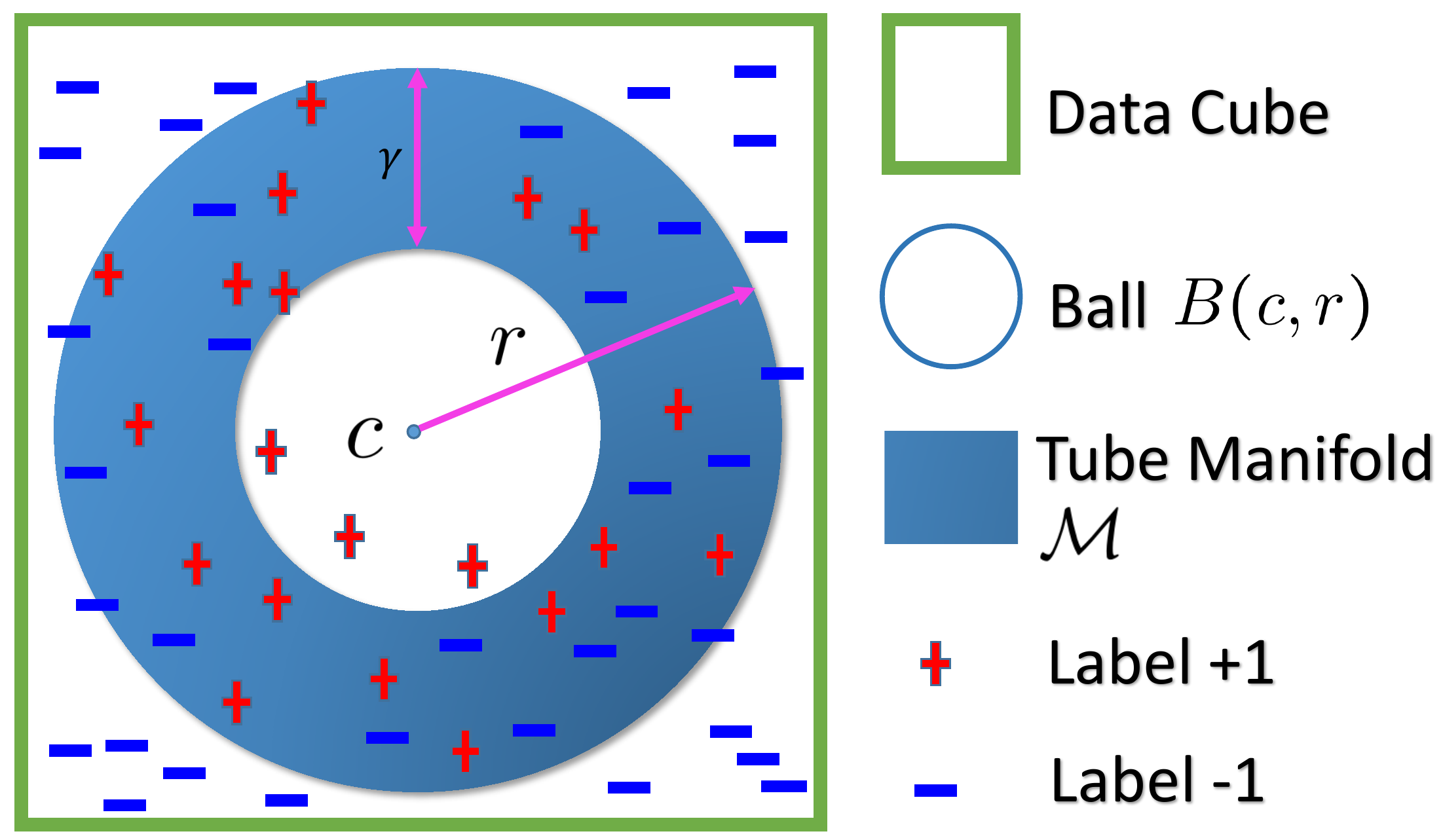}
\caption{Characterization of a 2D decision boundary by a tube manifold $\mathcal{M}$ with a width $\gamma$. }
\label{tube}
\end{figure}

\section{MHE and Decision Boundary}
Section~\ref{dec_boun} characterizes the  homologous topological properties of decision boundaries for tube manifolds. Section~\ref{intro_mhe} presents the MHE problem of physical geometry.  Section~\ref{seqopt_mhe} then proposes a sequential  solution for   $\ell_0$ expression of MHE. To find an alternative scheme for such a greedy solution, Section~\ref{low_bou_max_app} maximizes the lower bound of the optimal $\ell_0$ expression of MHE as a feasible approximation.

\subsection{Decision Boundary}\label{dec_boun}
We start by characterizing the decision boundaries.  Given a feature space $\mathcal{X}$ associating with a label space $\mathcal{Y}$, let $\mathcal{Y}=\{1,2,...,K\}$, we define $P_{\mathcal{X}\mathcal{Y}}$ as the joint distribution over the mapping from $\mathcal{X}$ to $\mathcal{Y}$. According to the Bayesian theorem, decision boundary of a class (cluster) can be constructed on  a tube  manifold $\mathcal{M}$ \cite{li2020finding}. We next give a conceptual description for decision boundaries in the multi-class setting.

\begin{definition}[Decision boundary]
Given a class i.i.d. drawn from  $P_{\mathcal{X}\mathcal{Y}}$ with label $k$ embedded in a label space $\mathcal{Y} \in \{1,2,3,...,K\}$, $k'\in \mathcal{Y}$ and $k'\neq k$,  the decision boundary of this class is distributed over a tube manifold $\mathcal{M}$ with uncertain label predictions, which satisfy $\mathcal{M}:=\{x\in \mathcal{X}| P_{\mathcal{Y}|\mathcal{X}}(k'|x)=P_{\mathcal{Y}|\mathcal{X}}(k|x)\}$, where the optimal Bayesian classifier is $f(x)=k$  if $P_{\mathcal{Y}|\mathcal{X}}(k|x) \geq 0.5$, otherwise $f(x)=k' \in \{1,2,...,k-1,k+1,...,K\}$.
\end{definition}

{\color{black} \begin{remark}
Definition~1 characterizes the decision boundary of a tube manifold over a  label space $\mathcal{Y}$, where  those samples  around the tube are with uncertain class labels.  For any high-dimensional sphere, the tube manifold  still covers those samples  with uncertain labels distributed near the boundary of the  class. 
\end{remark}

If $\mathcal{Y}$ is  a binary label space, that is,  $\mathcal{Y} \in \{+1,-1\}$, we then adapt Definition~1 into the following specific case.

\begin{example}[Binary case of decision boundary]
Given a class embedded in a label space $\mathcal{Y}=\{+1,-1\}$, its decision boundary is distributed over a tube manifold $\mathcal{M}$ with uncertain label predictions, which satisfy $\mathcal{M}:=\{x\in \mathcal{X}| P_{\mathcal{Y}|\mathcal{X}}(+1|x)=P_{\mathcal{Y}|\mathcal{X}}(-1|x)\}$, where the optimal Bayesian classifier is $f(x)=+1$  if $P_{\mathcal{Y}|\mathcal{X}}(+1|x) \geq 0.5$, otherwise $f(x)=-1$.
\end{example}

From a geometric perspective   \cite{ben2008relating}, the manifold of decision boundary can be characterized by a tube with a width $\gamma$, which keeps  homologous  properties with a ball $B(c,r)$, where $c$ denotes the center and $r$ denotes the  radius (see Figure~\ref{tube}). \cao{We give the following assumption for the homeomorphic tube manifold, where $B(c,r-\gamma)$ denotes the concentric ball of $B(c,r)$. Specifically, $r-\gamma$ is the radius, $c$ is the center, and  $\gamma$ is a variable to denote the width of the tube manifold. }

\begin{assumption}
Assume  that the label complexity of a version space-based hypothesis is characterized by a set $S$ over the tube manifold $\mathcal{M}:=\{B(c,r)\backslash B(c,r-\gamma)\}$, if $S \subseteq D^{1}=\{x\in \mathcal{X}: f(x)=+1\}$, we have: 1)  $\bigcap_{x_i\in S}  \mathcal{M}(x_i, \gamma)\neq \emptyset$, where  $\mathcal{M}(x_i, \gamma)$ denotes the tube manifold covering $x_i$ with a width $\gamma$;
2) let $D^{-1}=\{x\in \mathcal{X}: f(x)=-1\}$,  if 
$x_i\subseteq D^{+1}$ and $x_j\subseteq D^{-1}$, there exists $\|x_i-x_j\|\leq \gamma$, where  $\|\cdot\|$ denotes the $\ell_2$-norm. 
\end{assumption}}
\cao{In Assumption~1, the first condition requires that there is a homeomorphic tube of $\mathcal{M}$ with a width $\gamma$  which covers $x_i\in S$ to contain shared elements from both classes. This means that $\mathcal{M}$ is a tight covering for $S$. The second condition describes the width of  $\mathcal{M}$, \ie, $\gamma$.
We have defined decision boundaries to characterize those feasible hypothesis class as a consistent concept, and then generalize it with MHE.}

\subsection{MHE}\label{intro_mhe}
MHE \cite{liu2018learning} seeks to find the configuration of $N$ mutually-repelling electrons that minimizes the potential energy. These electrons are distributed on a unit hypersphere. In physics, such a potential energy is used to describe a balanced state of the  distribution of electrons. More generally, MHE corresponds to the uniform distribution on the hypersphere~\cite{liu2021learning}.
 {\color{black}From a geometric perspective, it can be used to characterize the decision boundary distributed in a homeomorphic tube manifold  of a hypersphere.}

Given $N$ samples, \ie, $\mathcal{W}=\{w_1,w_2,...,w_N \in \mathbb{R}^{d} \}$, let $\hat w_i$ denote  the $\ell_2$-norm  projection of $w_i$ on the unit hypersphere. Then their hyperspherical energy can be defined as 
\begin{equation}
\begin{split}
\mathbb{E}_{s,d}(\hat w_i|_{i=1}^N):&=\sum_{i=1}^N \sum_{j=1}^N f_s(\|\hat w_i-\hat w_j  \|)\\
&=
\left\{
        \begin{array}{lr}
            \sum_{i> j}\|\hat w_i-\hat w_j\|^{-s},  & s>0  \\
            \sum_{i> j} {\rm log} (\|\hat w_i-\hat w_j\|^{-1}), & s=0\\
             \end{array}
\right. ,
\end{split}
\end{equation}
where  $f_s$ denotes an energy function, and $i, j\geq 1$. Specifically,  $\mathbb{E}_{0,d}$ denotes the logarithmic potential energy, that is, the $\ell_0$ expression of MHE (also termed as $\ell_0$ MHE), \ie, $\mathbb{E}_{0,d}= \sum_{i> j} {\rm log} (\|\hat w_i-\hat w_j\|^{-1})$. Then, minimizing the $\ell_0$ hyperspherical energy can be viewed as an approximation to minimizing $\mathbb{E}_{s,d}$~\cite{liu2018learning}. $\ell_0$ MHE is formally defined as
\begin{equation}
\begin{split}
 \argmin_{w_1,w_2,...,w_N}  \mathbb{E}_{0,d}&= \sum_{i> j} {\rm log} (\|\hat w_i-\hat w_j\|^{-1})\\
& = \argmax_{w_1,w_2,...,w_N} \prod_{i> j} \|\hat w_i-\hat w_j\|.
\end{split}
\label{ell_0_MHE}
\end{equation}
Eq.~(\ref{ell_0_MHE}) is essentially a limiting case of the MHE objective $\mathbb{E}_{s,d}$~\cite{liu2018learning}. Besides this, $\mathbb{E}_{1,d}$  and $\mathbb{E}_{2,d}$ are also   feasible alternatives, namely   $\ell_1$  and   $\ell_2$ expressions of MHE, respectively (also term $\ell_1$ MHE and $\ell_2$ MHE). However,   minimizing $\mathbb{E}_{1,d}$ is a typical NP hard problem, which usually requires a sequential search. $\mathbb{E}_{2,d}$ can be solved by gradient descent with the following gradient \cite{lin2020regularizing}:
\begin{equation}
\begin{split}
  \nabla_{\hat w_i}  \mathbb{E}_{2,d}=\sum_{j=1,j\neq i}\frac{-2(\hat w_i-\hat w_j)}{\|\hat w_i-\hat w_j  \|^4},  
\end{split}
\end{equation}
whose solution yields $\hat w_i=\frac{\sum_{j=1}^N  \|\hat w_i-\hat w_j  \|^{-4}  \hat w_j}{\sum_{j=1}^N \|\hat w_i-\hat w_j  \|^{-4}}$ s.t. $j\neq i$.

 {\color{black}
\subsection{Sequential Optimization of MHE}\label{seqopt_mhe}

We here present the optimal solution of $\ell_0$ MHE by employing a greedy strategy. Given  $\mathcal{P}$ containing 
$M$ feasible data points in $\mathbb{R}^d$, MHE is performed to find $N$ data points for $\mathcal{W}$. Let 
$\hat w_1 \in  {P}$ and $\mathcal{W}$ be initialized by $\mathcal{W}=\{\hat  w_1\}$. \cao{At the $t$-th step,  a sequential optimization  is adopted to find $\hat w_t$ from $\mathcal{P}$
\begin{equation}
\begin{split}
 \hat w_t &=\argmax_{ {\hat  p} \in  \mathcal{P}}     \prod_{u,v\in\{{\hat w_{1}},\cdots,   {\hat w_{t-1}}, {\hat  p} \}} 
 \|u-v\|\\
 &=\argmax_{ {\hat  p} \in  \mathcal{P}} \prod_{t-1\geq i\geq 1}\|\hat w_i - \hat p\|, 
 \end{split}
 \label{sequential_MHE}
\end{equation}
after which $\mathcal{W}$ is updated by adding a new data point, \ie, $\mathcal{W}\leftarrow\mathcal{W}\cup\{\hat w_t\}$ at the $t$-th step. In this setting, computing one sample costs
$\mathcal{O}(MN)$ complexity, where the argmax operation costs $\mathcal{O}(M)$, and the computation of $\prod_{t-1\geq i> 1}$ costs   $\mathcal{O}(N)$. To obtain $N$ feasible samples,  Eq.~(\ref{sequential_MHE})
is computed until $t$ arrives at $N$, thereby costing  $\mathcal{O}(MN^2)$.}


Let $\mathcal{L}_{\mathbb{E}_{0,d}}( {\hat w}_1)$ denote the $\ell_0$ hyperspherical energy of $N$ data points with an initial  sample $ {\hat w}_1$, the optimal initialization on $ {\hat w}_1$, \ie, $ {\hat w}_1^*$, is obtained by  
\begin{equation}
\begin{split}
{\hat w}_1^*= \argmax_{\hat p\in \mathcal{P}}     \mathcal{L}_{\mathbb{E}_{0,d}}(\hat p).
 \end{split}
 \label{optimal_w_1}
\end{equation}

For $M$ times  of the sequential optimization in Eq.~(\ref{optimal_w_1}), it will cost  $\mathcal{O}(M^2N^2)$ to obtain the optimal $ {\hat w}_1$.
A detailed process is then presented in Algorithm~1.

\begin{algorithm}[t]
 { \caption{Sequential optimization of $\ell_0$ MHE}
   \textbf{Input:} $ \mathcal{P}=\{\hat p_1,\hat p_2,...,\hat p_M \}$ containing 
$M$ feasible data points in $\mathbb{R}^d$. \\
   \textbf{Initialize:}  $l=0$.\\
   \While{$l<M$}{
       $\hat w_1=\hat p_l$,  $\mathcal{W}= \{{\hat w}_1\}$.\\
   \For{$t=2,...,N$}{
    $\hat w_t= \argmax_{ {\hat  p} \in  \mathcal{P}} \prod_{t-1\geq i\geq 1}\|\hat w_i - \hat p\|$\\
    $\mathcal{W}\leftarrow\mathcal{W}\cup\{\hat w_t\}$
   }
   Obtain the energy: $ \mathcal{L}_{\mathbb{E}_{0,d}}(\hat p_l)=  \prod_{\mathcal{W}, i> j} \|\hat w_i-\hat w_j\|.$\\
   }
 \textbf{Output:} The optimal   ${\hat w}_1^*$ from   $\argmax_{\hat p_l\in \mathcal{P}}  \mathcal{L}_{\mathbb{E}_{0,d}}(\hat p_l),$ and its optimal energy expression $\mathcal{L}_{\mathbb{E}_{0,d}}({\hat w}_1^*)$.}
\end{algorithm}

\subsection{Approximation by Lower Bound Maximization}\label{low_bou_max_app}
The sequential optimization comes with a computation complexity of $\mathcal{O}(M^2N^2)$, which is not efficient to solve. 
We thus aim to find a feasible approximation.

\begin{proposition}
Assume that $\hat w_1,\hat w_2,..., \hat w_{N-1}$ are known  samples for $\mathcal{W}$, the following step is to optimize $\hat w_N$, that is,   $\hat w_N=\argmax_{\hat p\in \mathcal{P}} \mathcal{L}_{\mathbb{E}_{0,d}}(\hat p), $ $s.t. \  \mathcal{W}=\{\hat w_1, \hat w_2,..., \hat w_{N-1} \}$, where this argmax operation invokes Eq.~(\ref{sequential_MHE}) one time since $\mathcal{W}$ has already included $N-1$ samples.  With such setting,  assume that  $\min_{\hat w_j \in \mathcal{W}} \|\hat w_N-\hat w_j\|\leq  \min_{\hat w_i, \hat w_j \in \mathcal{W}}\|\hat w_i-\hat w_j\|$,   recalling the energy definition of Eq.~(\ref{ell_0_MHE}),  there exists an inequality 
\begin{equation*}  
\begin{split}
&\Big (\min_{\hat w_j \in \mathcal{W}} \|\hat w_N-\hat w_j\|\Big)^{{\frac{N^2-N}{2}} } \leq  \\
 &\Big(\!\prod_{ N-1>i>j }\!  \|\hat w_i-\hat w_j\|\Big) \!\times\!  \Big (\!(N-1)\!\min_{\hat w_j \in \mathcal{W}} \|\hat w_N-\hat w_j\|      \Big)    \leq   \mathcal{L}_{ \mathbb{E}_{0,d}}( {\hat w}_N),\\
\end{split}
\end{equation*}
where     $\min_{\hat w_j \in \mathcal{W}} \|\hat w_N-\hat w_j\| $ returns the minimal geodesic distance of $\hat w_N$ to the data point of $\mathcal{W}$. 
\end{proposition}

 To tightly approximate the optimal energy of $\mathcal{L}_{ \mathbb{E}_{0,d}}( {\hat w}_N)$, an effective way is to maximize its lower bound $\big (\min_{\hat w_j \in \mathcal{W}} \|\hat w_N-\hat w_j\|      \big)^{\frac{N^2-N}{2}}$. We thus present Corollary~1 to state our alternative scheme.

\begin{corollary}[Lower bound maximization] To obtain an approximated  optimal $\hat w_t^*$, one feasible method is to maximize the lower bound of $\mathcal{L}_{ \mathbb{E}_{0,d}}({\hat w}_t)$, that is, 
\begin{equation}
  \max_{\hat p\in \mathcal{P}}  \big (\min_{  \hat w_j \in \mathcal{W}   } \| {\hat  p}  -\hat w_j\|      \big)^{\frac{t^2-t}{2}}.
\end{equation}
Equivalently, this max-min solution is simplified as 
\begin{equation}
\max_{\hat p\in \mathcal{P}}   \min_{  \hat w_j \in \mathcal{W}   } \| {\hat  p}  -\hat w_j\|.
\end{equation}
\end{corollary}

\noindent\textbf{Sequential max-min} With Corollary~1,  at $t$-time, acquiring $\hat w_t$  is sequentially obtained from
\begin{equation}
\begin{split}
&\argmax_{\hat p\in \mathcal{P}}   \min_{  \hat w_j \in \mathcal{W}   } \| {\hat  p}  -\hat w_j\|, \  \  
\end{split}
\label{se_max_min}
\end{equation}
\cao{where $\mathcal{W}$ is updated by $ {\hat w_{t-1}}$. In this max-min optimization, }
 the argmax needs to perform $M$ times of min operation. On such setting, 
acquiring a sample $ { \hat w_t}$ at $t$-time costs  about $\mathcal{O}(MN)$.  


\noindent\textbf{Computation complexity}  To obtain $N$ feasible data from  $\mathcal{P}$, Eq.~(\ref{se_max_min}) is iteratively invoked $N$ times, which costs  about $\mathcal{O}(MN^2)$. To find the optimal $ {\hat w_1}$,  Eq.~(\ref{optimal_w_1}) needs to be performed $M$ times. Therefore, the total computation complexity of our lower bound approximation scheme costs about $\mathcal{O}(M^2N^2)$.

\section{Theoretical Insights}
In physics, MHE projects the vectors on a unit sphere to maximize their potential energies.  Ata Kaban \emph{et al.} \cite{kaban2015improved} proposed Gaussian random projection to redefine it from a vector view. Following their works, we present  theoretical insights for our proposed optimization scheme.

\subsection{Geometric Projection}
We firstly review the Gaussian random projection that maps vectors from $\mathbb{R}^d$ to $\mathbb{R}^\kappa$, where $\kappa<d$.
\begin{lemma}
Given $\hat w_1, \hat w_2,...., \hat w_N\in \mathbb{R}^d$, $P\in \mathbb{R}^{\kappa\times d}$ denotes a   Gaussian random
projection matrix   \cite{kaban2015improved}  where  $P_{ij}=\frac{1}{\sqrt{n}}r_{ij}$, $r_{ij}$ is i.i.d. drawn from $\mathcal{N}(0, \sigma^2)$, and $P\hat w_1, P\hat w_2 \in \mathbb{R}^d$ are the random projection of $\hat w_1, \hat w_2$ under $P$. Then, for any variable $\varepsilon \in [0,1]$, there exists
\begin{equation*}
\begin{split}
&{\rm Pr} \Big \{(1-\varepsilon)\|\hat w_i- \hat w_j \|^2k \sigma^2<  \|P\hat w_i- P\hat w_j \|^2\\
& < (1+\varepsilon)\|\hat w_i- \hat w_j \|^2k \sigma^2\Big\}   \geq 1-2{\rm exp} \Big   ( \frac{-\kappa\varepsilon^2}{8}  \Big),  \\
\end{split}
\end{equation*}
which satisfies $\kappa \varepsilon^2>5.54517744$.
\end{lemma}
Note that the inequality of $\kappa \varepsilon^2$ is limited by $2{\rm exp}   (\frac{-\kappa\varepsilon^2}{8} )\leq 1$.
With Lemma~1,   $\argmax_{\hat w_1,\hat w_2,...,\hat w_N} \prod_{i> j} \|\hat w_i-\hat w_j\|$ is approximated by $\argmax_{\hat w_1,\hat w_2,...,\hat w_N} \prod_{i> j} \max_{P}\|\hat Pw_i-P\hat w_j\|$. Liu \emph{et al.}  \cite{liu2018learning} used this idea to  regularize the network generalization. In our scheme, how to project $\{\hat w_1,\hat w_2,...,\hat w_N\}$ onto a sphere is a necessary process. However, most of the input distributions are aspherical \cite{cao2021distribution}, and we usually project the data into a spherical-likeness geometry. Typical data normalization methods are adopted to relieve this problem. For example, we project $\argmax_{\hat w_1,\hat w_2,...,\hat w_N}$ with a Gaussian projection to obtain a 0-mean and $\sigma^2$- variance distribution. With such idea, we present our projection proposition.

\begin{proposition}
Given $\hat w_1, \hat w_2,..., \hat w_N\in \mathbb{R}^d$, $P\in \mathbb{R}^{\kappa\times d}$ denotes a   Gaussian random
projection matrix  where  $P_{ij}=\frac{1}{\sqrt{n}}r_{ij}$, $r_{ij}$ is i.i.d. drawn from $\mathcal{N}(0, \sigma^2)$, and $P\hat w_1, P\hat w_2 \in \mathbb{R}^d$ are the random projection of $\hat w_1, \hat w_2$ under $P$. To satisfy the condition of $\kappa<d$ in Lemma~1, we extend a  vector with zero-element to the   $(d+1)$ dimension of  $\hat w_1, \hat w_2, \hat w_3, ..., \hat w_N\in \mathbb{R}^{d}$,  and obtain $\hat w_1', \hat w_2', \hat w_3', ..., \hat w_N'\in \mathbb{R}^{d+1}$, which require $\hat w_{id'}=0$ for any $1 \leq i\leq N$, where $d'=d+1$.  Considering that $P$ still is a $d$-dimensional projection matrix,   for any variable $\varepsilon \in [0,1]$, the inequality of Lemma~1 still holds.
\end{proposition}



In short, Lemma~1 still holds if the Gaussian projection limits $\kappa=d$. Then, the projection is a more specific normalization guaranteed by Lemma~1.
\subsection{Geometric Preservation}
With Proposition~2, under the limitation of $\kappa=d$, we observe the geometric preservation for our projection idea.

\cao{
\begin{lemma}[Preservation of geodesic inequality] Given $\hat w_1, \hat w_2,..., \hat w_N\in \mathbb{R}^d$, $P\in \mathbb{R}^{d\times \kappa}$ denotes a   Gaussian   random
projection matrix $P$  where  $P_{ij}=\frac{1}{\sqrt{n}}r_{ij}$, $r_{ij}$ is i.i.d. drawn from $\mathcal{N}(0,\sigma^2)$, and $P\hat w_1, P\hat w_2 \in \mathbb{R}^d$ are the random projections of $\hat w_1, \hat w_2$ under $P$, respectively. Let $d_\mathcal{M}$ be the geodesic metric over a curve $g$ of $\mathbb{R}^d$ which satisfies $d_\mathcal{M}(g(w_1), g(w_2)):=\nu\|w_1-  w_2\|$,  where $\nu\leq 0$,   $  d'_\mathcal{M}$ be the geodesic metric over a curve $g'$ of $\mathbb{R}^\kappa$ which satisfies $ d'_\mathcal{M}( g'(w_1), g(w_2)):=\nu'\|w_1-  w_2\|$,  where $\nu'\leq 0$, if $ d_\mathcal{M}(\hat w_1,\hat w_2)\leq   d_\mathcal{M}(\hat w_2,\hat w_3)$, if  $\nu=\nu'$, the inequality still hods for $  d'_\mathcal{M}$, \ie, there exists  $ d'_\mathcal{M}(P\hat w_1,P\hat w_2)\leq   d'_\mathcal{M}(P\hat w_2,P\hat w_3)$.
\end{lemma}}

\begin{lemma}[Preservation of angle inequality] Given $\hat w_1, \hat w_2,..., \hat w_N\in \mathbb{R}^d$,  $P\in \mathbb{R}^{d\times \kappa}$ denotes a   Gaussian random
projection matrix  where  $P_{ij}=\frac{1}{\sqrt{n}}r_{ij}$, $r_{ij}$ is i.i.d. drawn from $\mathcal{N}(0,\sigma^2)$, and $P\hat w_1, P\hat w_2 \in \mathbb{R}^d$ are the random projections of $\hat w_1, \hat w_2$ under $P$. Let $\theta_\mathcal{M}(\cdot,\cdot)$ be the angle metric over $\mathbb{R}^d$,      if $ \theta_\mathcal{M}(\hat w_1,\hat w_2)\leq   \theta_\mathcal{M}(\hat w_2,\hat w_3)$,   there exists $   \theta_\mathcal{M}(P\hat w_1,P\hat w_2)\leq    \theta_\mathcal{M}(P\hat w_2,P\hat w_3)$, otherwise, given   $\theta_\mathcal{M}(\hat w_1,\hat w_2)= \frac{1}{e}\frac{ (P \hat w_1)^T (P \hat w_2)}{\|  (P \hat w_1)^T  \|\cdot \|  (P \hat w_2)^T  \|},$ following  \cite{lin2020regularizing}, there then exists 

\[  \frac{ \theta_\mathcal{M}(P\hat w_1,P\hat w_2)-e\varepsilon} {e+e\varepsilon} <   \theta_\mathcal{M}(P\hat w_2,P\hat w_3),  \]
 where $\varepsilon\in[0,1]$,   $e>0$, and $\kappa=d$.
\end{lemma}
 



 

\subsection{Convergence of Approximation}
 We present the bound  of the approximation loss of Proposition~1, and then extend it into a more general bound of acquiring $N$ samples.
 \cao{
\begin{proposition}[Controlled approximation loss] 
Let $ \ell^{upper}_{LBM}({\hat w}_{1:N})$ be the upper bound of the approximation loss of acquiring $N$ samples,  given $\ell^{\sqrt{upper}}_{LBM}({\hat w}_{1:N})= \sqrt[\frac{N^2-N}{2}]{\ell^{upper}_{LBM}({\hat w}_{1:N})}$, there exists  
\begin{equation*}
\begin{split}
 & \min_{\hat p \in \mathcal{P}}  2\Big (\max_{\hat w_j \in \mathcal{W}} \|\hat p -\hat w_j\| -  \min_{\hat w_j \in \mathcal{W}} \|\hat p-\hat w_j\|   \Big) \\
&  \leq \ell^{\sqrt{upper}}_{LBM}\Big({\hat w}_{1:N}\Big) 
   \leq \max_{\hat p \in \mathcal{P}}\Big (\max_{\hat w_j \in \mathcal{W}} \|\hat p-\hat w_j\|  \Big). \\
 \end{split}
\end{equation*}
Specifically, for any approximation, there exists 
\begin{equation*}
\begin{split}
 &   \mathcal{O}\Big(\min_{\hat w_i, \hat w_j \in \mathcal{W}} \| {\hat w}_i-\hat w_j\| \Big)  \!\!   \leq \!\! \ell^{upper}_{LBM}({\hat w}_{1:N})\leq \!\!  \mathcal{O}\Big(\max_{\hat w_i, \hat w_j \in \mathcal{W}} \|{\hat w}_i-\hat w_j\|  \Big),\\
 \end{split}
\end{equation*}
that is, any upper bound of the MHE approximation  to the optimal solution  is proportional to the maximal length of the sphere chord, and any of its lower bound  is proportional to the minimal length of the  sphere chord.
\end{proposition}}

Proposition~3 presents a  natural derivation process for the approximation loss of Proposition~1. We can see that our proposed lower bound maximization can  properly converge.

\section{MHEAL: MHE-based Active Learning}
Based on the theoretical results of Sections~3 and 4,   MHEAL adopts a max-min solution of Eq.~(\ref{se_max_min})  in a set of hyperspheres as shown in   Figure~\ref{Pipeline}. This section thus interprets this process beginning from the hyperspherical clustering.
\subsection{Hyperspherical Energies}
 {\textbf{MHE with Clustering} Our goal is to characterize the decision boundaries over each cluster. Therefore, it is necessary to perform clustering  in  input distribution. Moreover,  adopting clustering to reduce $N,M$ into $N/k,M/k$, respectively, can effectively reduce the calculation complexity of sequential solution of $\ell_0$ MHE. In this way, the calculation complexity  of our optimization scheme reduces  to  $\mathcal{O}(\frac{M^2N^2}{k^4})$, which is  lower than previous $\mathcal{O}({M^2N^2})$.
}

\vspace{1.5mm}
\noindent\textbf{Hyperspherical Clustering} To characterize the hyperspherical energies, we use hyperspherical $k$-means (termed SphericalKmeans) to obtain $k$ hyperspheres. Following \cite{buchta2012spherical}, the standard   hyperspherical $k$-means is to optimize a set of sphere centers $C=\{c_1,c_2,...,c_k\}$: 
\begin{equation}
 \argmin_{c_k}   \sum_{x_i} \Big(1-{\rm cos}(c_k,x_i)\Big),
 \label{center}
\end{equation} 
where $c_k=\sum_{x_i } \mathbbm{1}_{y_i=c_k}  \frac{x_i}{\|x_i\|}$, and $\mathbbm{1}$ denotes the indication function. With Eq.~(\ref{center}), minimizing hyperspherical energies  over each hypersphere is transferred into learning 
tube energies in decision boundaries around those hyperspheres. Note that applying SphericalKmeans needs to normalize the input features with Gaussian projection \footnote{https://scikit-learn.org/stable/modules/preprocessing.html} following the theoretical results of Section~4.3, where we specify $\mu=0$ and $\sigma=1$ for the Gaussian projection.

\vspace{1.5mm}
\noindent\textbf{MHE over  Each Pre-estimated Cluster} Let    $\{\mathcal{B}_1, \mathcal{B}_2,..., \mathcal{B}_k  \}$ be the clustered $k$ hyperspheres where $\mathcal{B}_i$ is  with a center $c_i$ and  radius $r_i$, $\forall 1,2,3,...,k$.
We employ Eq.~(\ref{se_max_min}) to perform MHE over each pre-estimated cluster. The selection of $ {\hat w}_1$ in $\mathcal{B}_i$  follows the sequential optimization of Eq.~(\ref{optimal_w_1}), that is,
\begin{equation}
\argmax_{{\hat w}_1 \in   \mathcal{B}_i}      \mathcal{L}_{\mathbb{E}_{0,d}}( {\hat w}_1 ).
\label{mhe_cluster}
\end{equation}

Assume  that $\mathcal{B}_i$ has ${N}_i$ data, $\hat{ w}_1 \in   \mathcal{B}_i$ thus needs to perform ${N}_i$ times of argmax operation, that is, Eq.~(\ref{mhe_cluster}) needs to initialize  $\hat w_1$ for  ${N}_i$ times.
To reduce the calculation complexity, we degenerate the sequential selection into   vector rotation maximization, which maximizes the vector rotation of $\hat w_1^{t-1}$ and its next initialization $\hat w_1^t$, that is, 
\begin{equation}
\hat{ w}_1^t=  \argmax_{x_t\in \mathcal{B}_i}  \Big(1-{\rm cos}\big(\gamma_{c_i}^{\hat{ w}_1^{t-1}},\gamma_{c_i}^{x_t}\big)\Big),
\label{w_1_t}
\end{equation}
where  $\gamma_{u}^{v}$ denotes the    geodesic from   $u$ to $v$. 
On such setting, we  perform Eq.~(\ref{w_1_t}) $m$ times and obtain an initialization set for $\hat{ w}_1$, that is,  $W=\{\hat{ w}_1^1, {  w}_1^2,...,\hat{ w}_1^m\}$ with $m$ candidates, covering the original $\mathcal{B}_i$.    See Figure~\ref{StartPoint}. With this degeneration, the sequential optimization of Eq.~(\ref{mhe_cluster}) is redefined as 
  \begin{equation}
\argmax_{\hat{ w}_1 \in   W}      \mathcal{L}_{\mathbb{E}_{0,d}}(\hat{ w}_1 ).
\label{degeneration_mhe}
\end{equation}

\begin{figure} 
 \centering
\includegraphics[scale=0.58]{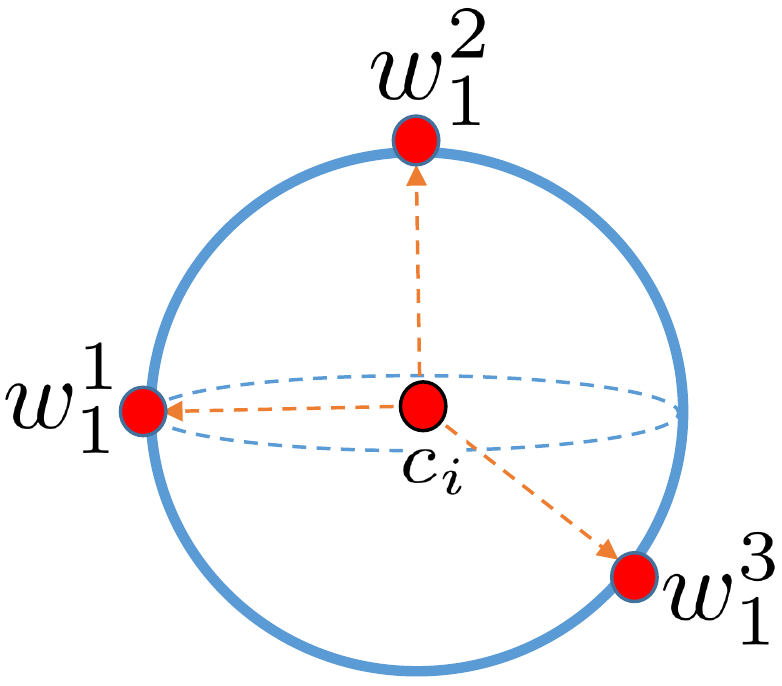}
\caption{Vector rotation maximization.  MHEAL selects the   optimal ${\hat w}_1$   by observing a set of candidate ${\hat w}_1$, $t=1,2,3,..., m$, where  the  vector  over the geodesic  $\gamma_{c_i}^{\hat w_1^{t}}$ holds the maximal disagreement with  $\gamma_{c_i}^{\hat w_1^{t-1}}$.}  
\label{StartPoint}
\vspace{-6pt}
\end{figure}

 {\begin{remark}
 With the vector rotation maximization of  Eq.~(\ref{w_1_t}),  then $\hat w_1 \in \mathcal{B}_i$ of Eq.~(\ref{mhe_cluster})  is updated  into $\hat w_1 \in W$ of Eq~(11). In this way, the calculation complexity of   Eq.~(\ref{mhe_cluster}) degenerates  into $\mathcal{O}(\frac{m^2N^2}{k^4})$, that is, Eq~(\ref{degeneration_mhe}) results in a much lower calculation complexity after restricting $\hat w_1 \in W$.
\end{remark}}

\subsection{Algorithm}
We present MHEAL  algorithm to learn    representative data, which  characterizes decision boundaries    in a balanced manner over each  pre-estiamted  hyperspherical cluster. 

The detailed steps of the MHEAL algorithm are as follows.  Step~3 optimizes a set of spherical clusters adopting hyperspeherical clustering w.r.t. Eq.~(\ref{center}), where we randomly initialize $c_1,c_2,c_3,...,c_k$ from $\mathcal{X}$. For each hypersphere $\mathcal{B}_i$,  Step~5 calculates the optimal initialization on $\hat{ w}_1^*$ following  Eq.~(\ref{degeneration_mhe}), where the  initialization set $W$ is obtained from Eq.~(\ref{w_1_t});    Steps~6 to 8 select  $\tau$  data  to match $\mathcal{B}_i$. Step~9 merges  $\tau$ representative data from $\mathcal{B}_i$ into the representation set $\mathcal{S}$, for any $i\leq k$. In this way, the final outputs on    $\mathcal{S}$  (\ie, $\mathcal{S}^*$) has $\tau k$  representative data, which approximately represent the $k$ spherical clusters.

\begin{algorithm}[t]
  \caption{Minimizing Hyperspherical  Energy-based Active Learning (MHEAL)}
   \textbf{Input:} Data set $\mathcal{X}$, number of clusters $k$, number of representative data $\tau k$.  \\
   \textbf{Initialize:} Randomly initialize $c_1,c_2,c_3,...,c_k$ from $\mathcal{X}$,  and then set   $\mathcal{S}, \mathcal{S}^* =\emptyset$. \\ 
   Solve $ \argmin_{c_k}   \sum_{x_i} \Big(1-{\rm cos}(c_k,x_i)\Big)$.\\

\For{$i=1,2,3,...,k$}{
    $\hat {w}_1^*=\argmax_{\hat{ w}_1 \in   W}      \mathcal{L}_{\mathbb{E}_{0,d}}(\hat{ w}_1 )$ and $\mathcal{S}=\hat{w}_1^*$.\\
\For{$t=2,3,...,\tau$}{
$ x_t^*=\argmax_{x_j\in \mathcal{B}_i}  \min_{w_i \in \mathcal{S}    }         \|w_i-x_j\|,$  $\mathcal{S} = \mathcal{S}\cup x_{t-1}^*$.
}

$\mathcal{S}^*=\mathcal{S}^*\cup \mathcal{S}$.
}
 \textbf{Output:}  Representative set  $\mathcal{S}^*$.
\end{algorithm}
}

\section{Improving Generalization for MHEAL}
 {Section~6.1 presents the generalization analysis of MHEAL on error and label complexity bounds. Section~6.2  then presents the empirical study to explain the theoretical results.}

 {\subsection{Error and Label Complexity Bounds}}
MHEAL adopts a version space view that yields a clustering manner to extract   representative data. This    reduces the typical label complexity bound  of VSAL into a  set of local polynomial  label complexities \cite{cortes2020region} over clusters.  Here, we follow the  IWAL  \cite{beygelzimer2009importance} scenario to present the convergence analysis  of MHEAL on the generalization error and label complexity bounds.

\vspace{1.5mm}
\noindent\textbf{IWAL scenario.} Given a finite hypothesis class $\mathcal{H}$, IWAL tries to update  the current hypothesis $h_t\in \mathcal{H}$  at $t$-time into the optimal hypothesis $h^*\in \mathcal{H}$. Let IWAL perform $T$ rounds of querying from $\mathcal{X}$,  assume that $\ell(\cdot,\cdot)$ denotes the loss of mapping $\mathcal{X}$ into $\mathcal{Y}$ with multi-class setting, we define the total loss of the  $T$ rounds of querying as $R(h_T)=\sum_{i=1}^{T}\frac{Q_t}{p_i}\ell(h(x_t), y_t)$, where $y_t$ denotes the label of $x_t$, $Q_t$ satisfies the Bernoulli distribution of $Q_t\in \{0,1\}$, and $\frac{1}{p_i}$ denotes the weight of sampling $x_t$. The sampling process adopts an error disagreement to control the hypothesis updates: 
\begin{equation}
\theta_{\rm IWAL}= \mathbb{E}_{x_t\in \mathcal{X}}   \mathop{{\rm sup}}\limits_{h\in B(h^*,r)} \left \{  \frac{ \ell(h(x_t),\mathcal{Y})-\ell(h^*(x_t),\mathcal{Y})   }{r}     \right\},
\end{equation}
 \cao{where $B(h^*,r)$ denotes the ball with a center  $h^*$ and radius $r$. 
 In hypothesis class,
$\ell(h(x_t),\mathcal{Y})-\ell(h^*(x_t),\mathcal{Y}) $ denotes the maximal hypothesis disagreement over the loss $\ell$. On this setting, $ \theta_{\rm IWAL} $ denotes the minimal step on the significant update of hypothesis.} 
 Once the hypothesis update w.r.t. error  after adding $x_t$ is larger than $\theta_{\rm IWAL}$, IWAL solicits $x_t$ as a significant update. Otherwise,  $x_t$ will not be an ideal update.

\vspace{1.5mm}
\noindent\textbf{MHEAL scenario.} Given the input dataset $\mathcal{X}$ with $n$ samples, it is divided into $k$ clusters: $\{\mathcal{B}_1, \mathcal{B}_2,..., \mathcal{B}_k\}$, where  $\mathcal{B}_i$ has $N_i$ samples. MHEAL performs IWAL for any $\mathcal{B}_i$. Specifically, MHEAL uses a new error disagreement $\theta_{\rm MHEAL}$ to control the hypothesis updates: 
\begin{equation}
\theta_{\rm MHEAL}= \mathbb{E}_{x_t\in \mathcal{B}_i}   \mathop{{\rm sup}}\limits_{h\in B(h^*,r)} \left \{  \frac{ \ell(h(x_t),\mathcal{Y})-\ell(h^*(x_t),\mathcal{Y})   }{r}       \right\}.\end{equation}

\begin{theorem}
Given $T$ rounds of querying by employing IWAL, let $\mathcal{Q}$ be the number of ground-truth queries, i.e., label complexity. If   MHEAL performs IWAL for any $\mathcal{B}_i$, each cluster   will have $\tau=T/k$ rounds of querying. Then, with a probability at least  $1-\delta$, for all $\delta>0$, for any $T>0$,  the error disagreement of   $R(h_\tau)$ and   $R(h^*)$ of MHEAL  is bounded by $k$ times of  polynomial  label complexities over each cluster
\begin{equation*}
\begin{split}
&R(h_\tau)-R(h^*) \\
&\leq k\times \max_{\mathcal{H}_i, i=1,...,k} \Bigg\{\frac{2}{\tau}  \Bigg[\sqrt{\sum_{t=1}^{\tau}p_t}+6\sqrt{{\rm log}\Big[\frac{2(3+\tau)\tau^2}{\delta}\Big] } \Bigg] \\
&\times  \sqrt{{\rm log}\Big[\frac{16\tau^2|\mathcal{H}_i|^2 {\rm log}\tau}{\delta}\Big]}\Bigg\}.
\end{split}
\end{equation*} 
Then, with a probability at least  $1-2\delta$, for all $\delta>0$, the label complexity of MHEAL can be bounded by
\begin{equation*}
\begin{split}
& \mathcal{Q} \leq 8k   \times \max_{\mathcal{H}_i, i=1,...,k} K_\ell\Bigg\{ \Big[\sum_{j=1}^{N_i} \theta_{\rm MHEAL} R_j^*\tau p_j\Big]\\
&\ \ +\sum_{j=1}^{N_i} O\Bigg(\sqrt{R_j^*\tau p_j{\rm log}\Big[\frac{\tau|\mathcal{H}_i|N_i}{\delta} \Big]}\Bigg)\!
+\!O\Bigg(N_i {\rm log}^3\Big(\frac{\tau|\mathcal{H}_i|N_i}{\delta}\Big)\Bigg)\Bigg\}
\end{split}
\end{equation*} 
where $K_f$ is the slope asymmetry  over the limited loss $\ell$ on $\mathcal{B}_i$, i.e. 
$\ell_{\mathcal{B}_i}$,   $K_f= \mathop{{\rm sup}}\limits_{x_t', x_t\in \mathcal{B}_i} \left |\frac{{\rm max} \ \ell_{\mathcal{B}_i}(h(x_t), \mathcal{Y})-\ell_{\mathcal{B}_i}(h(x_t'), \mathcal{Y}) }    {{\rm min} \ \ell_{\mathcal{B}_i}(h(x_t), \mathcal{Y})- \ell_{\mathcal{B}_i}(h(x_t'), \mathcal{Y}) } \right|$, $R_j^*$ denotes the best-in-class risk at $j$-time querying, and $|\mathcal{H}|$ denotes the element number of  $\mathcal{H}$ ($\ell_0$-norm).  
\end{theorem}

\begin{figure*} 
\subfloat[skin]{
\begin{minipage}[t]{0.156\textwidth}
\centering
\includegraphics[width=1.17in,height=0.975in]{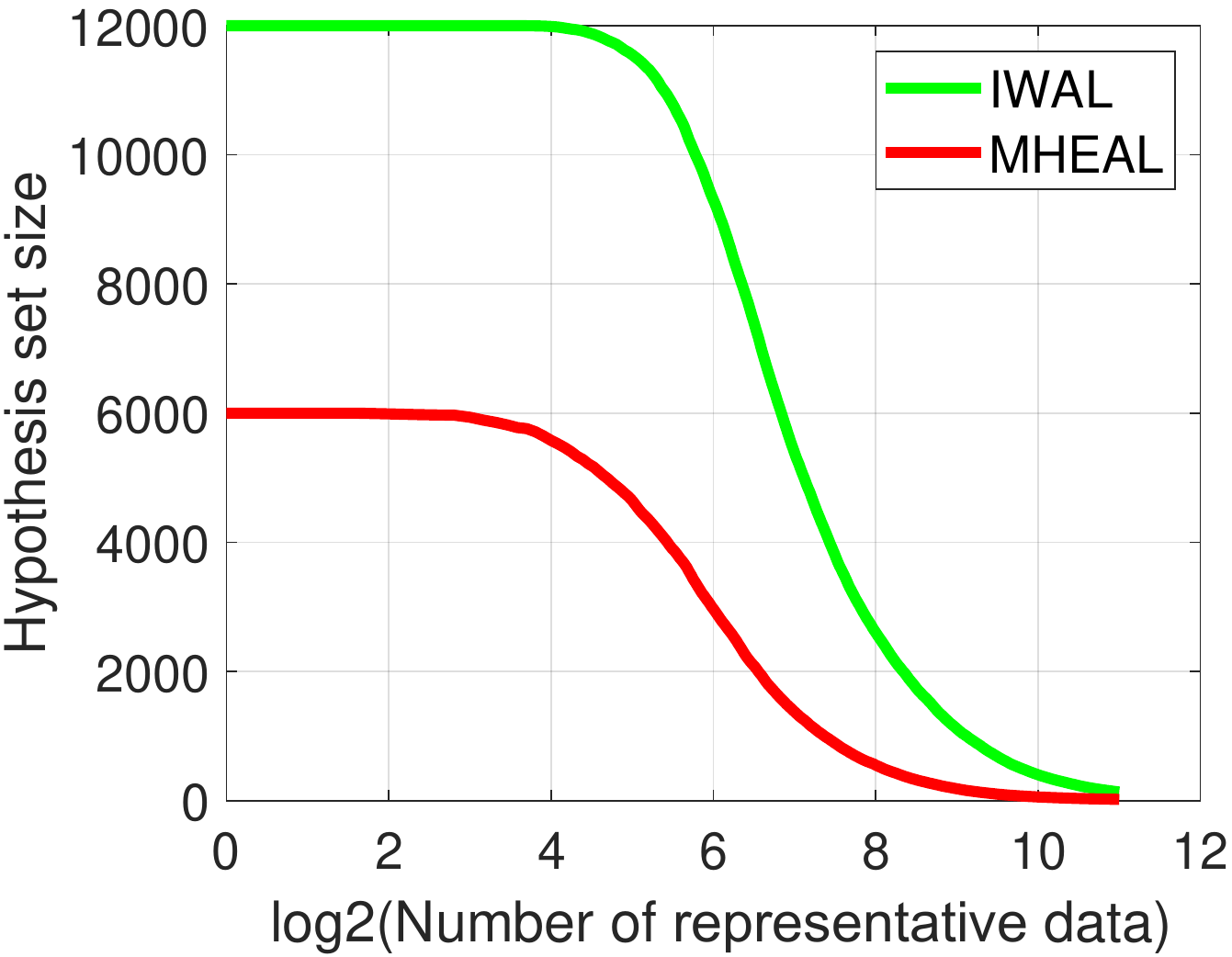}
\end{minipage}
}
\subfloat[shuttle]{
\begin{minipage}[t]{0.156\textwidth}
\centering
\includegraphics[width=1.17in,height=0.975in]{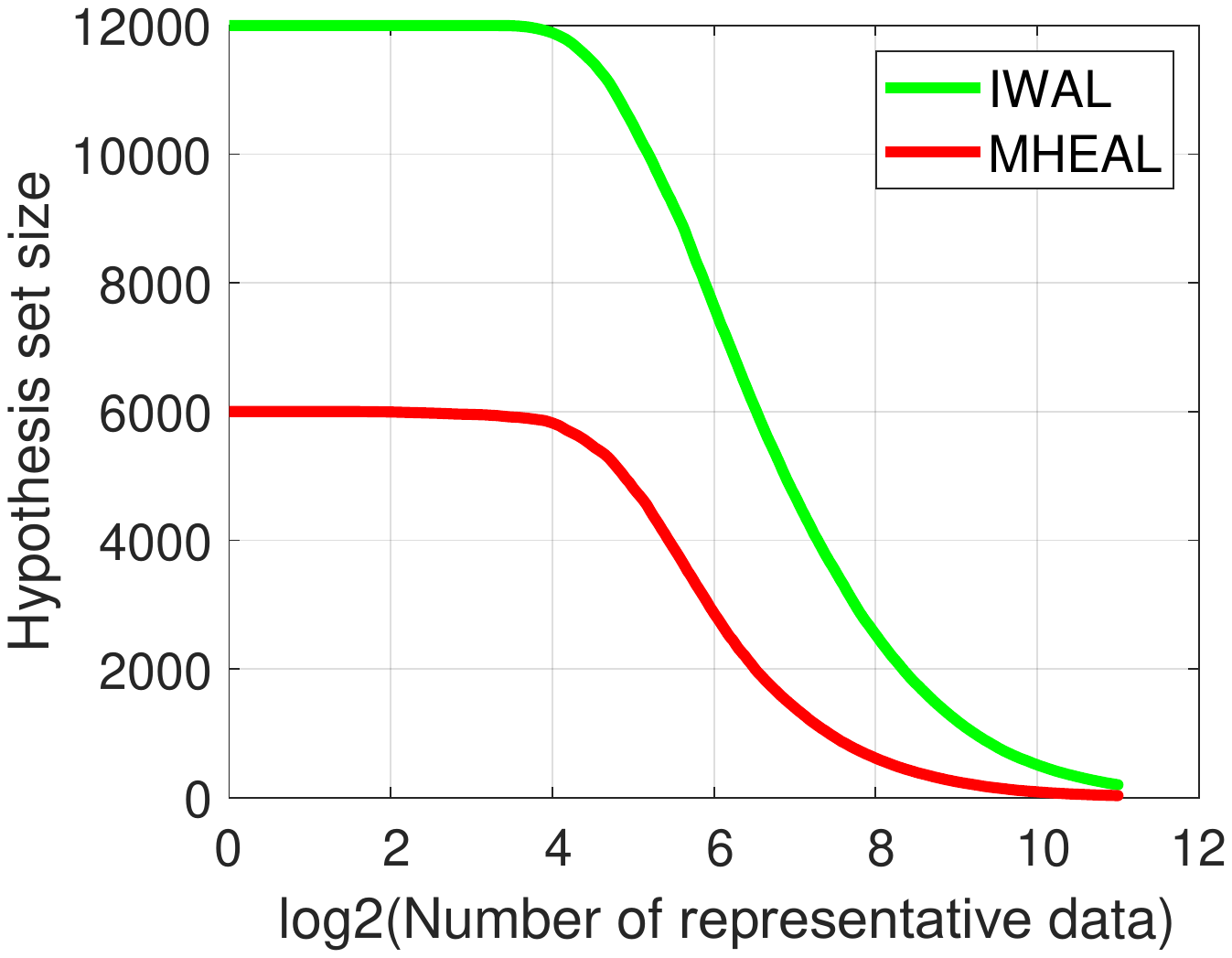}
\end{minipage}
} 
\subfloat[magic04]{
\begin{minipage}[t]{0.156\textwidth}
\centering
\includegraphics[width=1.17in,height=0.975in]{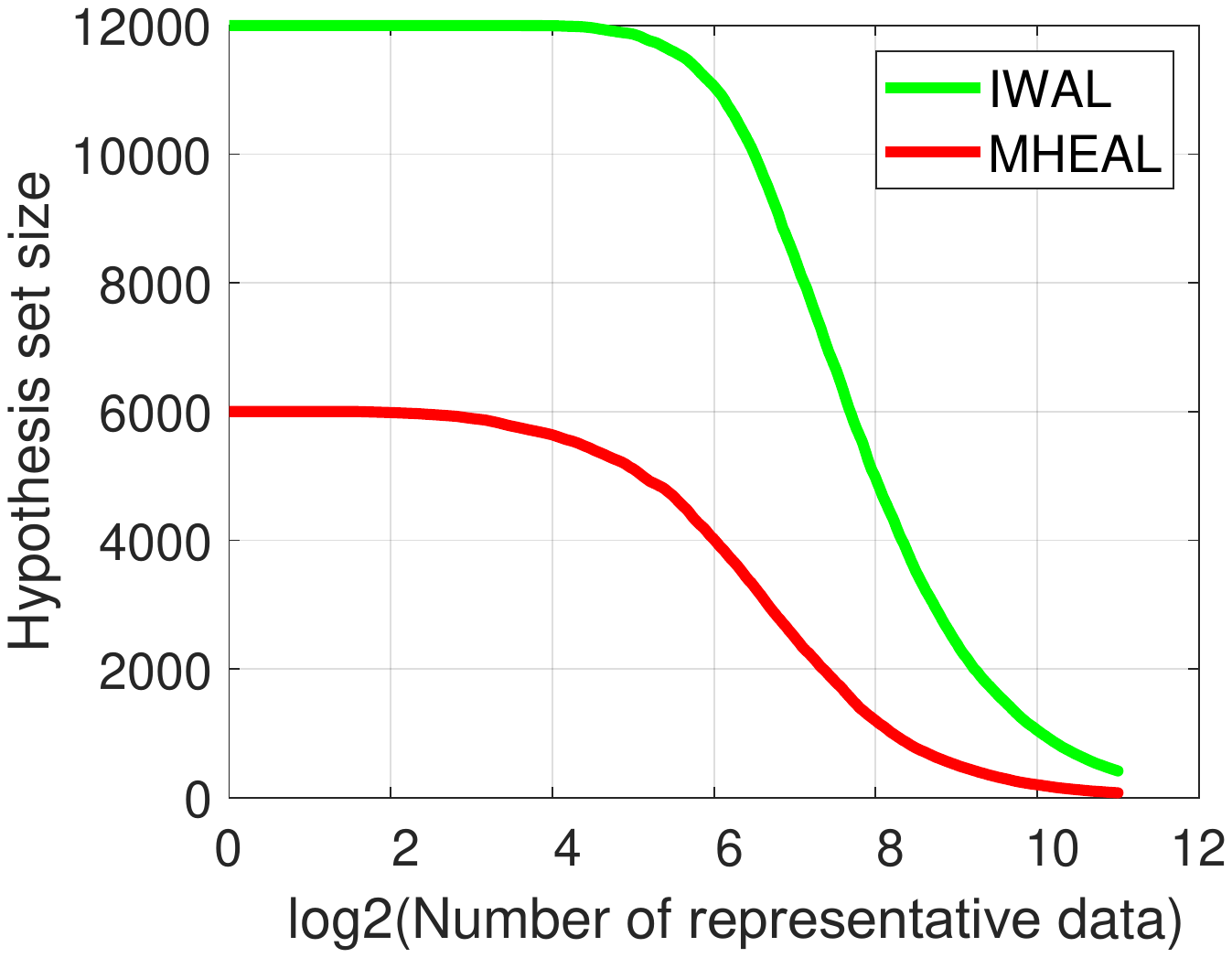}
\end{minipage}
} 
\subfloat[jm1]{
\begin{minipage}[t]{0.156\textwidth}
\centering
\includegraphics[width=1.17in,height=0.975in]{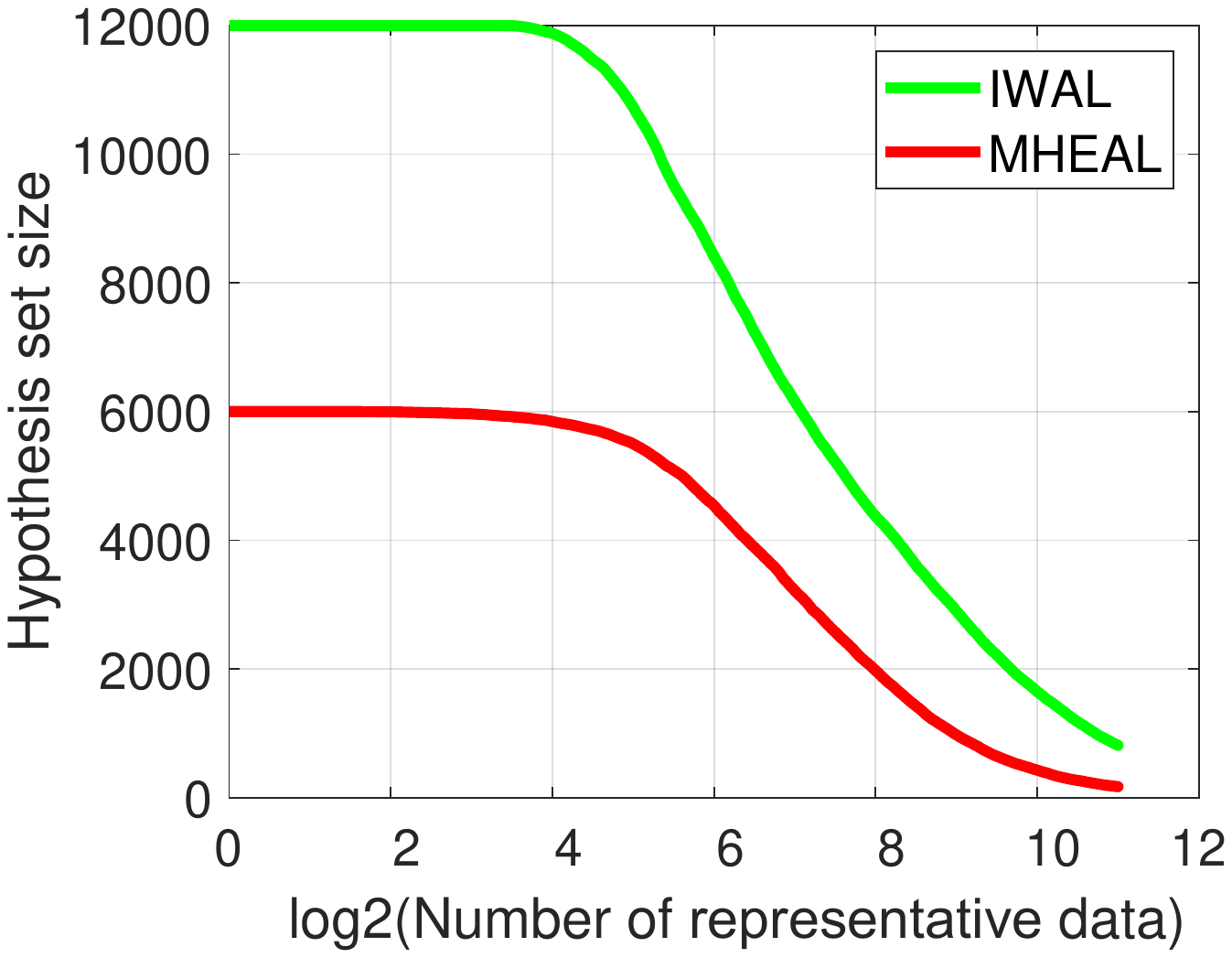}
\end{minipage}
} 
\subfloat[covtype]{
\begin{minipage}[t]{0.156\textwidth}
\centering
\includegraphics[width=1.17in,height=0.975in]{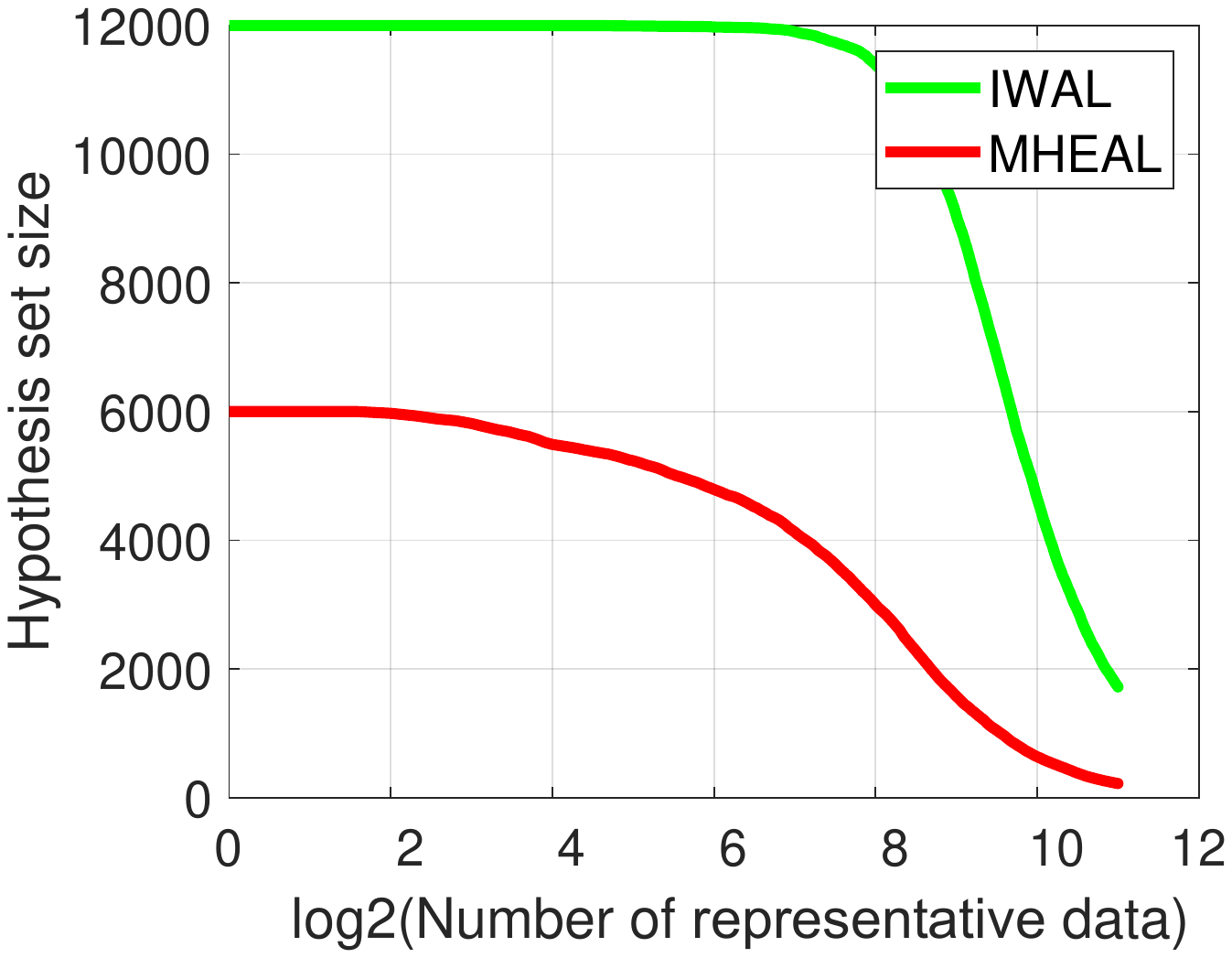}
\end{minipage}
} 
\subfloat[nomao]{
\begin{minipage}[t]{0.156\textwidth}
\centering
\includegraphics[width=1.17in,height=0.975in]{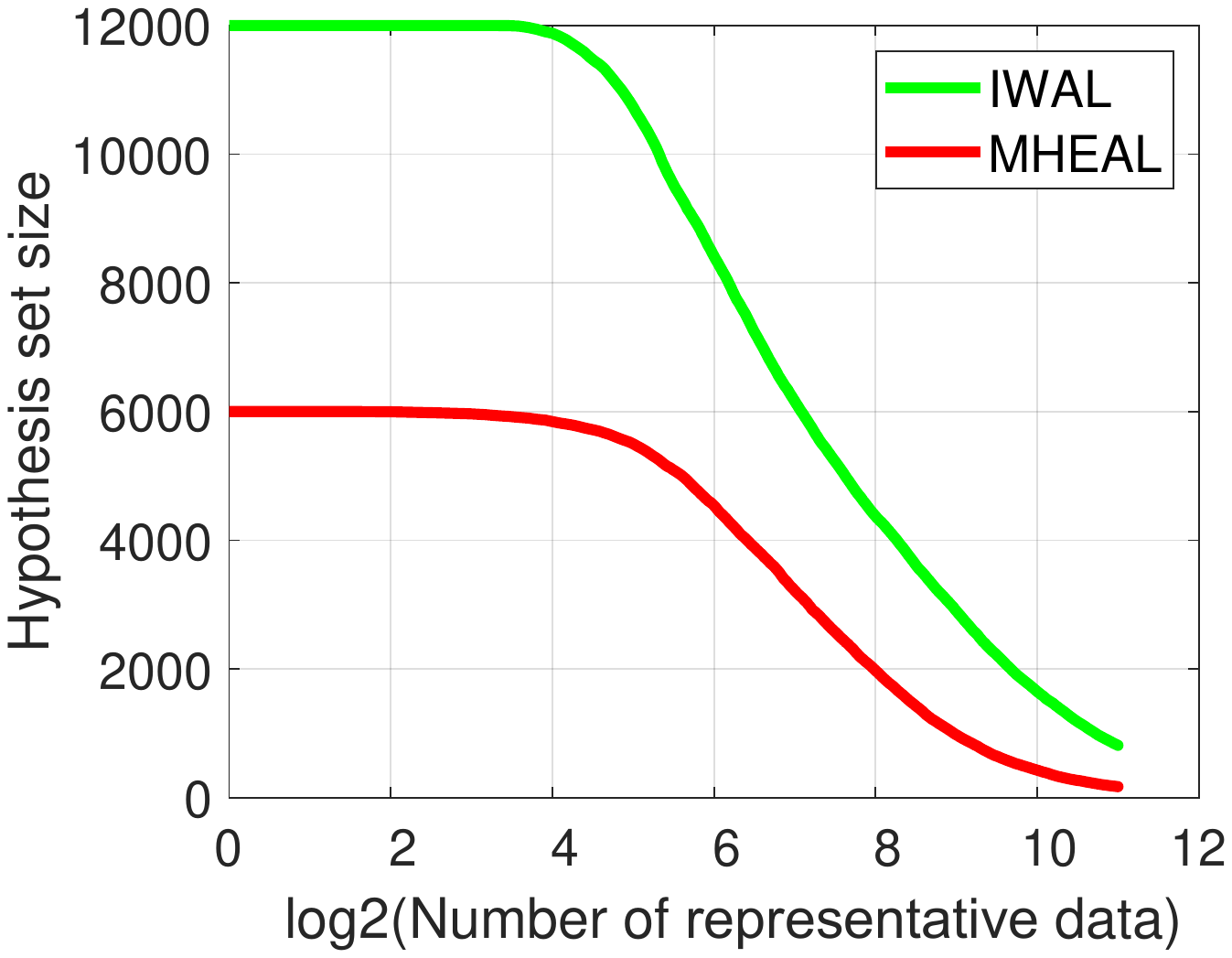}
\end{minipage}
} 
\caption{Hypothesis-pruning speeds of  IWAL and MHEAL. The results show that MHEAL can prune the hypothesis class faster than typical IWAL.
 }  
 \label{Hypothesis_pruning_speed}
\end{figure*}

\begin{figure*} 
\subfloat[skin]{
\begin{minipage}[t]{0.156\textwidth}
\centering
\includegraphics[width=1.17in,height=0.975in]{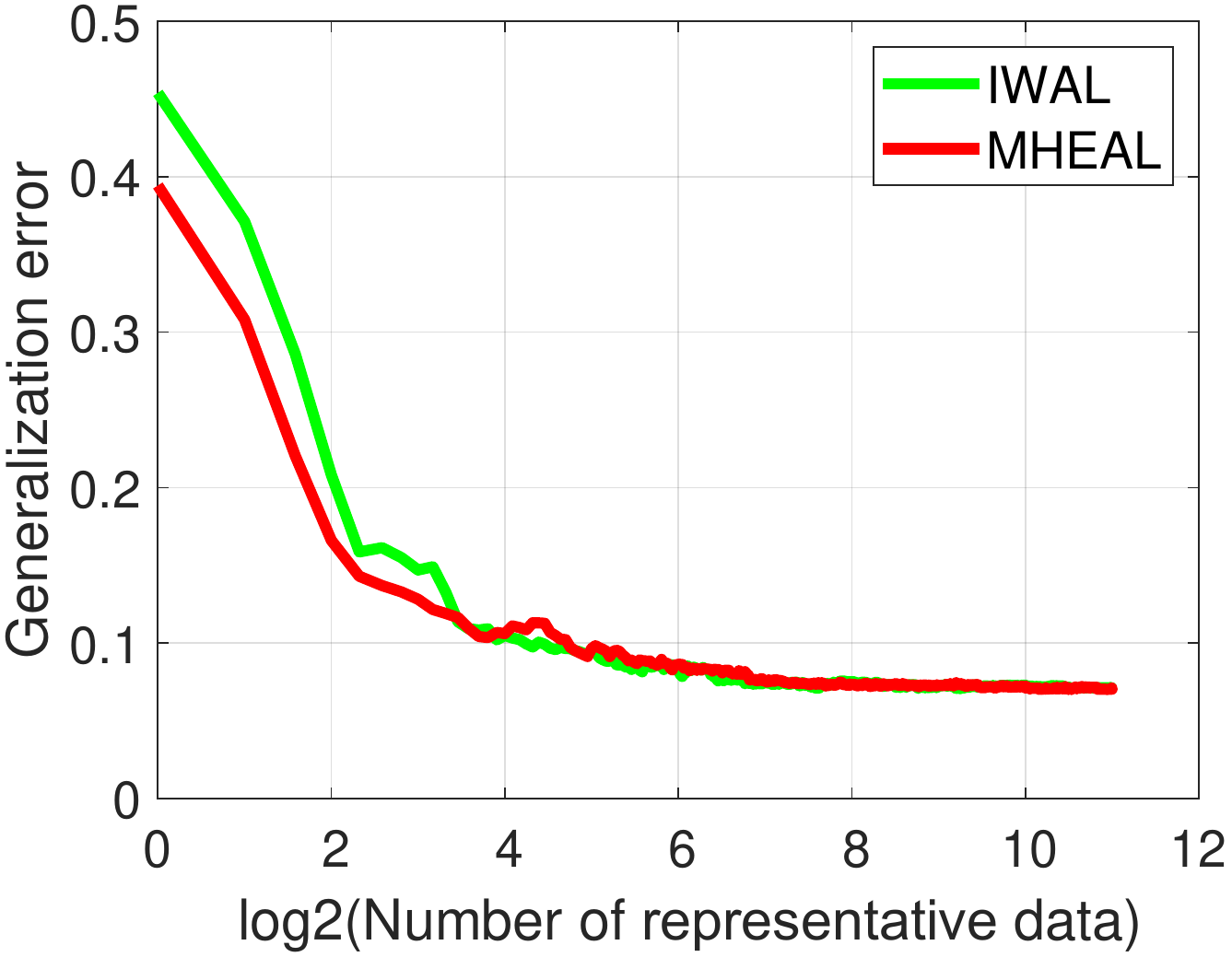}
\end{minipage}
}
\subfloat[shuttle]{
\begin{minipage}[t]{0.156\textwidth}
\centering
\includegraphics[width=1.17in,height=0.975in]{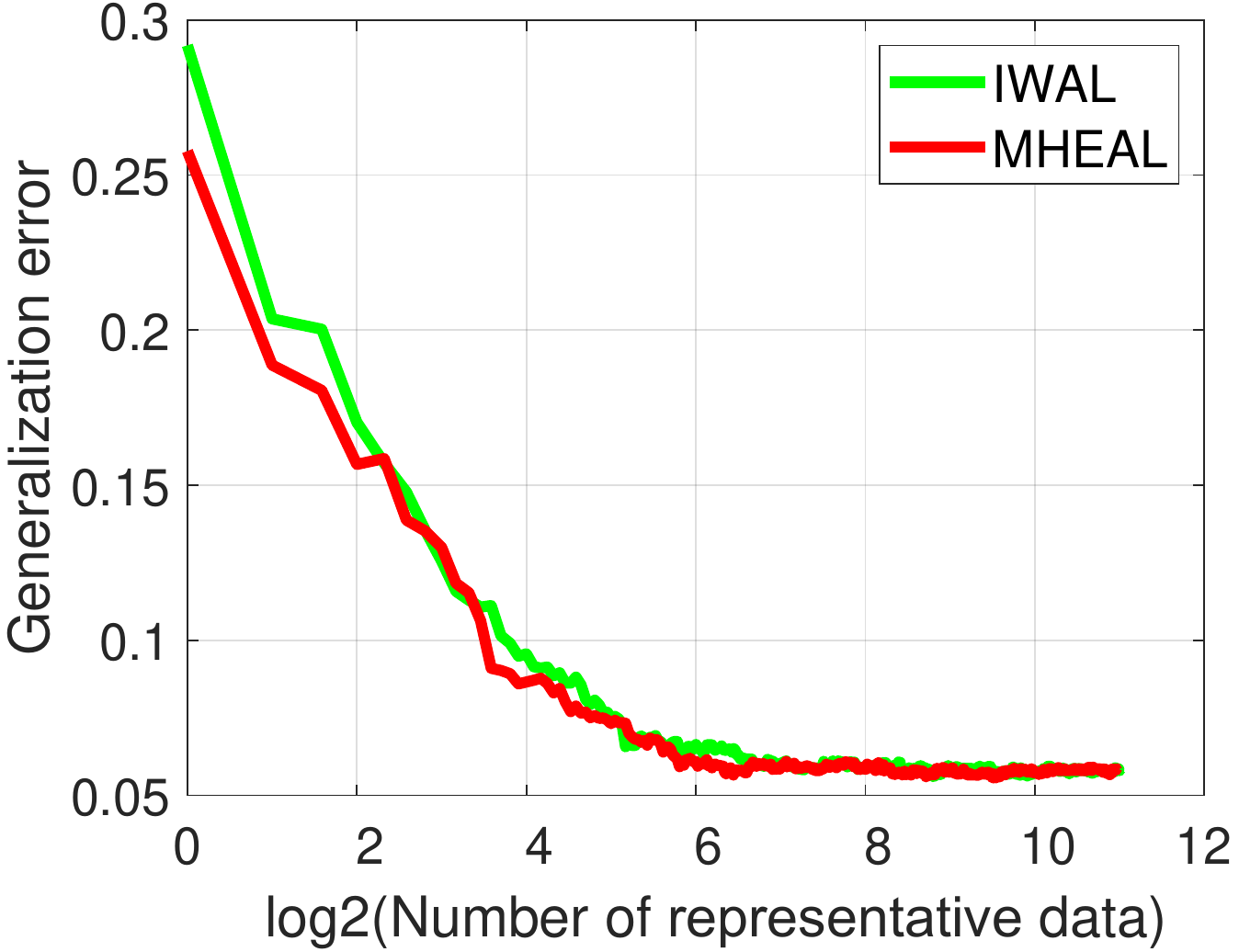}
\end{minipage}
} 
\subfloat[magic04]{
\begin{minipage}[t]{0.156\textwidth}
\centering
\includegraphics[width=1.17in,height=0.975in]{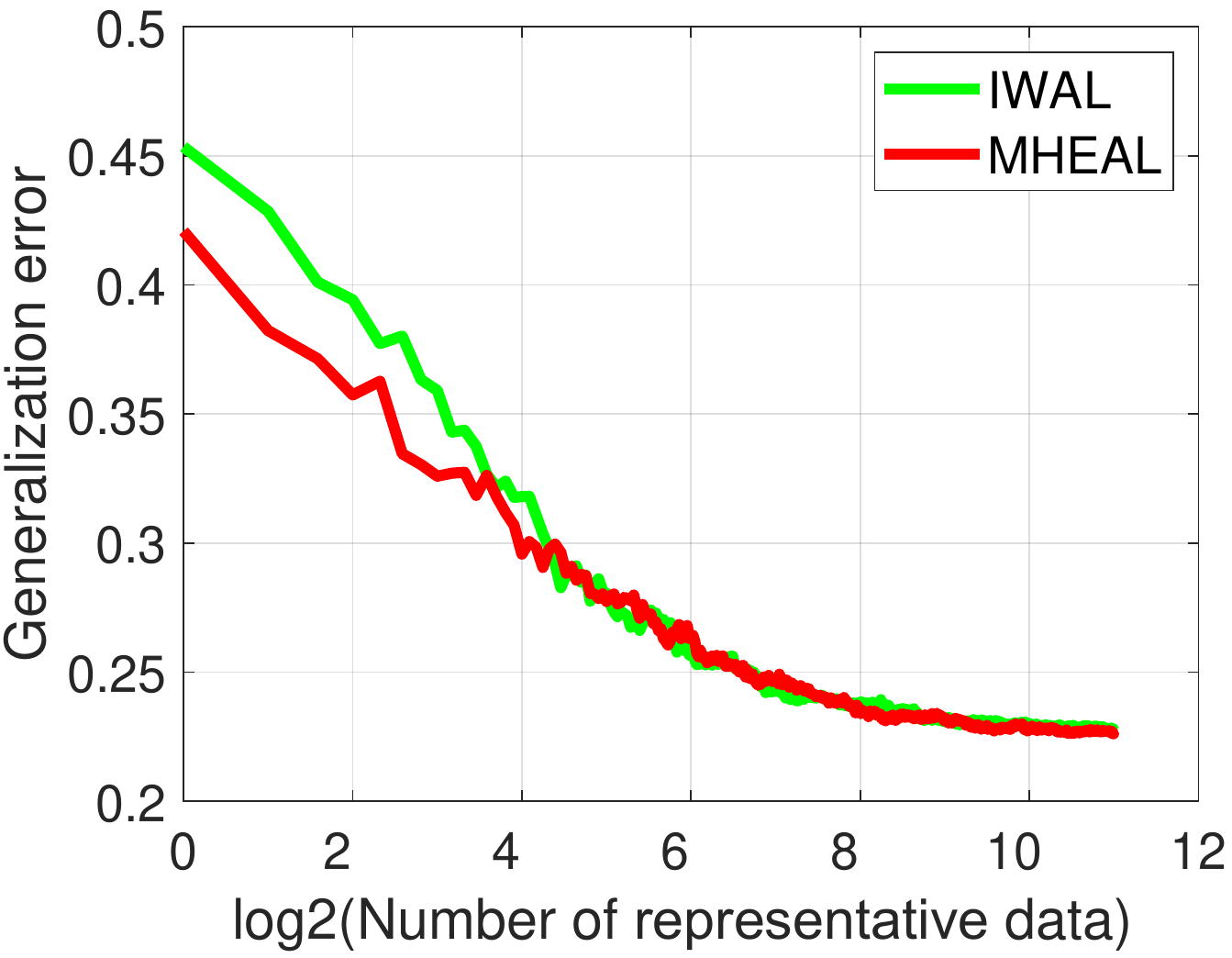}
\end{minipage}
} 
\subfloat[jm1]{
\begin{minipage}[t]{0.156\textwidth}
\centering
\includegraphics[width=1.17in,height=0.975in]{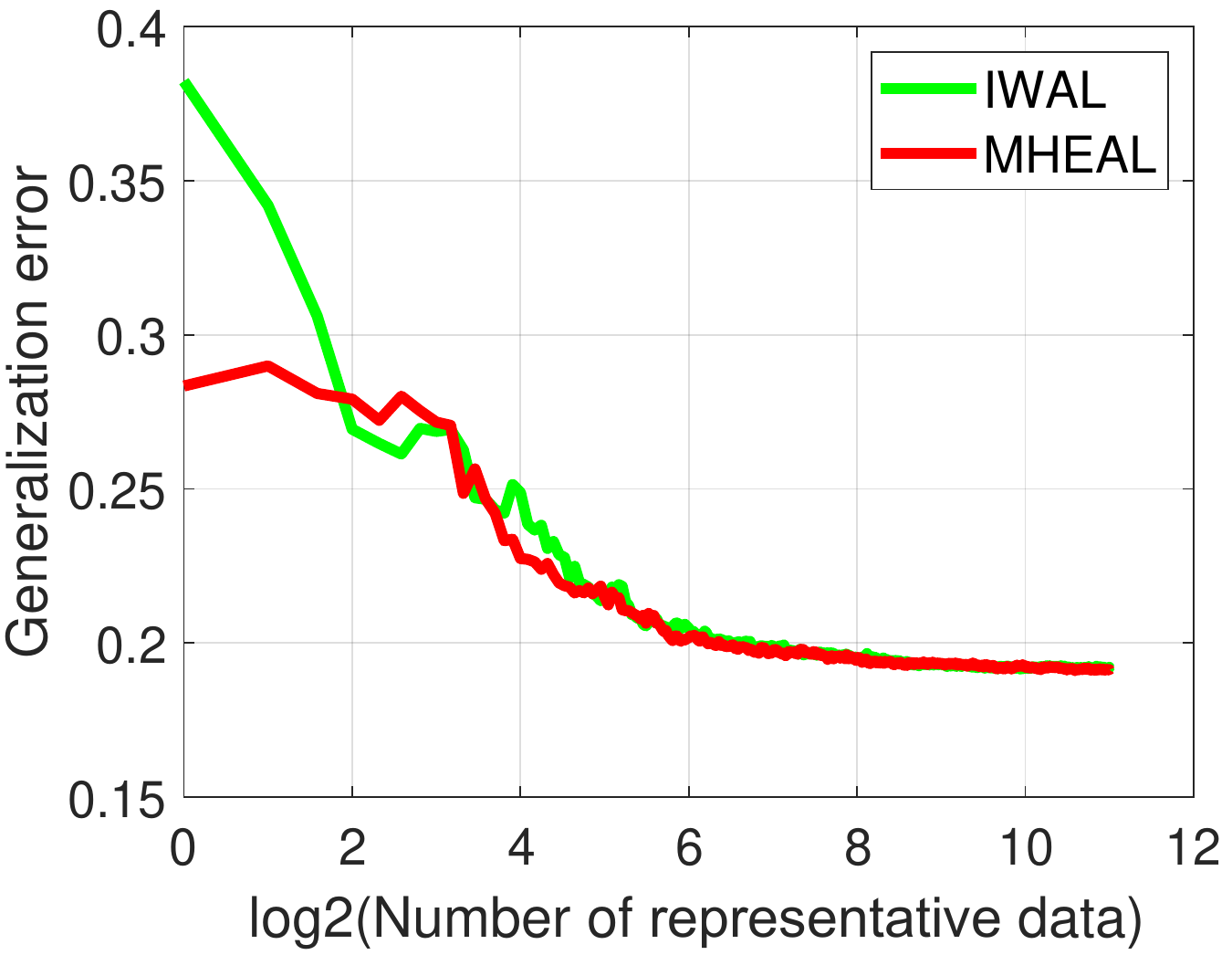}
\end{minipage}
} 
\subfloat[covtype]{
\begin{minipage}[t]{0.156\textwidth}
\centering
\includegraphics[width=1.17in,height=0.975in]{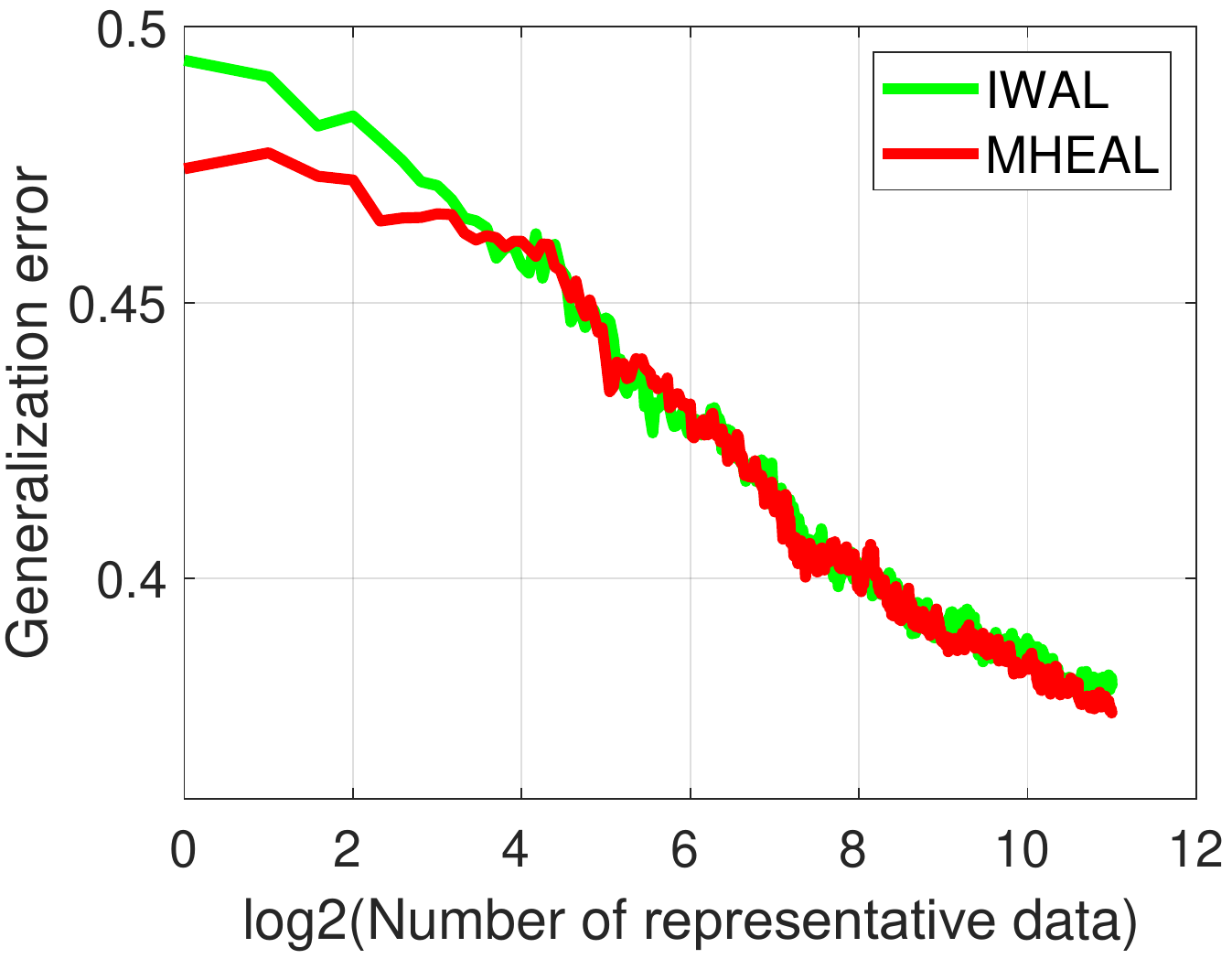}
\end{minipage}
} 
\subfloat[nomao]{
\begin{minipage}[t]{0.156\textwidth}
\centering
\includegraphics[width=1.17in,height=0.975in]{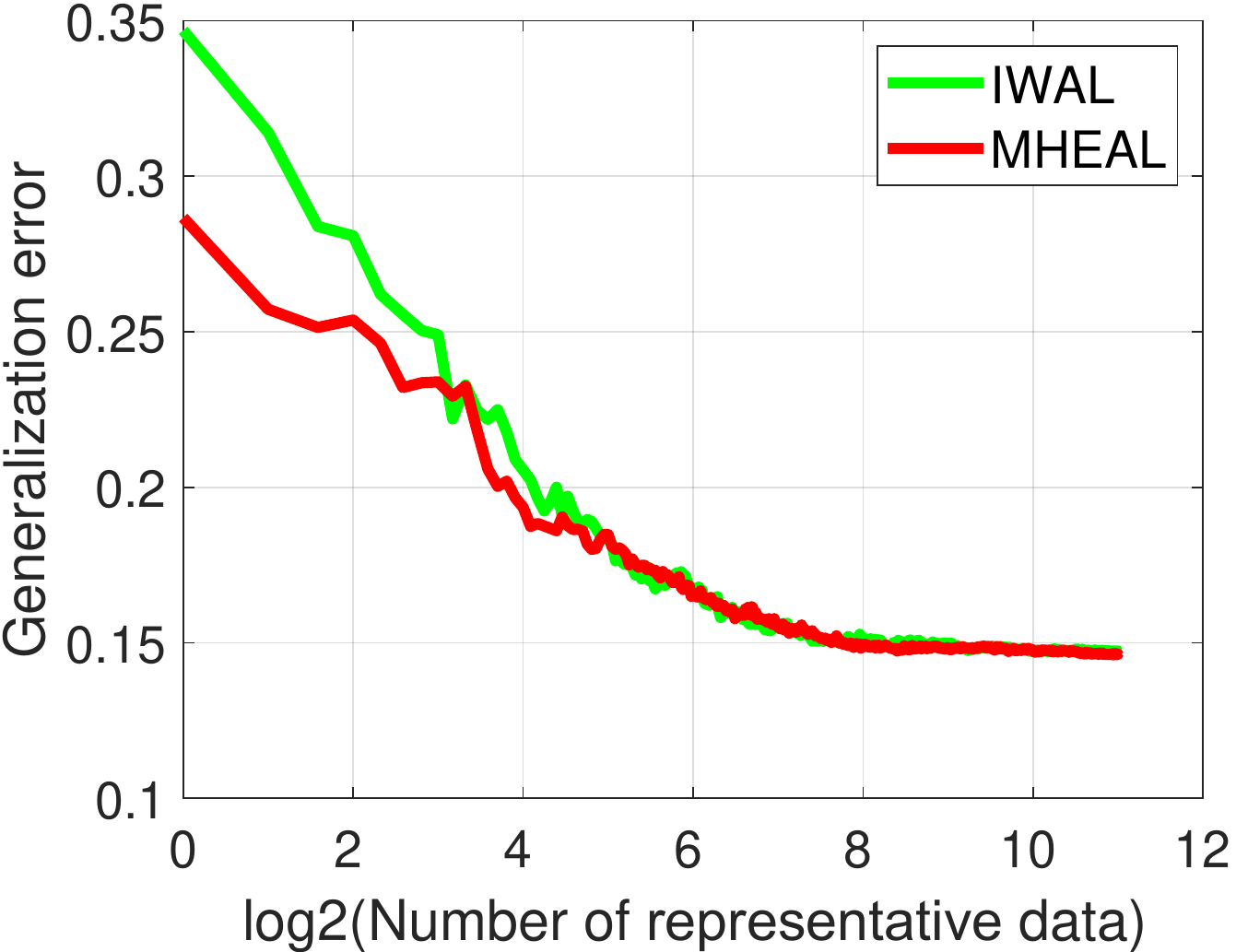}
\end{minipage}
} 
\caption{Generalization errors of  IWAL and MHEAL. The results show that  MHEAL can obtain a tighter error than  IWAL.
 }  
 \label{Generalization_errors}
\end{figure*}

\begin{figure*} 
\subfloat[skin]{
\begin{minipage}[t]{0.156\textwidth}
\centering
\includegraphics[width=1.17in,height=0.975in]{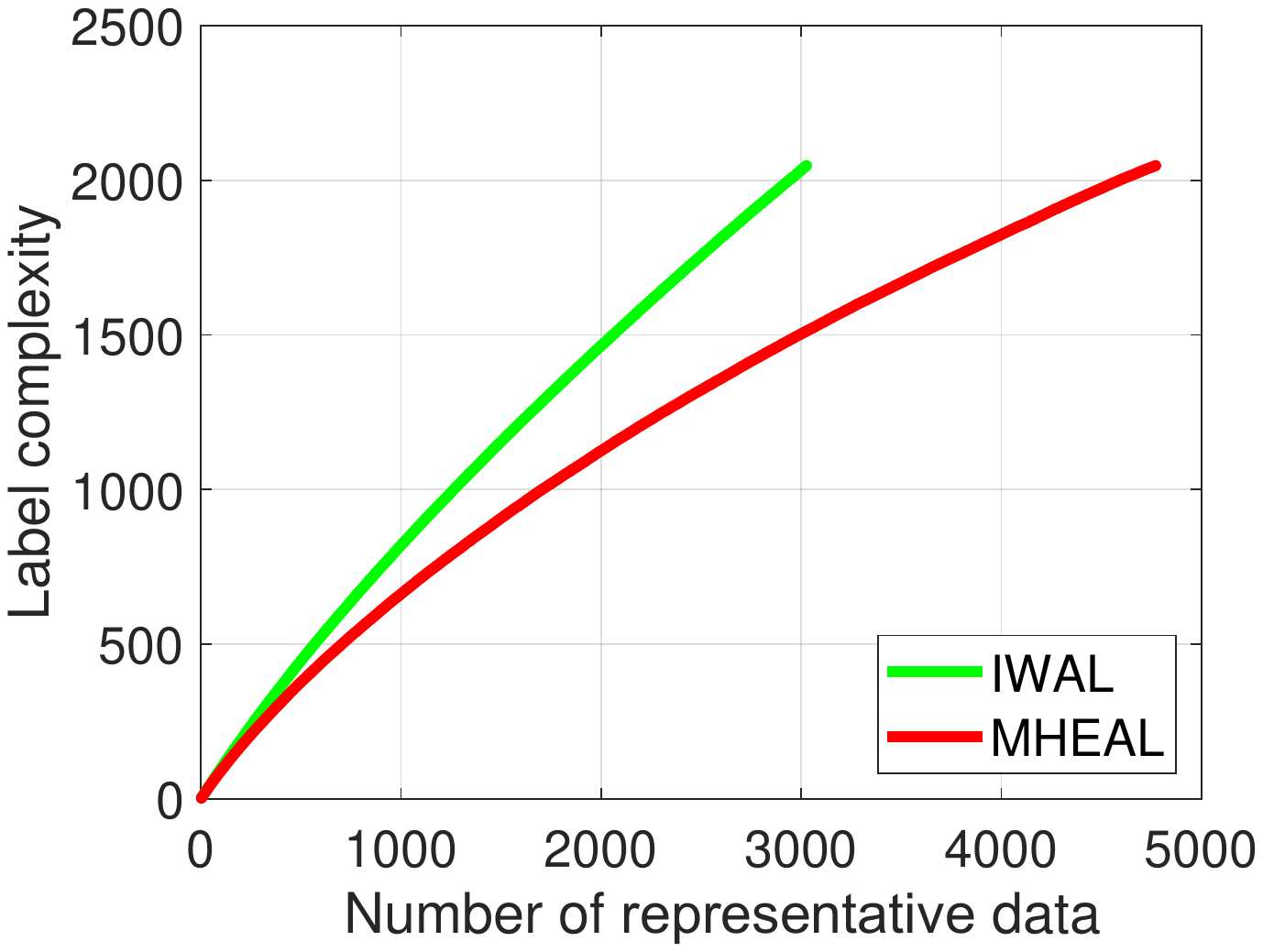}
\end{minipage}
}
\subfloat[shuttle]{
\begin{minipage}[t]{0.156\textwidth}
\centering
\includegraphics[width=1.17in,height=0.975in]{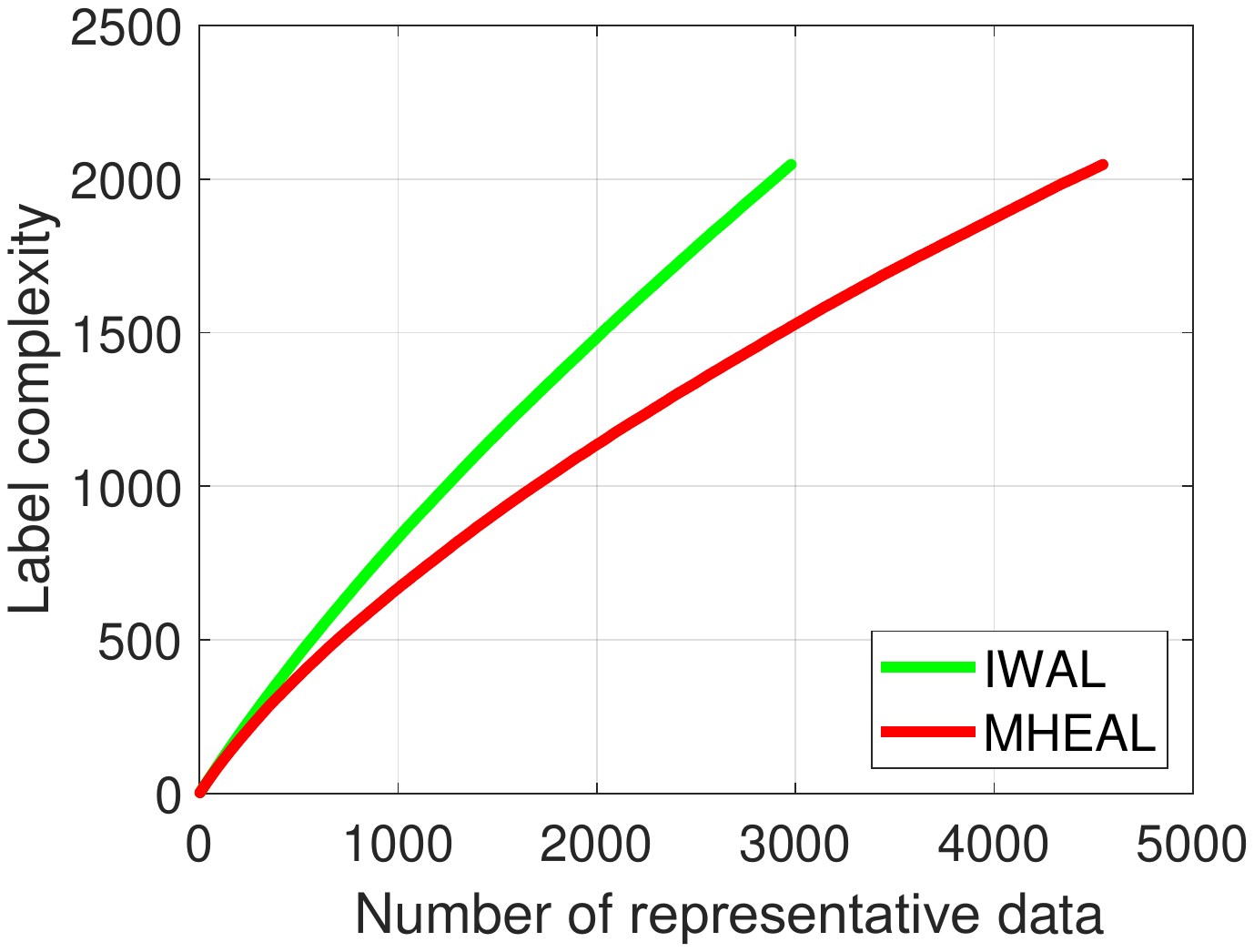}
\end{minipage}
} 
\subfloat[magic04]{
\begin{minipage}[t]{0.156\textwidth}
\centering
\includegraphics[width=1.17in,height=0.975in]{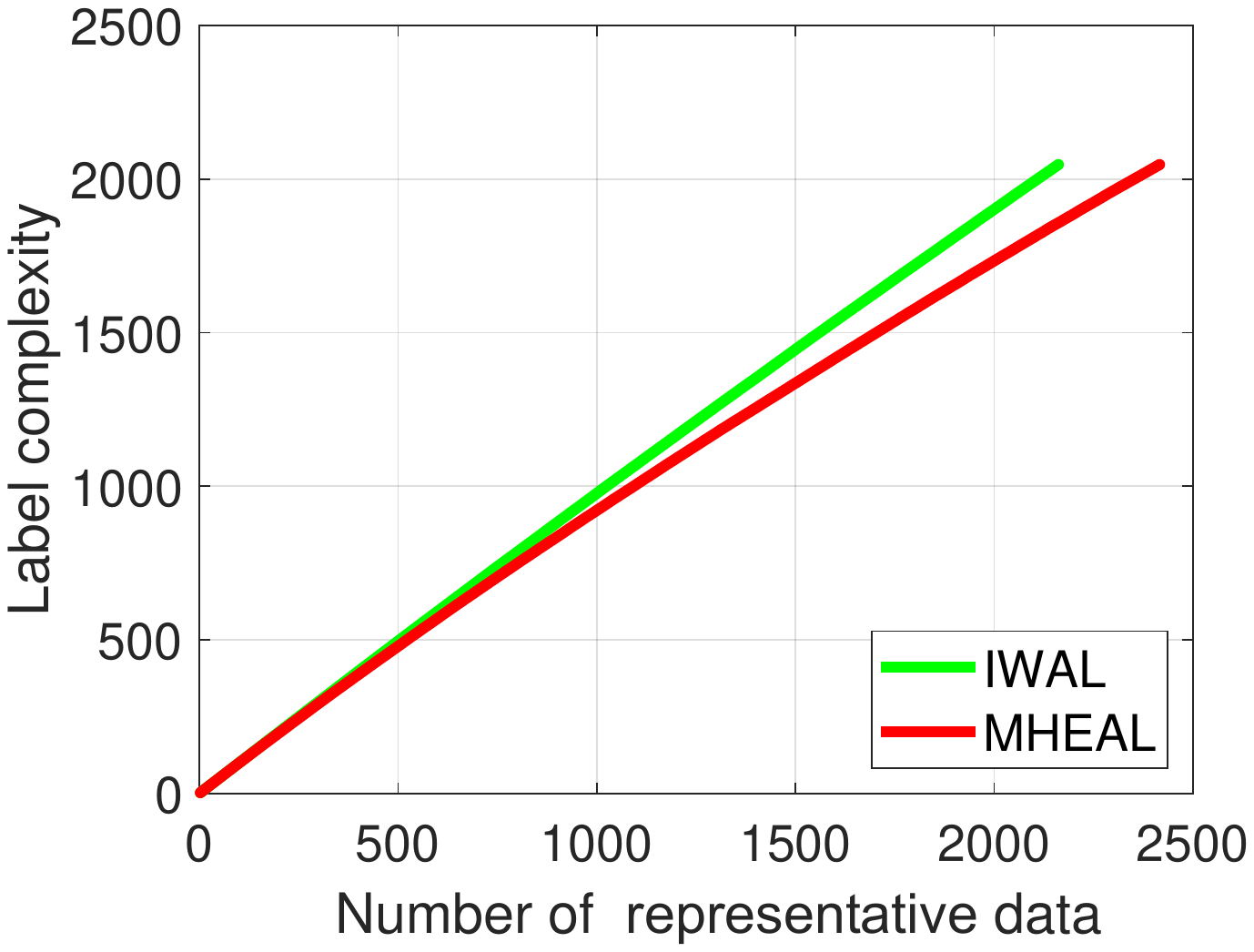}
\end{minipage}
} 
\subfloat[jm1]{
\begin{minipage}[t]{0.156\textwidth}
\centering
\includegraphics[width=1.17in,height=0.975in]{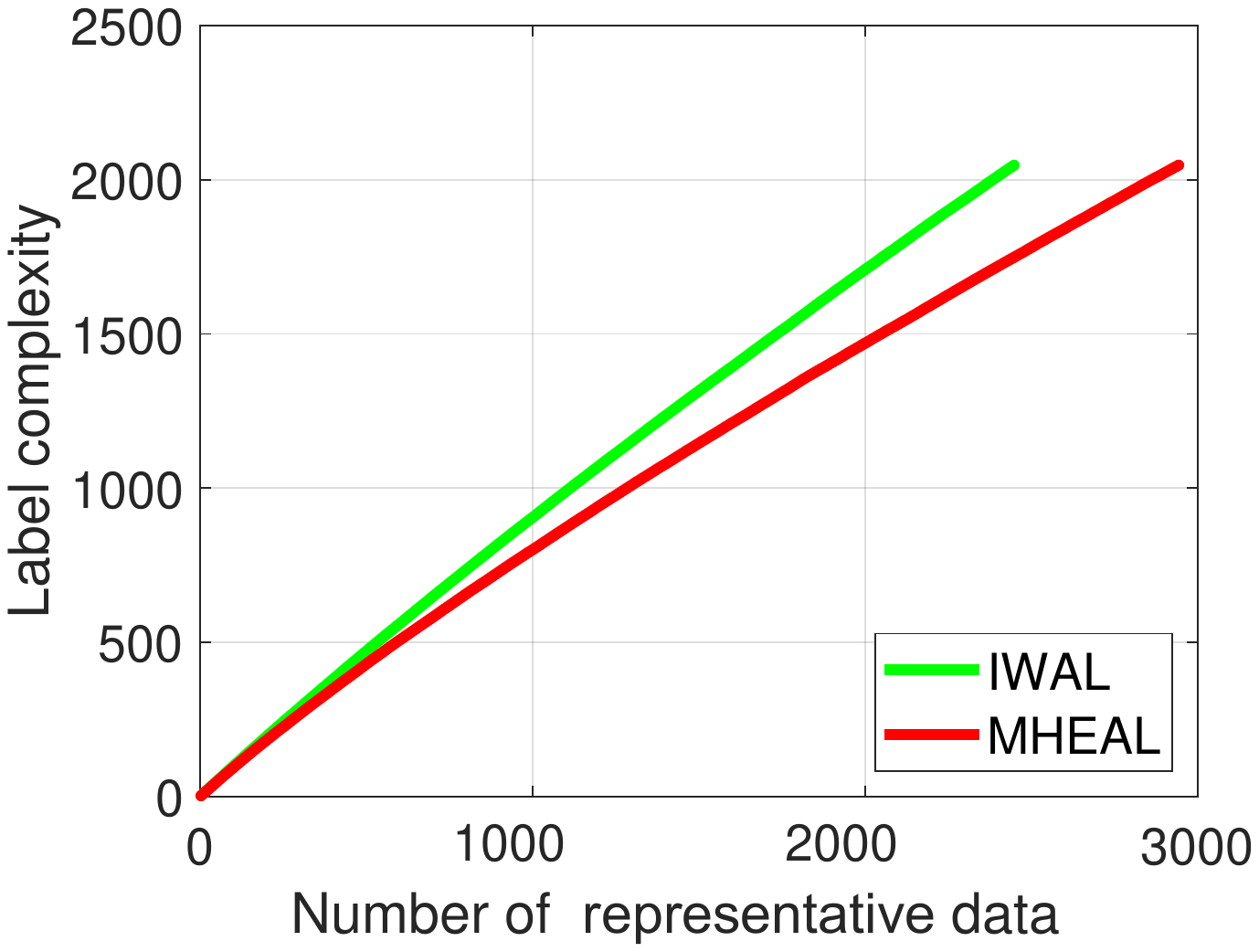}
\end{minipage}
} 
\subfloat[covtype]{
\begin{minipage}[t]{0.156\textwidth}
\centering
\includegraphics[width=1.17in,height=0.975in]{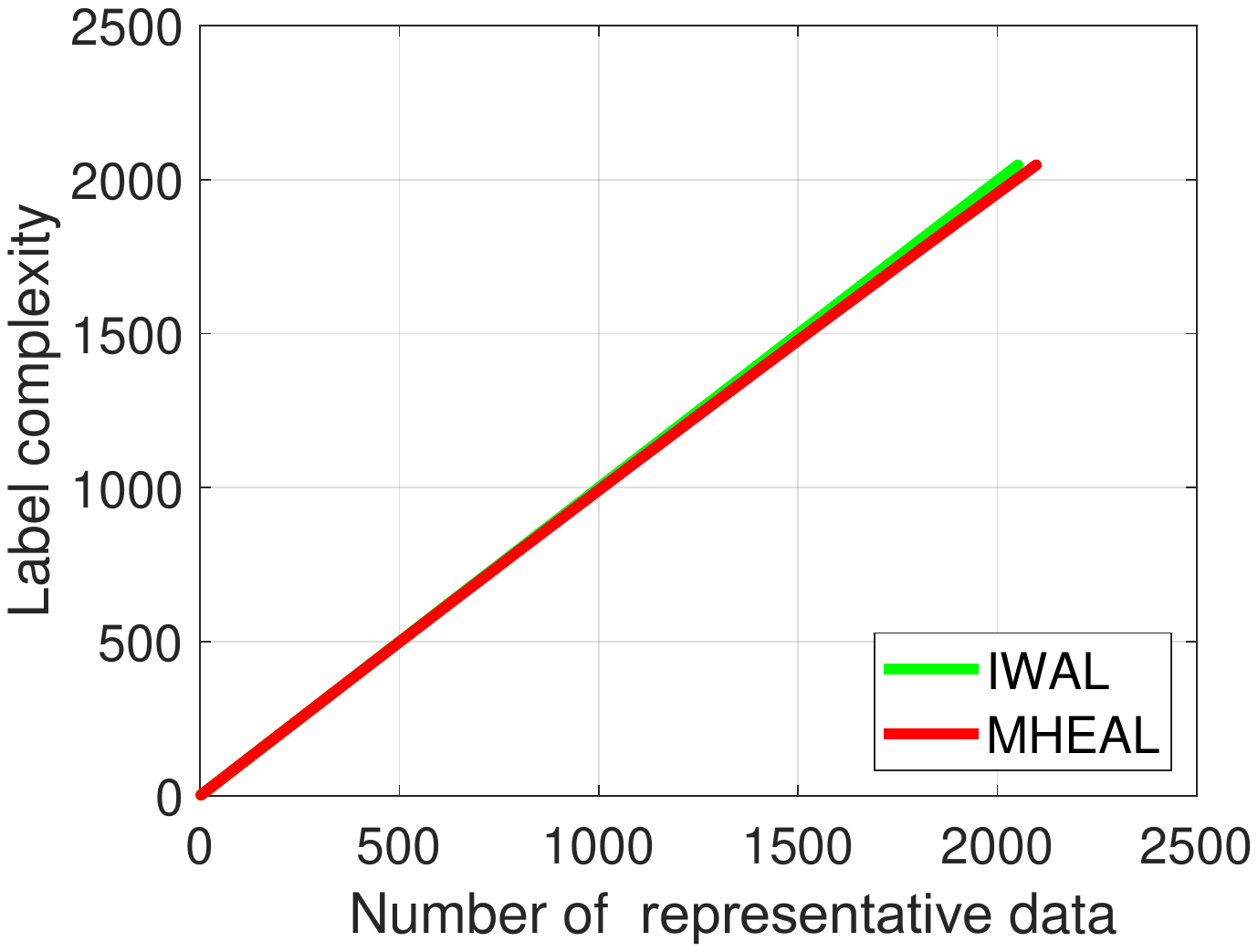}
\end{minipage}
} 
\subfloat[nomao]{
\begin{minipage}[t]{0.156\textwidth}
\centering
\includegraphics[width=1.17in,height=0.975in]{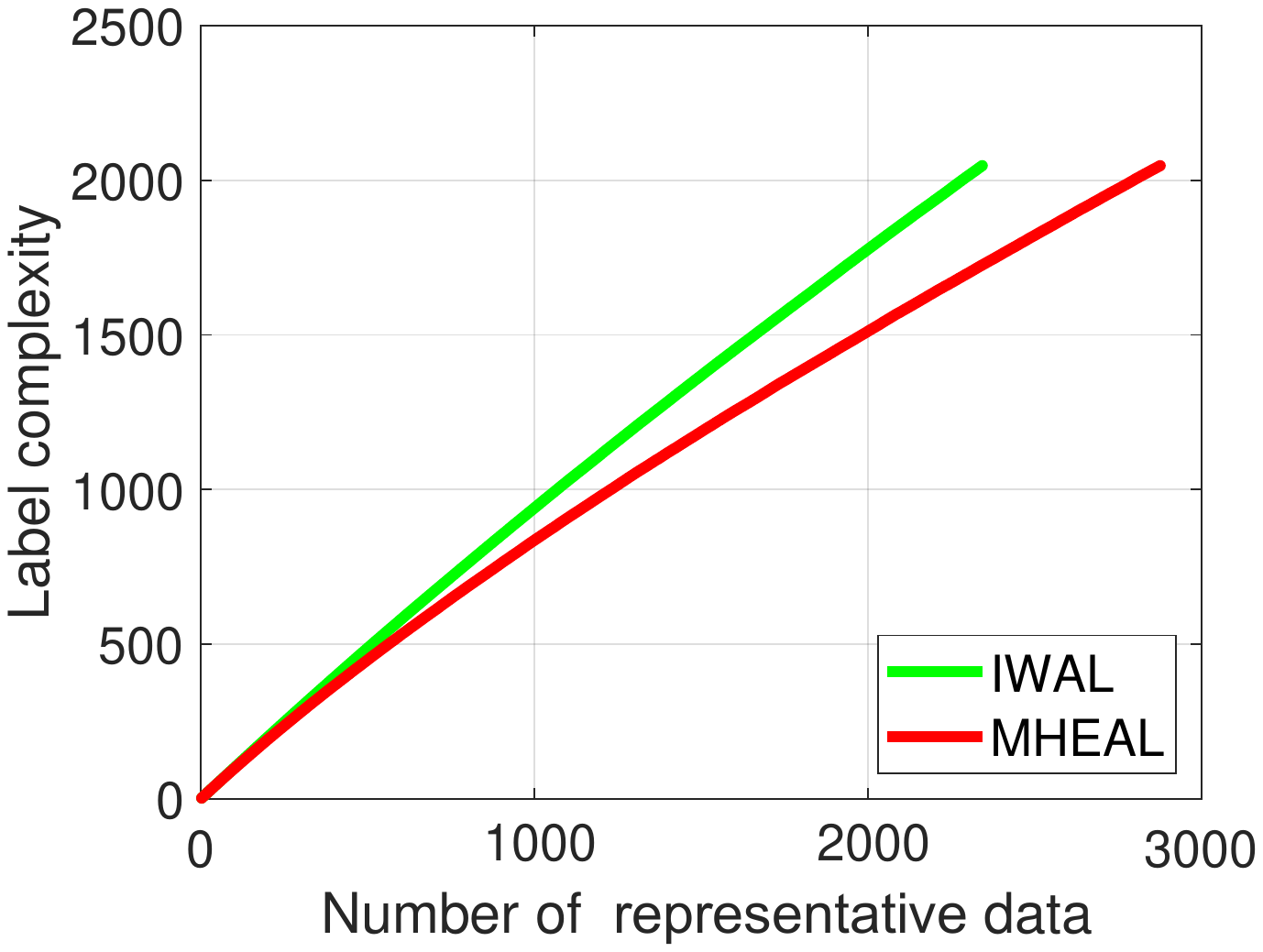}
\end{minipage}
} 
\caption{Label complexities of  IWAL and MHEAL. The results show that MHEAL can spend fewer labels to converge than IWAL.
 }  
 \label{Label_complexities}
\end{figure*}

 {\color{black}
 {\subsection{Empirical Study}}

\vspace{1.5mm}
\noindent\textbf{Datasets selection} Following Cortes's work in \cite{cortes2020region}, we select six UCI datasets for our empirical study, including   skin, shuttle, magic04, jm1, covtype, and nomao.

\vspace{1.5mm}
\noindent\textbf{Hypotheses generation}  We take logistic loss defined by $\log(1+\exp(-yh(x)))$ as the hypothesis prototype.  For $d$-dimensional data space, we randomly generate $10,000$ $d$-dimensional hyperplanes
with a prototype of $\sum_i^n w_ix_i+b=0$, where  any $w_i \in \mathcal{N}(0,1)$, $b$ also follows, and $h^{*}$ is defined as that  hyperplane which generates the minimum  loss.  Note that $ \mathcal{N}(0,1)$ denotes a standard Gaussian distribution.

 \vspace{1.5mm}
\noindent\textbf{Specification on $\theta_{\rm IWAL}$ and $\theta_{\rm MHEAL}$} With Theorem~1, we know the error disagreement of MHEAL  is a tighter polynomial expression of  IWAL. We thus need to  specify $ \theta_{\rm IWAL}$ and $\theta_{\rm MHEAL}$ which satisfy  $\theta_{\rm MHEAL}\leq \theta_{\rm IWAL}$.  Recalling Eqs.~(\ref{w_1_t}) and (\ref{degeneration_mhe}), we know 1) $\ell(h(x_t),\mathcal{Y})-\ell(h^*(x_t),\mathcal{Y})\leq 1$, and 2) $\ell(h(x_t),\mathcal{Y})-\ell(h^*(x_t),\mathcal{Y})$ of learning from $\mathcal{X}$ is tighter than that of learning from   $\mathcal{B}_i$ since $\mathcal{B}_i \subset  \mathcal{X}$. 
 Let ${\rm Vol}(\cdot)$ denote the volume function, we know ${\rm Vol}(\mathcal{B_i})={\rm Vol}(\mathcal{X})$. Let $\xi^{\rm IWAL}_t=\ell(h(x_t),\mathcal{Y})-\ell(h^*(x_t),\mathcal{Y})$ and Let $\xi^{\rm MHEAL}_t=\ell(h(x_t),\mathcal{Y})-\ell(h^*(x_t),\mathcal{Y})$, we know  $\xi^{\rm MHEAl}_t=\frac{1}{k}\xi^{\rm IWAL}_t$.  
 To enlarge the disagreement of  $\theta_{\rm IWAL}$ and $\theta_{\rm MHEAL}$, we force $\xi^{\rm IWAL}_t$ arrive at its upper bound, that is, $\xi^{\rm IWAL}_t=1$. Given a radius $r=0.1$, we then specify $\theta_{\rm IWAL}=10$.  Given that there are three clusters in $\mathcal{X}$, \ie, $k=3$, with the condition of  $\xi^{\rm MHEAL}_t=\frac{1}{k}\xi^{\rm IWAL}_t$, we then know $\xi^{\rm IWLA}_t=\frac{1}{3}\times 1=0.3333$. With a same setting on $r=0.1$, we then know $\theta_{\rm MHEAL}=3.333$.  
 
\vspace{1.5mm}
\noindent\textbf{Hypothesis-pruning} We  start the hypothesis-pruning using $\theta_{\rm IWAL}$ and $\theta_{\rm MHEAL}$, where the pruning  condition is that any hypothesis update needs to satisfy a minimal  disagreement. Following  \cite{cortes2020region}, we invoke a feasible pruning manner to prune those $10,000$ hyperplanes, which can guarantee a near-optimal convergence. The specified way is to iteratively  update $\theta_{\rm IWAL}$ and $\theta_{\rm MHEAL}$ using  $\sqrt{10/t}$ for IWAL and $\sqrt{3.333/t}$ for    MHEAL to eliminate more loose hypotheses.  The specified pruning operation is to eliminate the hyperplane which holds a logistic loss looser than $\sqrt{10/t}$ for IWAL or  $\sqrt{3.333/t}$ for MHEAL will be eliminated.    Figure~\ref{Hypothesis_pruning_speed} presents the comparison of MHEAL and IWAL on their hypothesis-pruning speeds. The results show that MHEAL can prune the hypothesis class faster than typical IWAL.
 
\vspace{1.5mm}
\noindent{\textbf{Generalization on error} Error disagreement of MHEAL  is a tighter polynomial expression of  IWAL, which  prunes the   hypothesis faster. We thus study its hypothesis-pruning speed, that is, whether MHEAL can eliminate more insignificant hypotheses using a given coefficient.  We thus follow the specification on $\theta_{\rm IWAL}$ and $\theta_{\rm MHEAL}$, and present the generalization error of pruning those $10,000$ hypotheses into their optimal hypothesis. The error curves are presented in Figure~\ref{Generalization_errors}. The results show that  MHEAL can obtain a tighter error than   IWAL. }

\vspace{1.5mm}
\noindent{\textbf{Generalization on  label complexity} Label complexity bound of MHEAL is also a tighter polynomial expression of  IWAL. We thus try to observe their label costs of converging into a desired hypothesis.  Figure~\ref{Label_complexities} presents the label complexities of IWAL and MHEAL.   The results show that MHEAL can spend fewer labels to converge than IWAL.}

 In conclusion, MHEAL employs a tighter error disagreement coefficient than IWAL to pure the hypothesis class, but results in a faster pruning speed. The generalization results on  error and label complexity then show  tighter bounds than that of IWAL, which verifies  our theoretical results in Theorem~1.  
}

\section{Experiments}
According to the statement of Section~5.1,
 MHEAL  algorithm  adopts hyperspherical clustering to learn  the  representative data   in each pre-estimated  cluster. The experiments   discuss 
  the following five questions.
 
\begin{itemize}
\item  Why SphericalKmeans clustering is applied? How about other  spherical  clustering approaches, e.g. K-means and Gaussian mixture model  (GMM)?

\item Why do we follow $\ell_0$ expression of MHE?  How about  $\ell_1$  and $\ell_2$  expressions of MHE?

\item Why decision boundaries can characterize better  representative data? What is the difference of sampling in  out-version-space and in-version-space?

\item Can MHEAL derive more expressive  representative data than  the state-of-the-art deep AL baselines?

 \item  {Can MHEAL  keep solid performance against special  data-efficient learning settings, such as repeated and noisy scenario?}
\end{itemize}

\subsection{Data-Efficient Spherical Clustering}
\begin{figure}
 \centering
\includegraphics[scale=0.32]{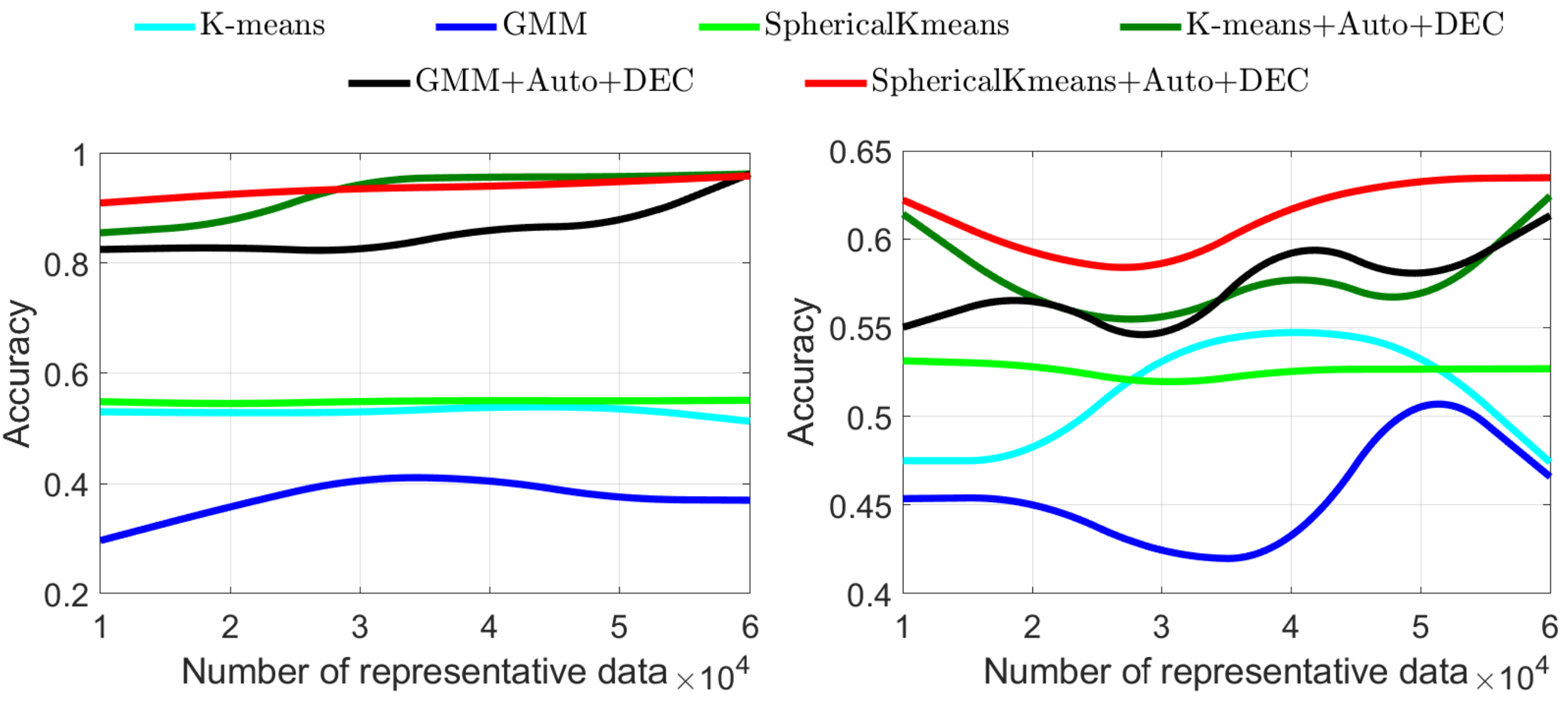}
\caption{Data-efficient spherical clustering using representative data. To  enhance the unsupervised learning performance of K-means, GMM, SphericalKmeans, we adopt the AutoEncoder and 
Kullback–Leibler (KL) divergence   loss of deep embedding clustering (DEC) which performs end-to-end learning of convnets. }  
\label{clustering}
\end{figure}

Clustering model on  representative data can be transferred into large-scale learning \cite{dai2008self}. In this section,
we focus on  spherical clustering on  representative data, and the related baselines are  K-means, GMM, and SphericalKmeans. Figure~\ref{clustering} presents the performance of clustering on  representative data  of the three baselines on  
  MNIST and Fashion-MNIST, where the  representative data are randomly sampled from the original training sets of the datasets, with a  varying number  from 1,000 to 6,0000.  
The maximum iterations of these baselines are set as 60.

To improve the non-deep clustering algorithms, we adopt the AutoEncoder and 
Kullback–Leibler (KL) divergence   loss of deep embedding clustering (DEC) \cite{xie2016unsupervised} to enhance their unsupervised learning results. The dimension  of the input layer is 784,   the encoding layers follow the (input,output) dimension settings of (784,500),  (500,500), (500,2000),  (2000,10), then decoding layers follow 
the (input,output) dimension settings of (10,2000),  (2000,500), (500,500),  (500,784). We  also build one clustering layer, which converts the input features to   label probability of clusters calculated by student’s t-distribution, measuring the similarity between   embedded data and   centroids.

The deep clustering results  also are presented in Figure~\ref{clustering}.  It is clear that the typical clustering baselines have significant accuracy improvements by adopting AutoEncoder  and DEC, where SphericalKmeans performs best most of the time whether in non-deep or deep clustering. 
 To analyze their perturbations to the number of  representative data, Table~\ref{table_Accuracy_statistics} presents the unsupervised  accuracy  statistics for Figure~\ref{clustering}, where optimal accuracy denotes the best accuracy with  a given number of  representative data, varying from \{10,000,20,000,...,60,000\}, and  the best `optimal accuracy' and the lowest `standard deviation' are marked in bold. From those metrics, we   find that SphericalKmeans  can achieve better clustering performance on  representative data than other spherical clustering approaches, characterizing better geometric clustering features.

\begin{table*}
 \caption{Accuracy statistics of  breakpoints of the learning curves of  clustering on  representative data. }
 \setlength{\tabcolsep}{4.5pt}{
\begin{center}
\scalebox{1.29}{
 \begin{tabular}{c|c  c c   |c  c c c    } 
 \hline
\multirow{2}{*}{Algorithms}  &    \multicolumn{3}{c|}{MNIST}  &    \multicolumn{3}{c}{Fashion-MNIST} \\
    &  Optimal Acc   &   Mean Acc   &      Std   &  Optimal Acc    &   Mean Acc   &      Std    \\

\hline
 {K-means} &    (4,0000,  0.53961)   & 0.5298  &  0.0095  & (40,000, 0.55031)  & 0.5083 &   0.0372 \\

{GMM}    & (30000, 0.41239)   & 0.3695  &  0.0420  & (50,000,  0.52611) & 0.4566 &  0.0391\\
{SphericalKmeans}      &  (6,0000, 0.55188)   & 0.5490  &  \textbf{0.0025}      &(10,000,  0.53125) & 0.5262 &   \textbf{0.0050}  \\

{K-means+Auto+DEC}     &(60000, 0.96221)  &0.9257  &  0.0490     & (60,000, 0.62419) & 0.5821 &   0.0309   \\

{GMM+Auto+DEC}      &(6,0000, 0.96256)   &0.8615  &  0.0536     & (60,000, 0.61319) & 0.5743 &  0.0307 \\

{SphericalKmeans+Auto+DEC}     &\textbf{(60,000, 0.96864)}   &\textbf{0.9362}  &  0.0172     &  \textbf{(60,000, 0.63448)}& \textbf{0.6133} & 0.0230   \\
\hline
\end{tabular}}
\end{center}}
\label{table_Accuracy_statistics}
\end{table*}

\subsection{Data-Efficient Distribution Matching}
\begin{figure*} 
\subfloat[{$\ell_{\rm KL}$ on MNIST}]{
\begin{minipage}[t]{0.156\textwidth}
\centering
\includegraphics[width=1.17in,height=0.975in]{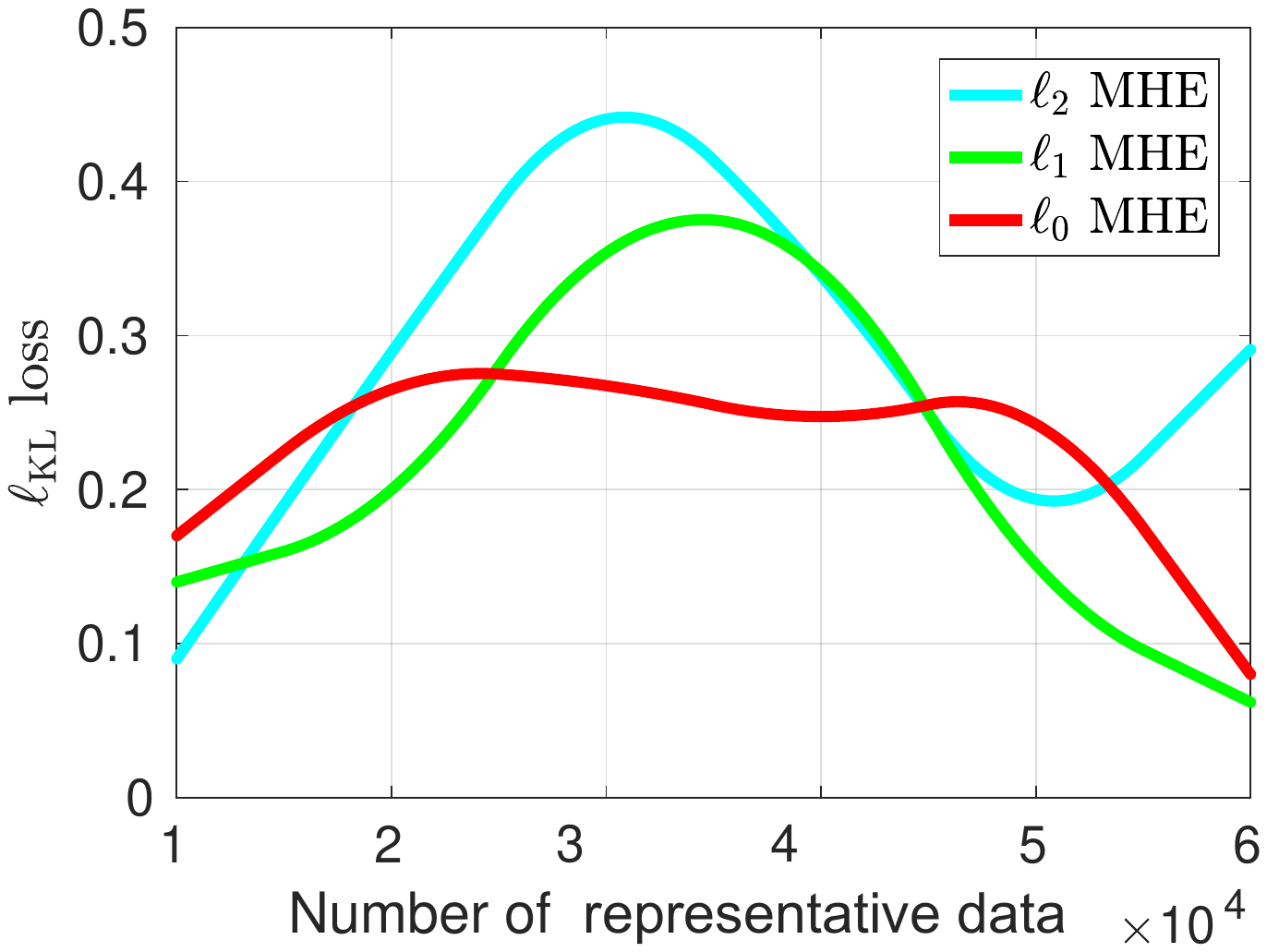}
\end{minipage}
}
\subfloat[{$\ell_{\rm KL}$ on F-MNIST}]{
\begin{minipage}[t]{0.156\textwidth}
\centering
\includegraphics[width=1.17in,height=0.975in]{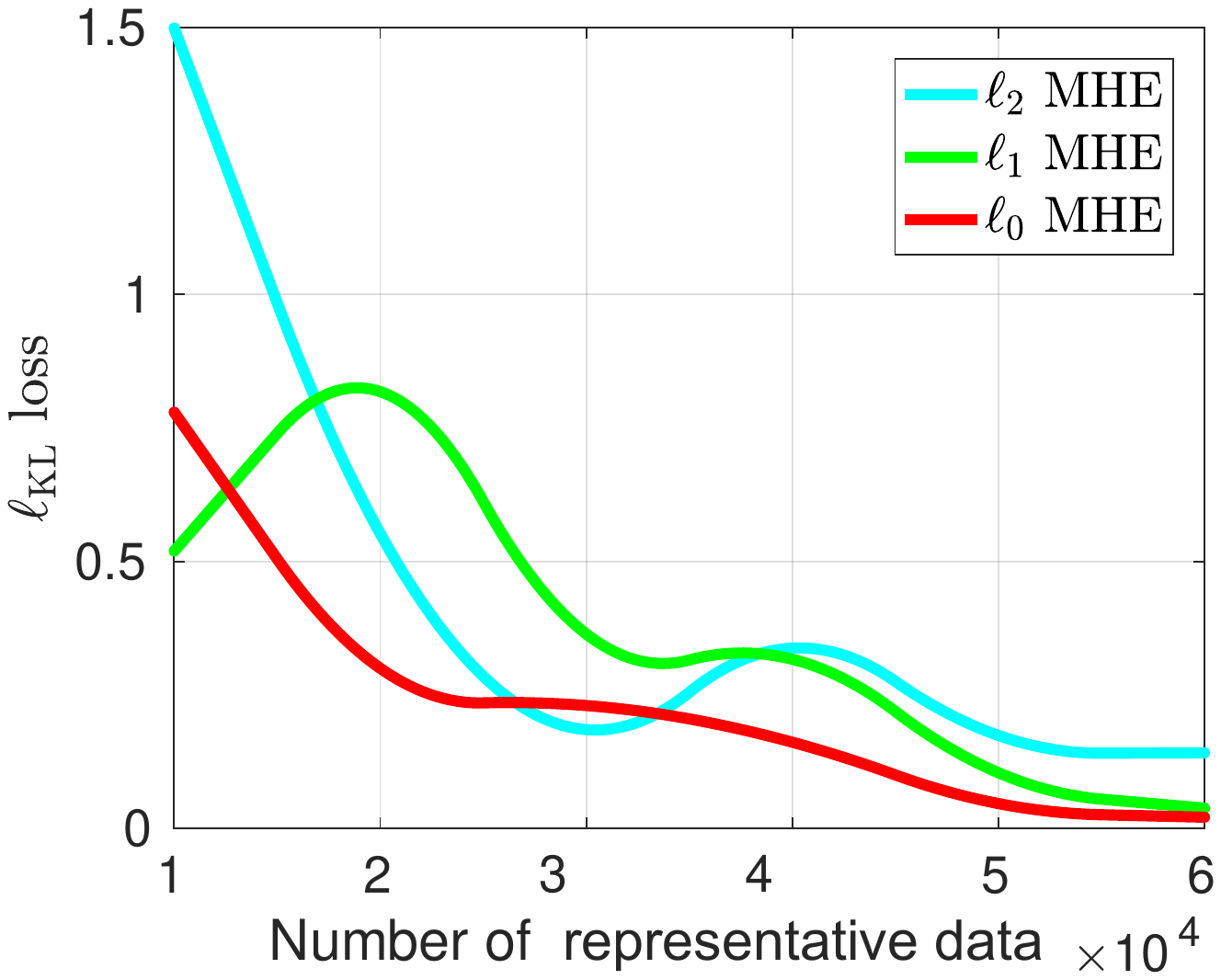}
\end{minipage}
}
\subfloat[{$\ell_{\rm MMD}$  on MNIST}]{
\begin{minipage}[t]{0.156\textwidth}
\centering
\includegraphics[width=1.17in,height=0.975in]{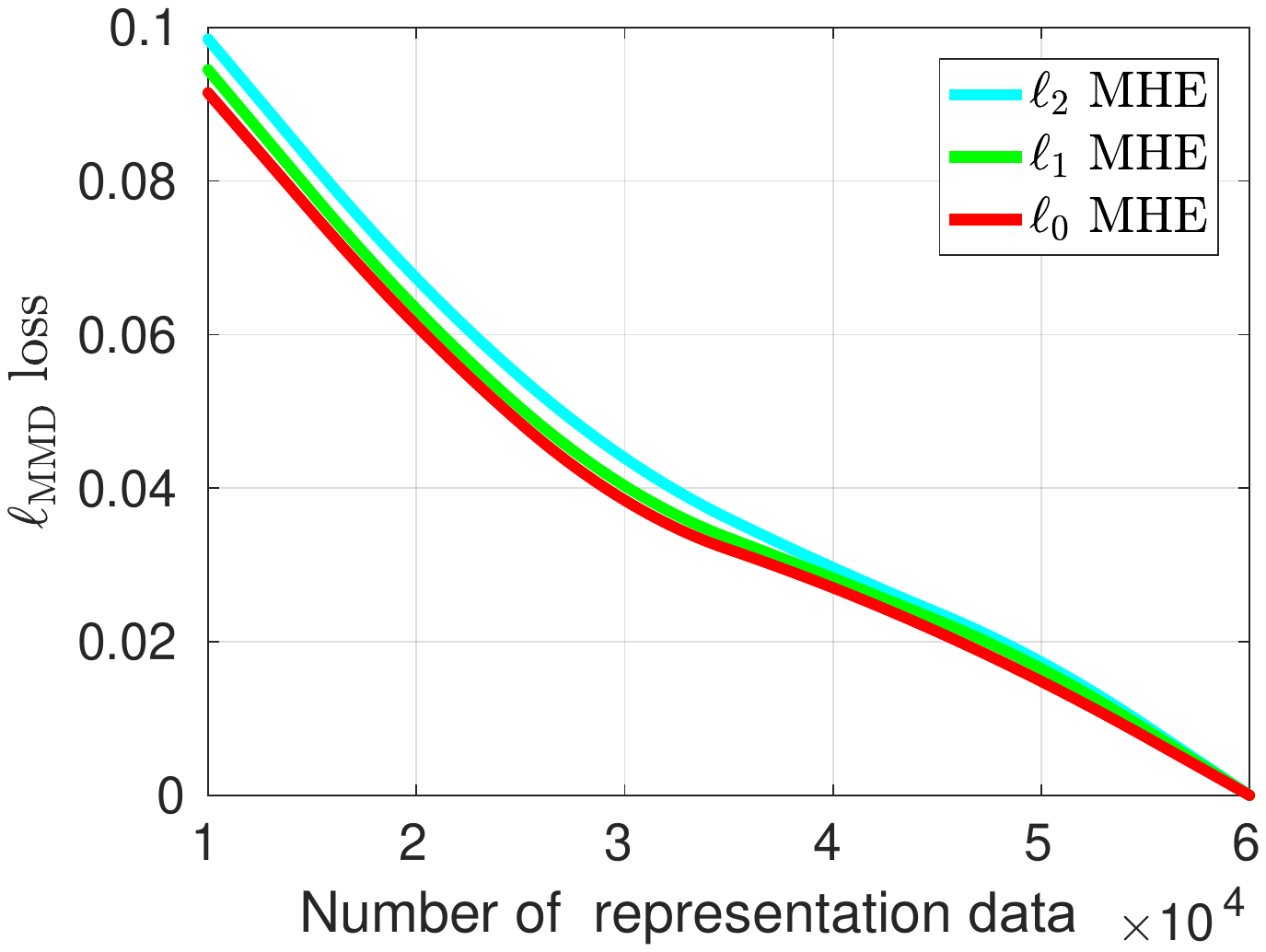}
\end{minipage}
} 
\subfloat[{$\ell_{\rm MMD}$  on F-MNIST}]{
\begin{minipage}[t]{0.156\textwidth}
\centering
\includegraphics[width=1.17in,height=0.975in]{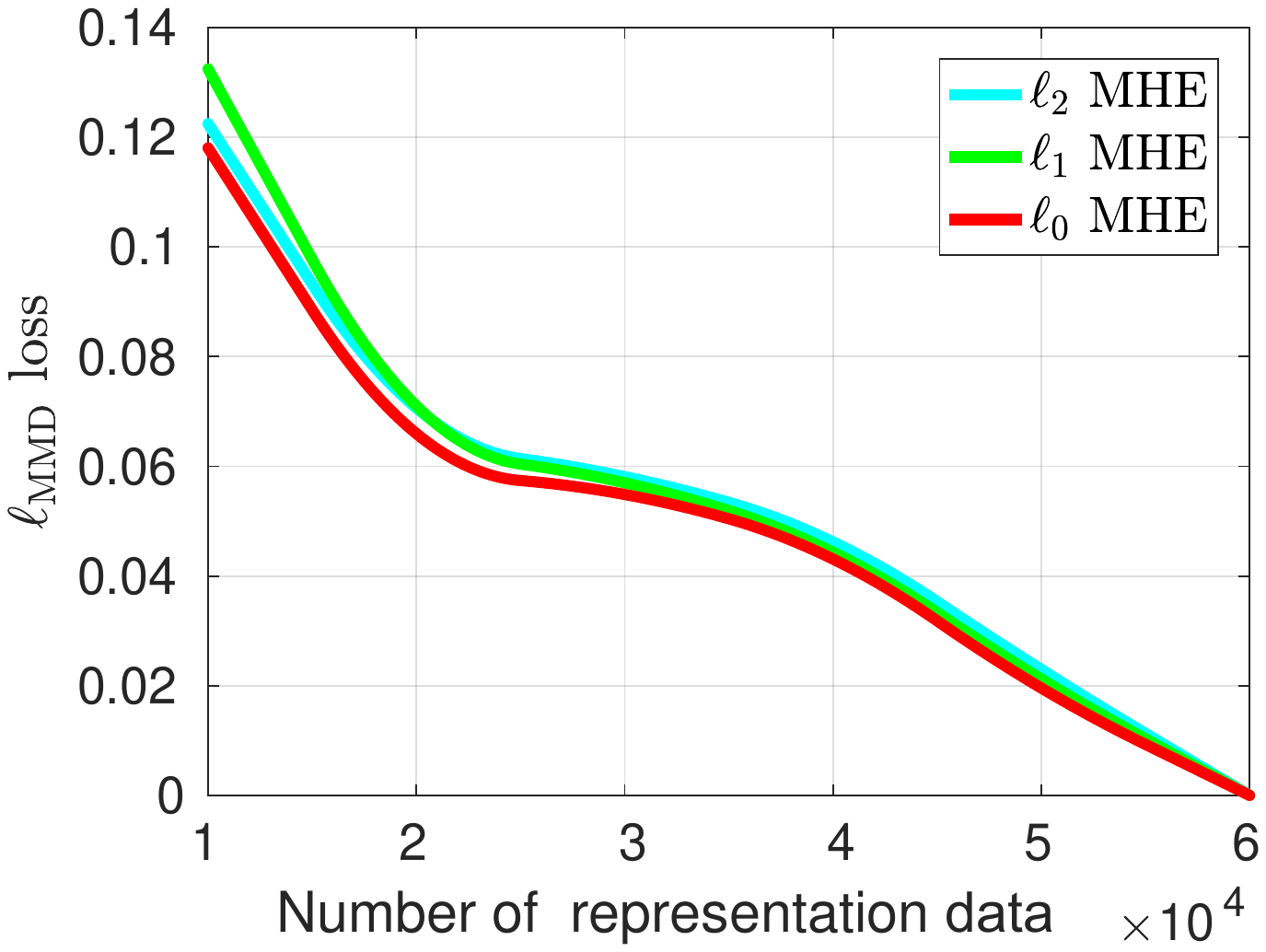}
\end{minipage}
}
\subfloat[{${\rm MMD}_{\mu}$ on MNIST}]{
\begin{minipage}[t]{0.156\textwidth}
\centering
\includegraphics[width=1.17in,height=0.975in]{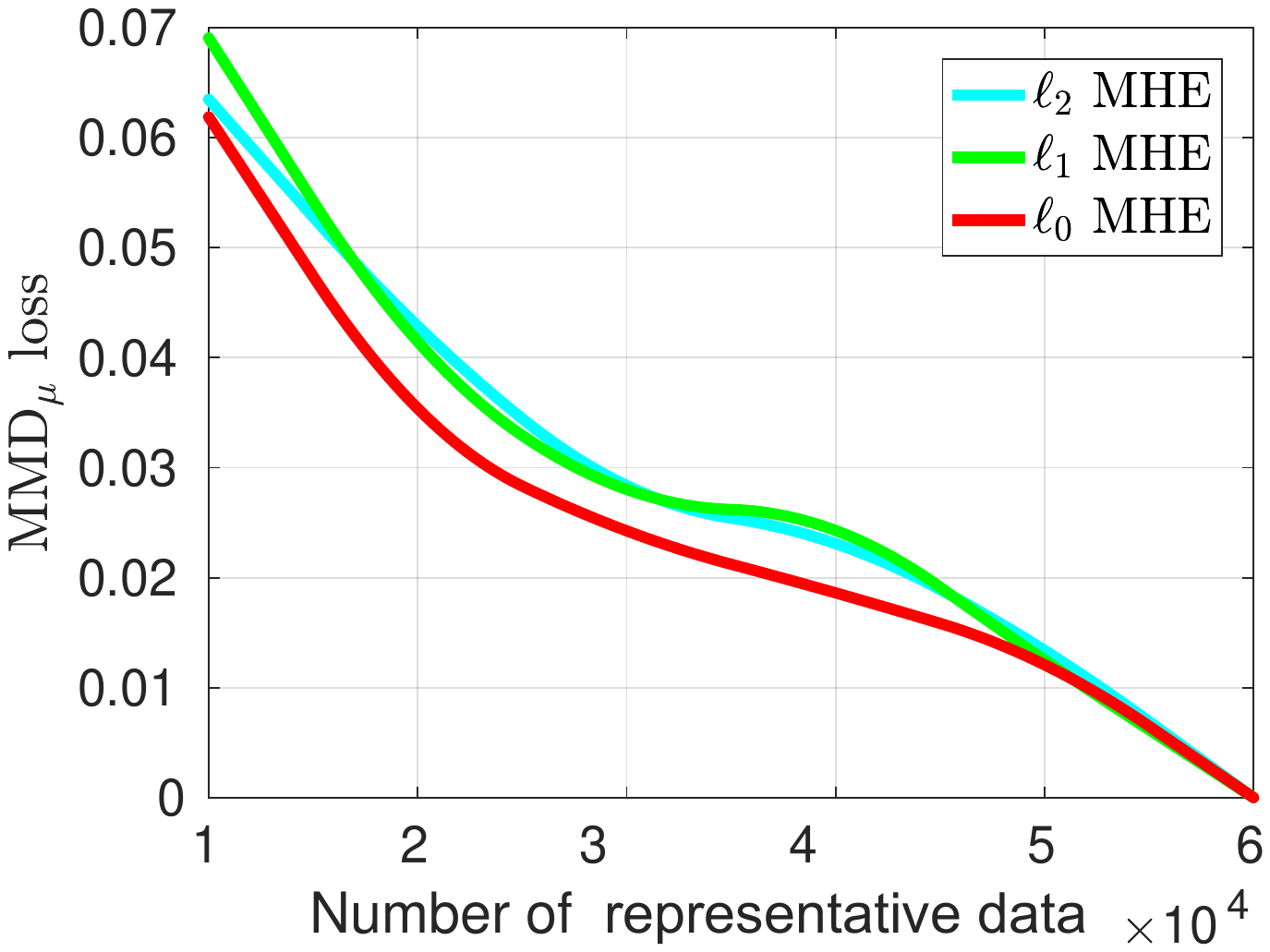}
\end{minipage}
}
\subfloat[{${\rm MMD}_{\mu}$ on F-MNIST}]{
\begin{minipage}[t]{0.156\textwidth}
\centering
\includegraphics[width=1.17in,height=0.975in]{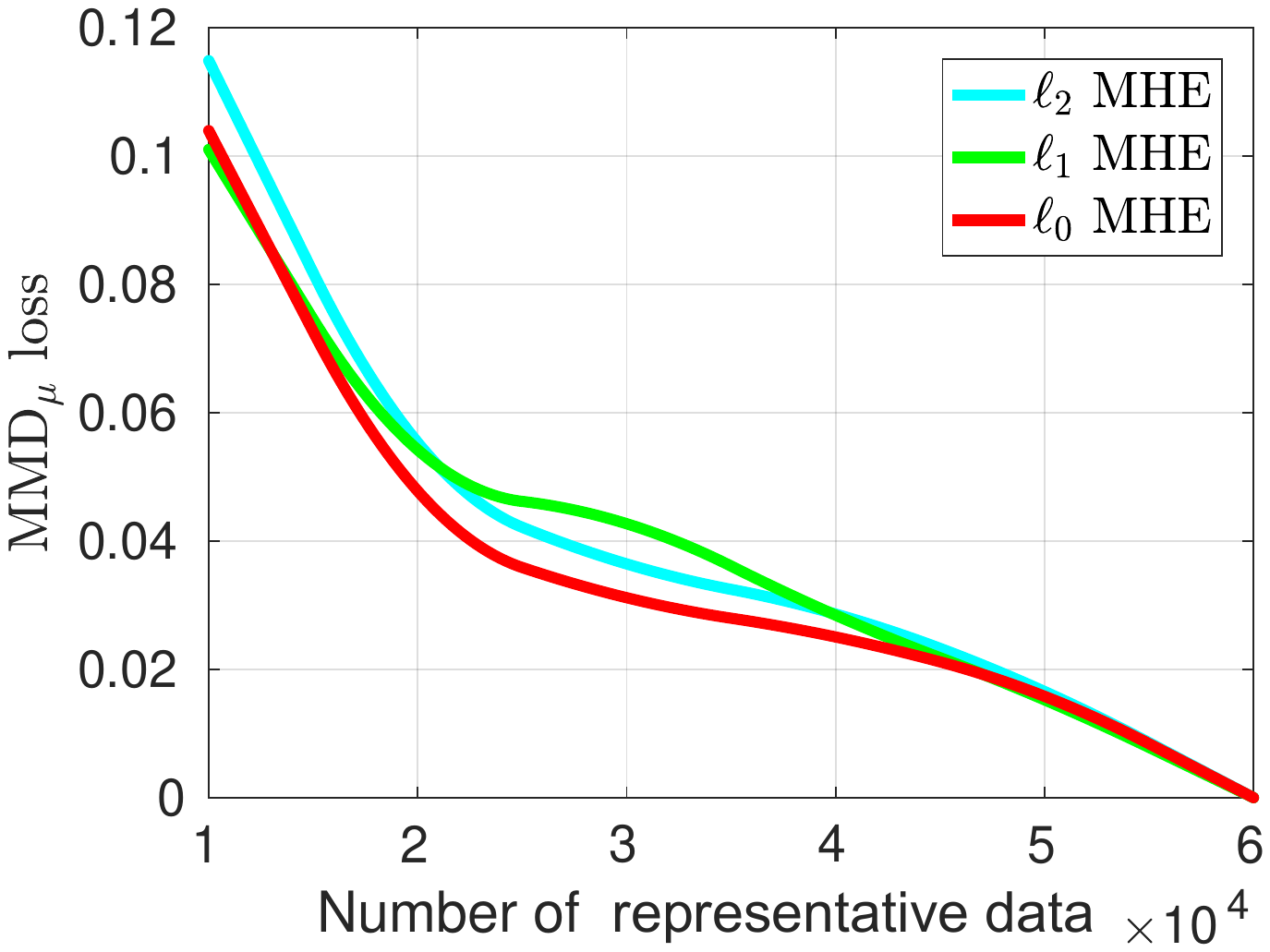}
\end{minipage}
} 
\caption{$\ell_{\rm KL}$, $\ell_{\rm MMD}$, and ${\rm MMD}_{\mu}$  loss curves of distribution matching on  representative data via  MHE. The curves show that distribution matching by $\ell_0$ MHE  approximates the input distribution  more tightly than $\ell_1$ and $\ell_2$  MHE.  (F-MNIST denotes Fashion-MNIST )
 }  
\label{KL_and_MMD}
\end{figure*}

Distribution matching \cite{bickel2008transfer} on  representative data presents expressive modeling for input features. We compare the performance disagreements of distribution matching using  $\ell_0$, $\ell_1$, and $\ell_2$ expressions of MHE. 
Specifically,  $\ell_0$ MHE minimizes  $\mathbb{E}_{0,d}$ by adopting our max-min solution of Eq.~(\ref{se_max_min}), $\ell_1$ MHE  minimizes   $\mathbb{E}_{1,d}$ by adopting a heuristic search with a random beginning, and $\ell_2$ MHE  minimizes  $\mathbb{E}_{2,d}$ by adopting a gradient solver of Eq.~(3) with a learning rate 0.001. We select KL divergence and Maximum Mean Discrepancy (MMD)  \cite{gretton2012kernel} as the loss metrics of   learning by  representative data via  MHE. 

Given $\mathcal{X}'$ be the  representative data  of $\mathcal{X}$ with a size  $m$, we define $\ell_{\rm KL}:=\sum_i \mathcal{X}(i){\rm log}\frac{ \mathcal{X}(i)}{\mathcal{X}'(i)+\beta},$ and   
$\ell_{\rm MMD}:=\Big\| \frac{1}{n^2}\sum_{i,j}  \|\mathcal{X}(i)-\mathcal{X}(j)\|-\frac{2}{nm}\sum_{i,j}  \|\mathcal{X}(i)-\mathcal{X}'(j)\| 
+\frac{1}{m^2}\sum_{i,j}  \|\mathcal{X}'(i)-\mathcal{X}'(j)\|\Big\|_1^{1/2},$
where 
$\beta$ avoids the calculation infeasible of $\mathcal{X}'(i)=0$, and the $\ell_1$ operation of  $\ell_{\rm MMD}$ avoids the negative values of the $\ell_2$ metrics on MMD.

\begin{table*}
 \setlength{\tabcolsep}{4.5pt}{
\begin{center}
\setlength\tabcolsep{8pt}
\caption{Loss statistics of the breakpoints of  distribution matching on  representative data via   MHE. } 
\scalebox{1.29}{
 \begin{tabular}{c|c|c c c |c  c cc } 
\hline
\multirow{2}{*}{Loss Metrics} & \multirow{2}{*}{Algorithms} &  \multicolumn{3}{c|}{MNIST}& \multicolumn{3}{c}{Fashion-MNIST} \\  
  &         &  Initial  Loss   &   Mean Loss  &      Std  &  Initial  Loss   &   Mean Loss  &      Std     \\
\hline
\multirow{3}{*}{$\ell_{\rm KL}$} &  $\ell_2$ MHE &0.1723 & 0.2751&0.1396&1.5013&
    0.4603&0.5325 \\
 &  $\ell_1$ MHE &0.1456 &  \textbf{0.2103}&0.1331&0.5194& 0.3698&0.3409 \\
 &  $\ell_0$ MHE &\textbf{0.0912} &0.2183&\textbf{0.0788}&0.7869& 0.2451
& 0.2788 \\
\hline
\multirow{3}{*}{ $\ell_{\rm MMD}$} &  $\ell_2$ MHE &0.0985&0.0402& 0.0354&0.1225&0.0527&
    0.0418 \\
 &  $\ell_1$ MHE &0.0945 & 0.0338& {0.0385} & 0.1325  &0.0533& 0.0455 \\
 &  $\ell_0$ MHE &\textbf{0.0915} &{0.0385}&\textbf{0.0330}&\textbf{0.1180}& \textbf{0.0494}& \textbf{0.0406} \\
\hline
\multirow{3}{*}{$ {\rm MMD}_{\mu}$} &  $\ell_2$ MHE &0.0634 &0.0284& 0.0222&0.1148&  0.0410&\textbf{0.0348} \\
 &  $\ell_1$ MHE &0.0690 &0.0290&0.0238&0.1039 & 0.0395& {0.0351} \\
 &  $\ell_0$ MHE &\textbf{0.0618} & \textbf{0.0249}&\textbf{0.0212}&\textbf{0.1010}  &\textbf{0.0364}&  0.0359\\
\hline
\end{tabular}}
\end{center}}
\label{Table_Loss_statistics}
\end{table*}

Note $\|\cdot\|_1$ denotes the $\ell^1$-norm.
Since kernel MMD may result in high computational cost and has parameter perturbations,   we select the unbiased $\ell_2$ operator to define MMD from a vector level.  To approximate a nearly zero loss for $\ell_{\rm KL}$ due to its biased estimations, we set $\beta=0.552$ for MNIST and  $\beta=0.352$ for Fashion-MNIST. 
Before estimating $\ell_{\rm KL}$ and $\ell_{\rm MMD}$, each feature of the dataset is divided by 255 which scales into [0,1]. Besides those two metrics, we also present  the mean value disagreement as an overall MMD observation to verify the above two metrics:   ${\rm MMD}_{\mu}=\|\mu_\mathcal{X}-  \mu_{\mathcal{X}'}\|$, where $\mu_X$ denotes the mean vector of $X$. Figures~\ref{KL_and_MMD}(a)-(d) present the loss curves of $\ell_{\rm KL}$ and $\ell_{\rm MMD}$. These curves show that distribution matching by $\ell_0$ MMD  approximates the input distribution  more tightly than $\ell_1$ and $\ell_2$  MMD.

Table~1 then presents the loss statistics of the breakpoints (e.g., \{10,000, 20,000, 30,000, 40,000, 50,000, 60,000\}) of Figure~\ref{KL_and_MMD}, where ``Acc'' denotes Accuracy and ``Std'' denotes  Standard Deviation. 
The results show that $\ell_0$ MMD achieve lower initial and mean distribution loss than $\ell_1$ and $\ell_2$  MMD. To verify the effectiveness of   $\ell_{\rm KL}$ and $\ell_{\rm MMD}$ in Figure~10, we present an overall observation by estimating   ${\rm MMD}_{\mu}$ in Figures~\ref{KL_and_MMD}(e)-(f). The results keep consistent properties  as with Figures~\ref{KL_and_MMD}(a)-10(d), that is, $\ell_0$ MHE adopting Eq.~(7) usually results in   lower distribution losses than  its $\ell_1$ and $\ell_2$ expressions.

\vspace{1.5mm}
\noindent\subsection{Data-Efficient   Version Space  Representation }  
Decision boundaries \cite{beygelzimer2010agnostic} are distributed in the version space over the tube manifold of   hyperspherical distribution. Effective  representative data are derived from decision boundaries. This section thus revels the effectiveness of MHE in characterizing the topology of  decision boundaries.  Following the cluster boundary detection \cite{cao2018multidimensional} and out-of-distribution detection \cite{liang2018enhancing},  we also use a hyper-parameterized  threshold     to divide one cluster into two parts: 1) in-version-space i.e. tube manifold $\mathcal{M}$,  and 2) out-version-space i.e. inner regions of $B(c,R-\gamma)$. To specify $\mathcal{M}$ or  $B(c,R-\gamma)$, we use hyperspherical  energy as the  parameterized  variable.  

\textbf{Specification of Definition~1.} Given $K$ samples around  $x_i$, $\mathbb{E}_{0,d}(x_i)$ denotes the hyperspherical  energy of $B(c,r)$. Assume that there is a threshold $g'$ which divides $B(c,r)$ into  $\mathcal{M}$ and $B(c,R-\gamma)$: if $\mathbb{E}_{0,d}(x_i)>g'$, $x_i\in \mathcal{M}$, else $x_i\in B(c,R-\gamma)$. Specifically, we use PCA to project 2  dimensions  of MNIST and Fashion-MNIST as its extracted features.
To free $g'$,  we  set $K$=5   and collect the top 30\% training data with large $\mathbb{E}_{0,d}$    to specify $\mathcal{M}$.  

 \begin{figure}
\subfloat[{MNIST}]{
\begin{minipage}[t]{0.24\textwidth}
\centering
\includegraphics[width=1.7in,height=1.42in]{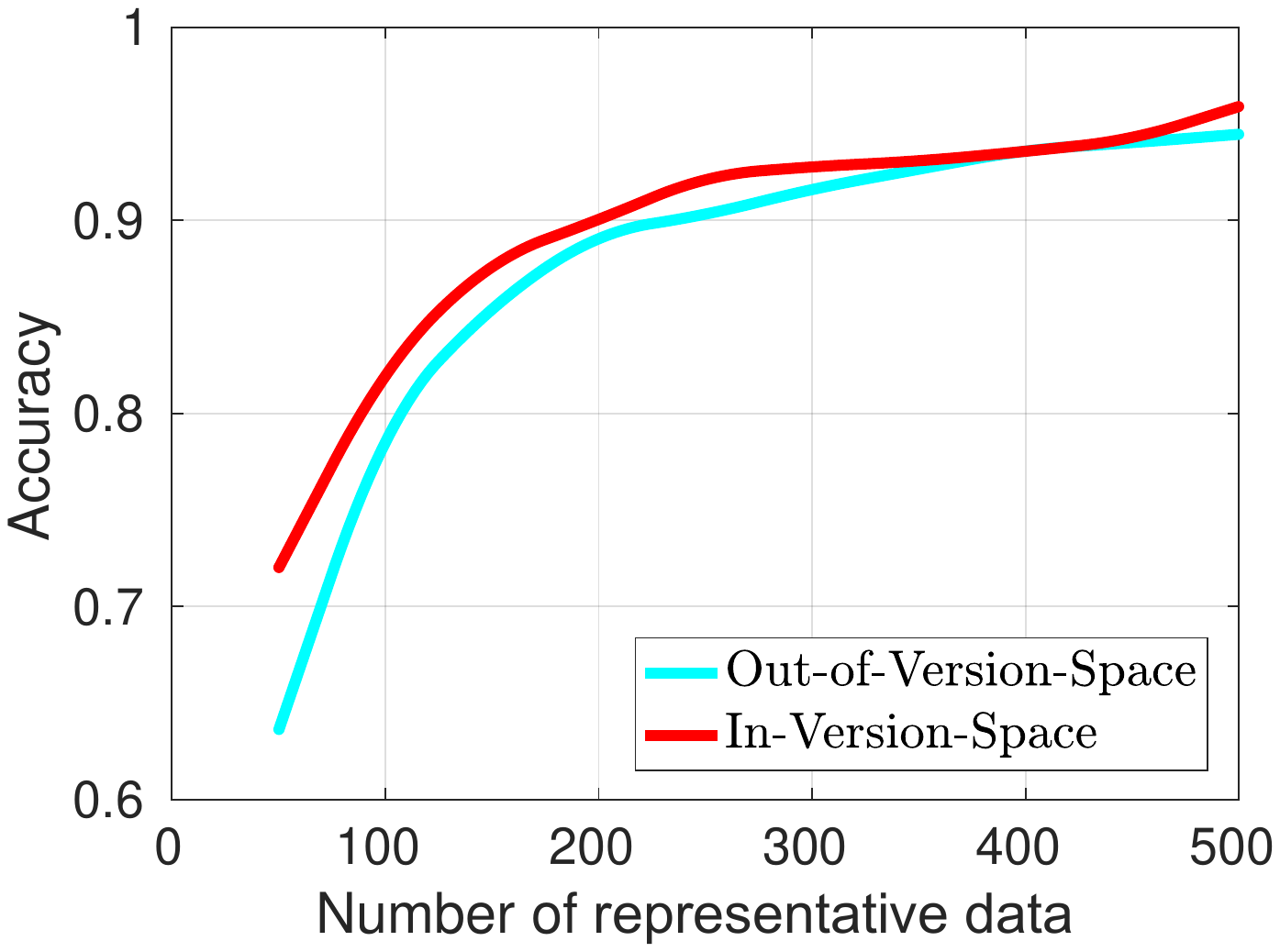}
\end{minipage}
}
\subfloat[{Fashion-MNIST}]{
\begin{minipage}[t]{0.24\textwidth}
\centering
\includegraphics[width=1.7in,height=1.42in]{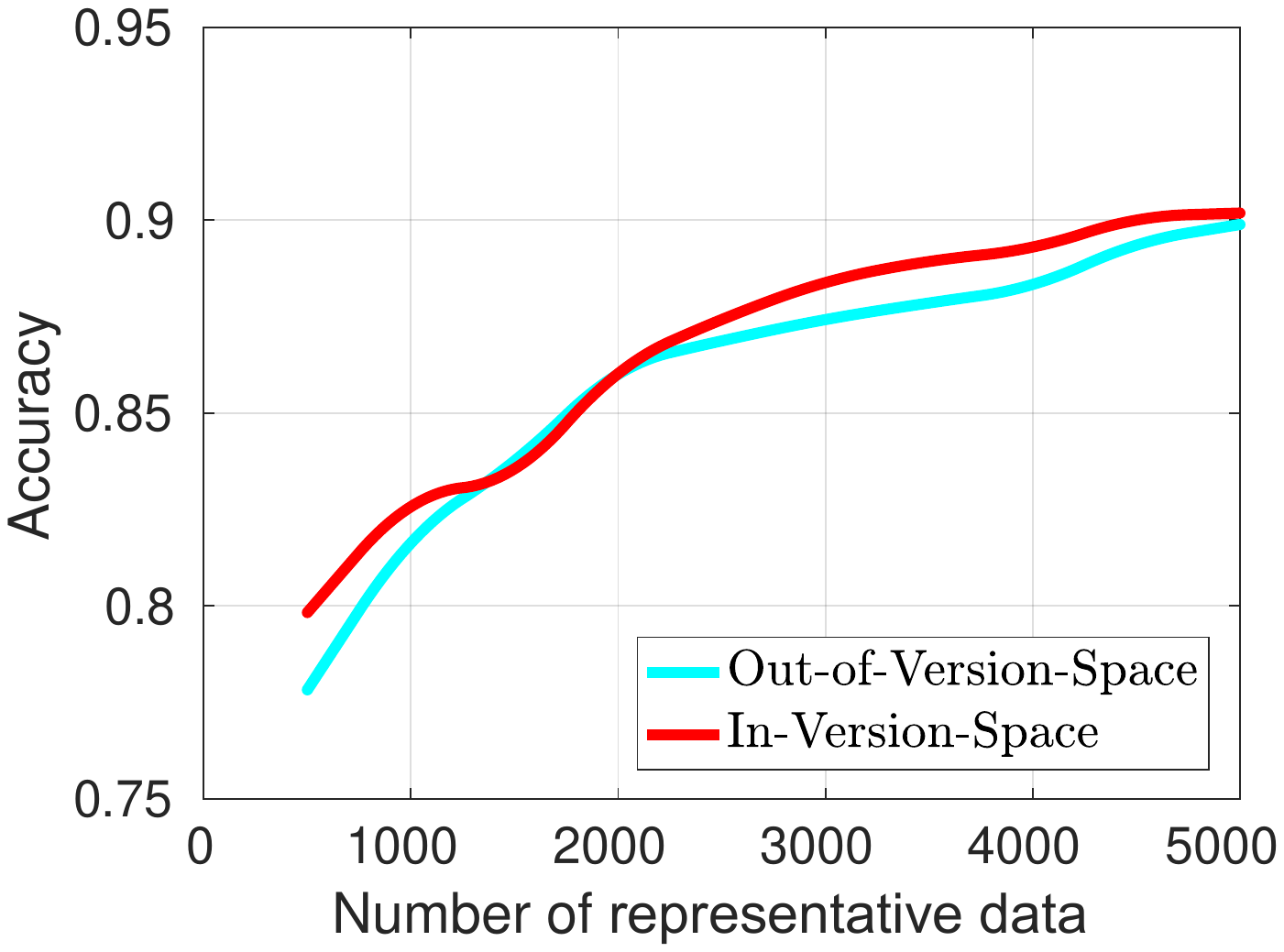}
\end{minipage}
}
\caption{Learning disagreements of random sampling in  out-of-version-space and in-version-space. The curves show that sampling  representative data in in-version-space has significant superiority than that of sampling in out-of-version-space. 
 }  
\label{out-of-version-space}
\end{figure}

We next randomly sample representative data from  $\mathcal{M}$ and $B(c,R-\gamma)$ to compare their learning disagreements. Figure~\ref{out-of-version-space} presents the average classification accuracy results with 10 times of random sampling, where the used prediction model is a CNN following  \cite{gal2017deep}.  The learning curves clearly show that sampling  representative data in in-version-space has significant superiority than that of sampling in out-of-version-space. This explains our motivation of sampling  representative data from in-versions-space regions, covering decision boundaries of $\gamma$-tube.

\begin{figure*} 
\subfloat[Phishing]{
\begin{minipage}[t]{0.24\textwidth}
\centering
\includegraphics[width=1.8in,height=1.52in]{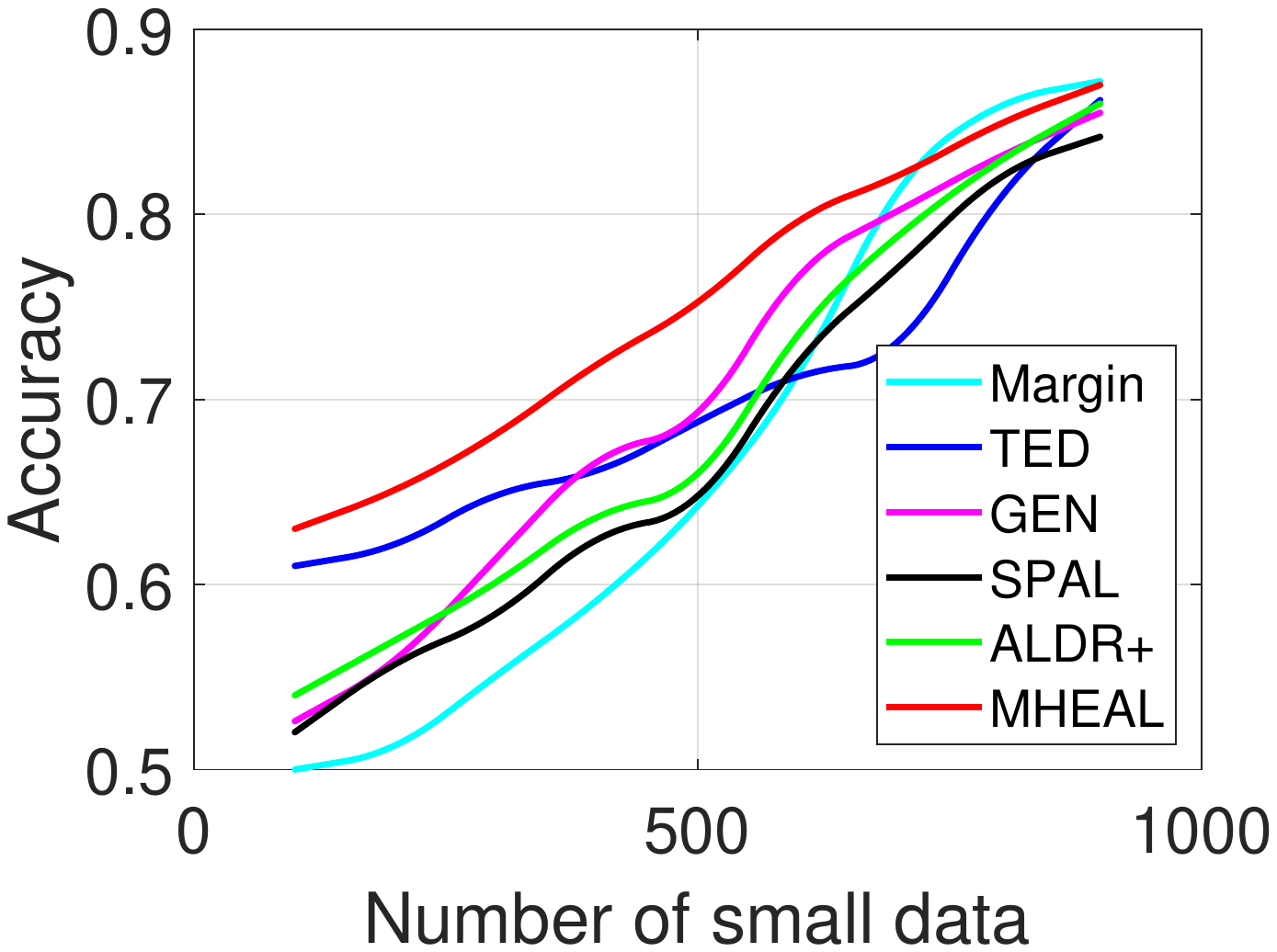}
\end{minipage}
}
\subfloat[Satimage]{
\begin{minipage}[t]{0.24\textwidth}
\centering
\includegraphics[width=1.8in,height=1.52in]{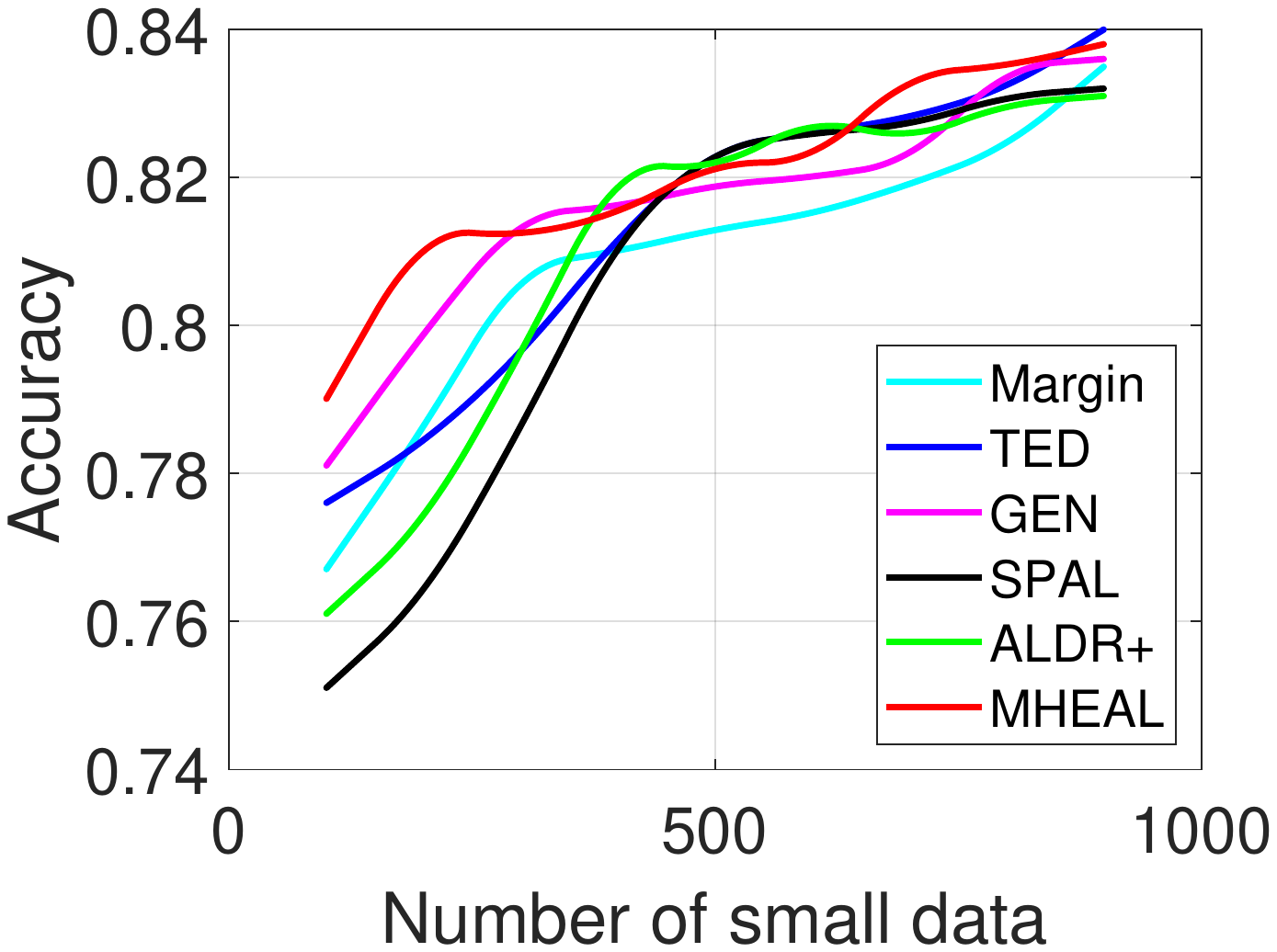}
\end{minipage}
} 
\subfloat[Adult]{
\begin{minipage}[t]{0.24\textwidth}
\centering
\includegraphics[width=1.8in,height=1.52in]{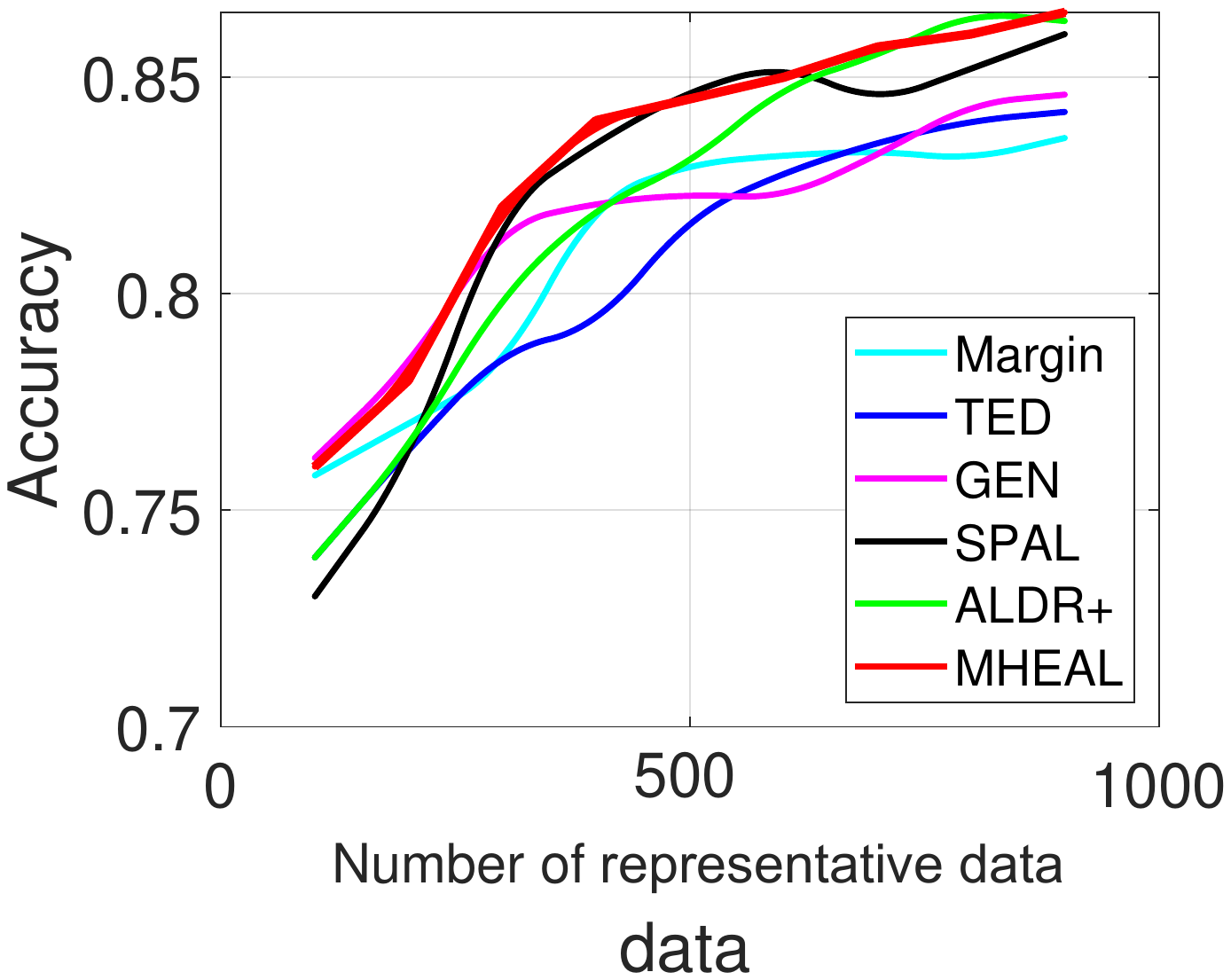}
\end{minipage}
} 
\subfloat[MNIST]{
\begin{minipage}[t]{0.24\textwidth}
\centering
\includegraphics[width=1.8in,height=1.52in]{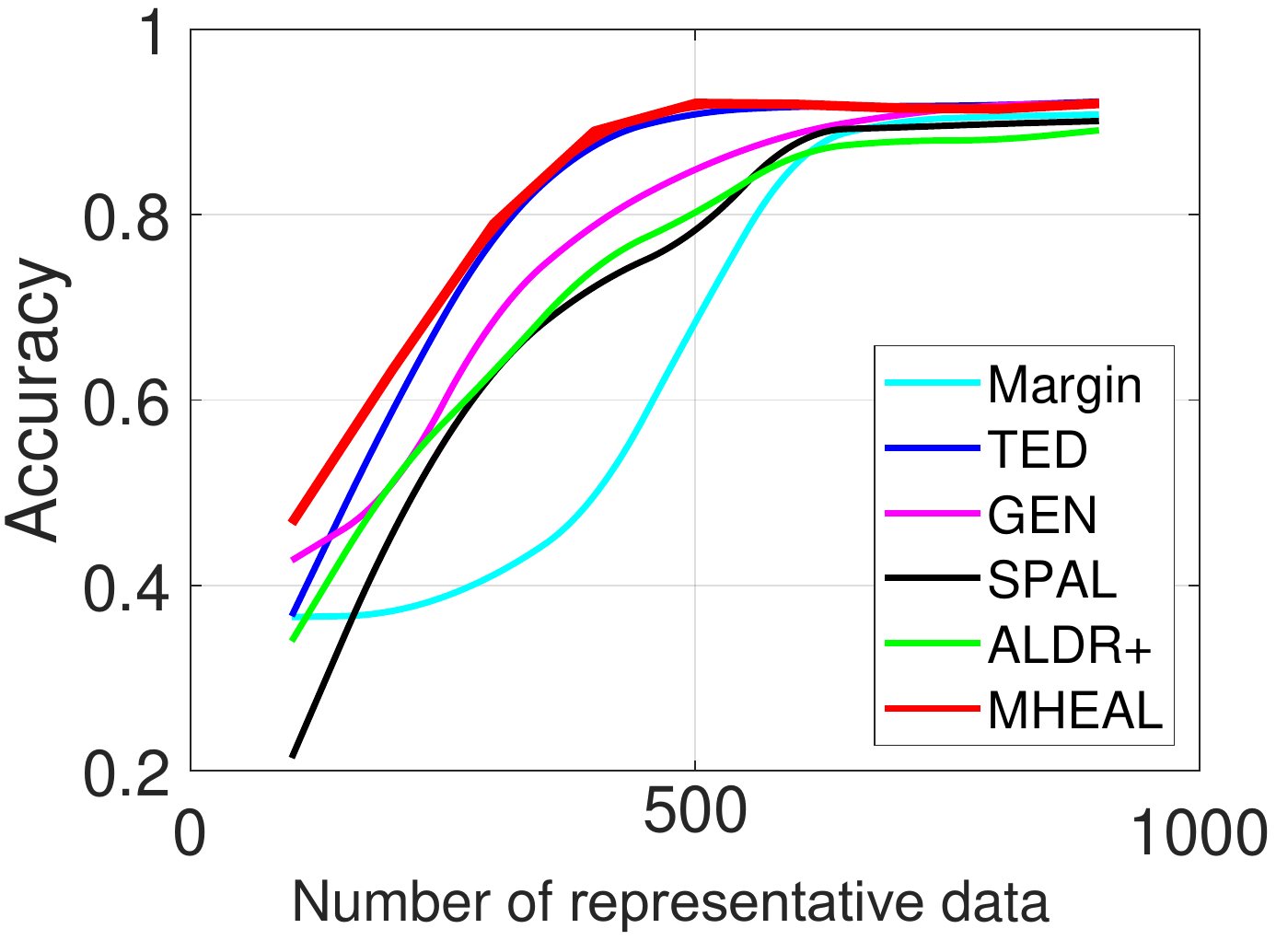}
\end{minipage}
} 
\caption{Data-efficient learning  using  representative data from Phishing, Adult, Satimage, and MNIST by invoking typical active learning baselines.
 }  
 \label{Learning_by_small_data_from_Phishing}
\end{figure*}

 \begin{table*} 
 \caption{Accuracy statistics of classification on  representative data via different deep AL baselines. }
  \renewcommand{\arraystretch}{1.2}
 \setlength{\tabcolsep}{1.5pt}{
\begin{center}
\scalebox{1.29}{
 \begin{tabular}{c|c  c c c  |c  c cc |c  c ccc  } 
\hline
\multirow{2}{*}{Algorithms}  &    \multicolumn{4}{c|}{MNIST (CNN)}  &         \multicolumn{4}{c|}{CIFAR-10 (ResNet20)}&   \multicolumn{4}{c }{CIFAR-100  (ResNet20)} \\
    &     100  &     200 &     300  &  500     &   1K    &   2K  &5K &10K &   2K    &   5K  &10K  &20K \\
\hline
K-means \cite{cao2020shattering}&   0.7139  & 0.8752       &  0.9369  &   0.9409    &0.3245&0.3889   &0.5015  &   0.7241       &  0.2223&    0.2631    &     0.3034    &   0.4467    \\
K-medoids \cite{cao2020shattering}&  0.8465  & 0.9229       &  0.9342  &   0.9478        & 0.4794    & 0.5680&0.7045 &   0.7676   &  0.2625   &  0.3031& 0.4456&0.5278   \\
Hierarchical  Tree \cite{dasgupta2008hierarchical} &0.8036 & 0.8764   &0.8956  & 0.9123        &        0.5194 & 0.5780  &0.7245      & 0.7636  &      0.2458 &      0.2931 &   0.4055  &0.4674  \\
Max Entropy \cite{gal2017deep}  &    0.6953   & 0.9056       & 0.9264  & 0.9655          &  0.4694 &0.5480&  0.6945   & 	0.7807     & 0.2245   & 0.3031 & 0.4250 &0.5156\\
Variation Ratios  \cite{gal2017deep}& 0.6707   & 0.8595       &  0.9226  & 0.9399          & 0.4894  &0.5980 &  0.7418    & 0.7902      & 0.2158   & 0.3232 & 0.4301 &0.5264  \\
 Core-set \cite{sener2018active}& 0.6310   &    0.8414     &  0.9035  &  0.9134           &  0.4546    &0.6369   & 0.7403     &  0.7858    &0.2645   &  0.33310&0.4654&0.5364 \\
BALD \cite{houlsby2011bayesian}&  0.7601   & 0.8864       & 0.9373  & 0.9636         &  0.5298    & \textbf{0.6510}& 0.7057&  0.7764    & 0.2047   & 0.3104 & 0.4208 &0.5298   \\
CDAL+CoreSet \cite{agarwal2020contextual}&     0.8578 & 0.8972   &0.9346        & 0.9569       & 0.5194& 0.5780&  0.6923         &   0.7718&0.2213 & 0.3031&0.4356&0.4854  \\
CDAL+Reinforcement  \cite{agarwal2020contextual}&  0.8347 & 0.8878   &0.9232        & 0.9456   &  0.4794          & 0.4880     & 0.6202& 0.7898&        0.2413    &  0.3231 &  0.4351&0.4295 \\
VAAL+VAE+Adversial \cite{sinha2019variational}&0.8105 & 0.8502   &0.9012        & 0.9312          &       0.3287     &0.4021   & 0.6032& 0.7312 &  0.2015&    0.2878   & 0.3761&0.4756\\
MHEAL&   \textbf{0.8648}   &   \textbf{0.9367}     & \textbf{0.9401}  & \textbf{0.9785}    & \textbf{0.5409} &  {0.6203}  &  \textbf{0.7842}    &\textbf{0.8146}   & \textbf{0.3201} &   \textbf{0.3831}   & \textbf{0.5052} & \textbf{0.5505} \\
\hline
\end{tabular}}
\end{center}}
\label{table_Accuracy_statistics_small_data}
\end{table*}

\begin{figure*} 
\subfloat[Unsupervised learning vs. MHEAL]{
\begin{minipage}[t]{0.33\textwidth}
\centering
\includegraphics[width=1.8in,height=1.52in]{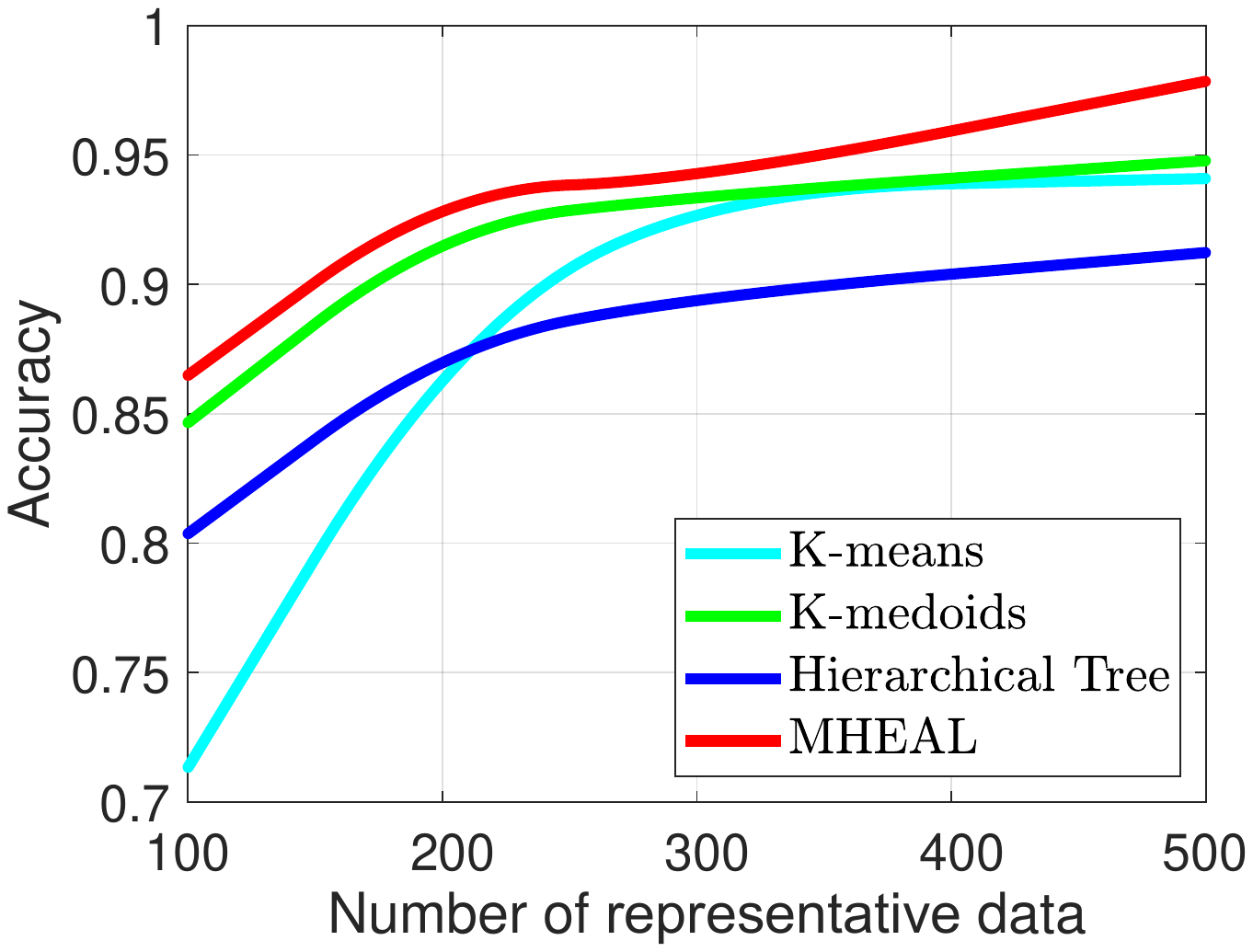}
\end{minipage}
}
\subfloat[Supervised learning vs. MHEAL]{
\begin{minipage}[t]{0.33\textwidth}
\centering
\includegraphics[width=1.8in,height=1.52in]{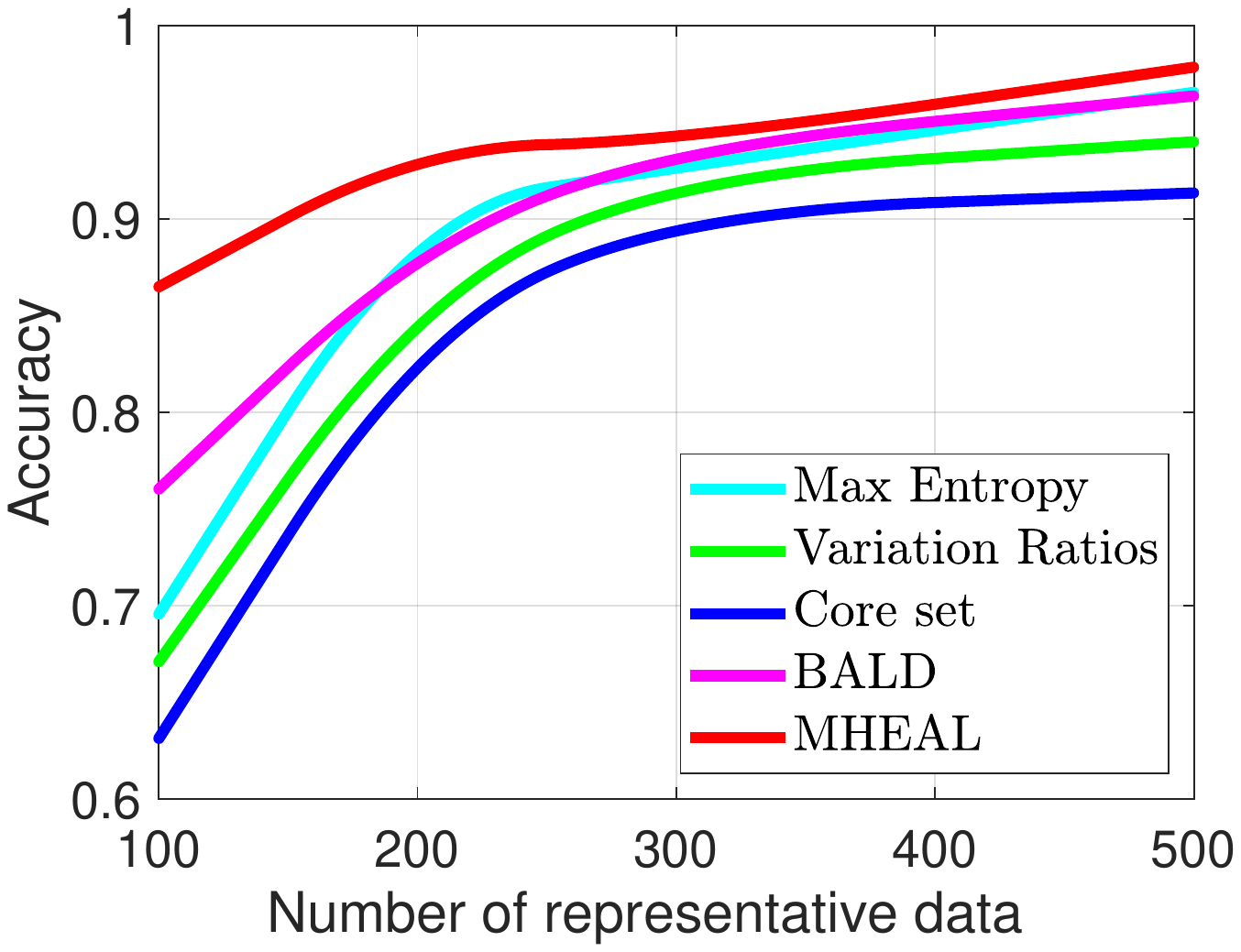}
\end{minipage}
} 
\subfloat[Deep learning vs. MHEAL]{
\begin{minipage}[t]{0.33\textwidth}
\centering
\includegraphics[width=1.8in,height=1.52in]{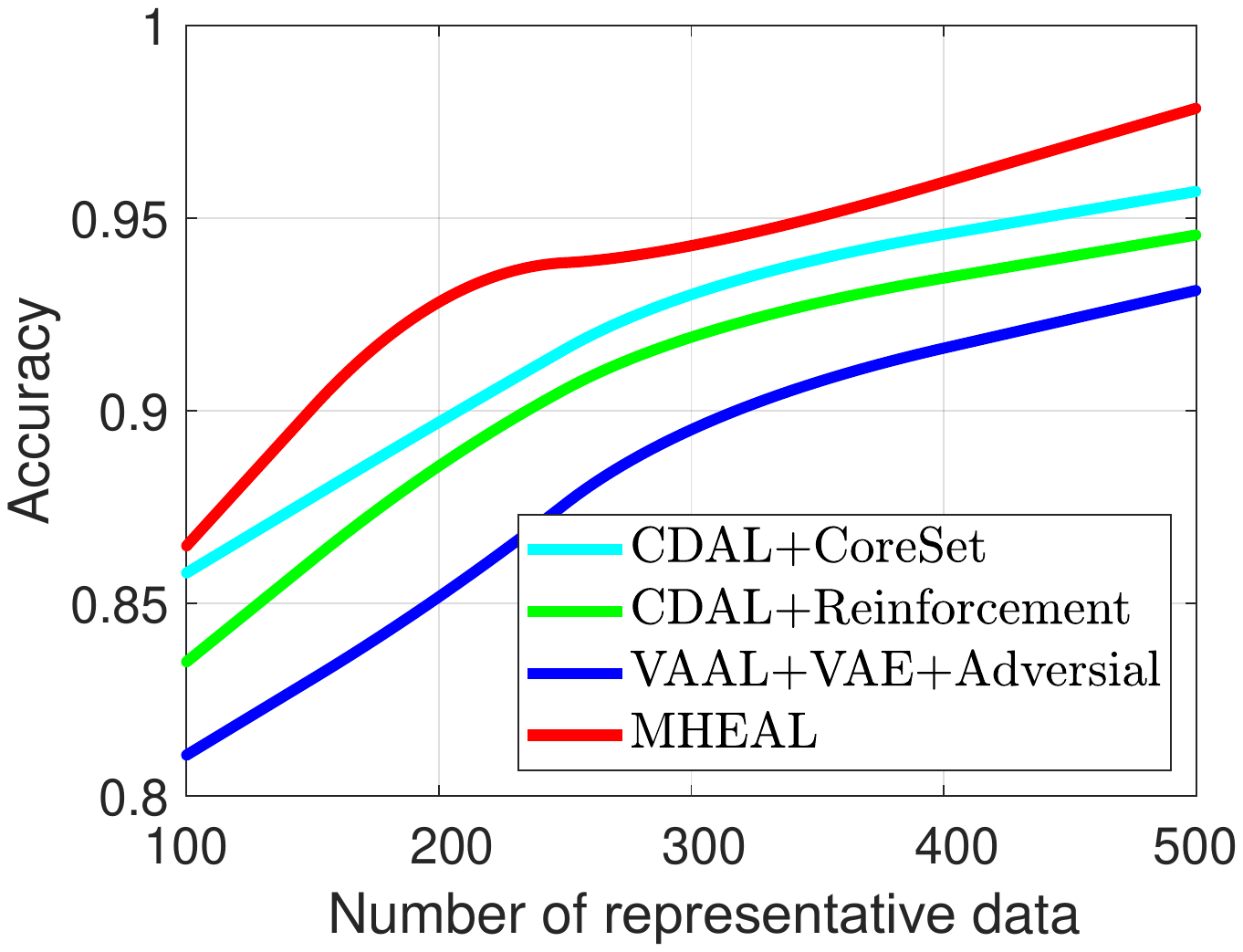}
\end{minipage}
} 
\caption{Data-efficient learning  using  representative data from  MNIST. (a)Unsupervised learning vs. MHEAL. (b)Supervised learning vs. MHEAL. (c)Deep learning vs. MHEAL.
 }  
\end{figure*}

\begin{figure*} 
\subfloat[Unsupervised learning vs. MHEAL]{
\begin{minipage}[t]{0.33\textwidth}
\centering
\includegraphics[width=1.8in,height=1.52in]{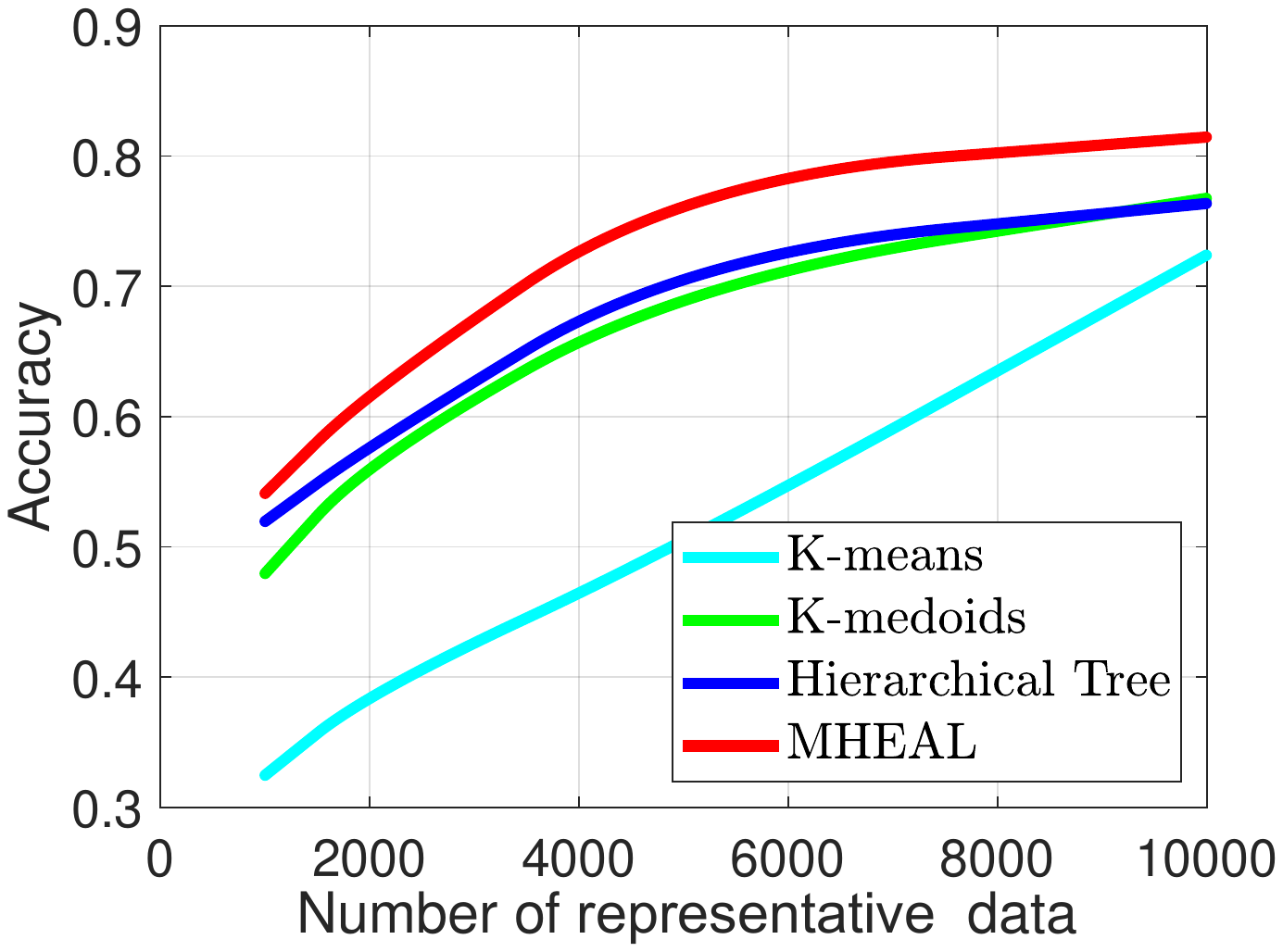}
\end{minipage}
}
\subfloat[Supervised learning vs. MHEAL]{
\begin{minipage}[t]{0.33\textwidth}
\centering
\includegraphics[width=1.8in,height=1.52in]{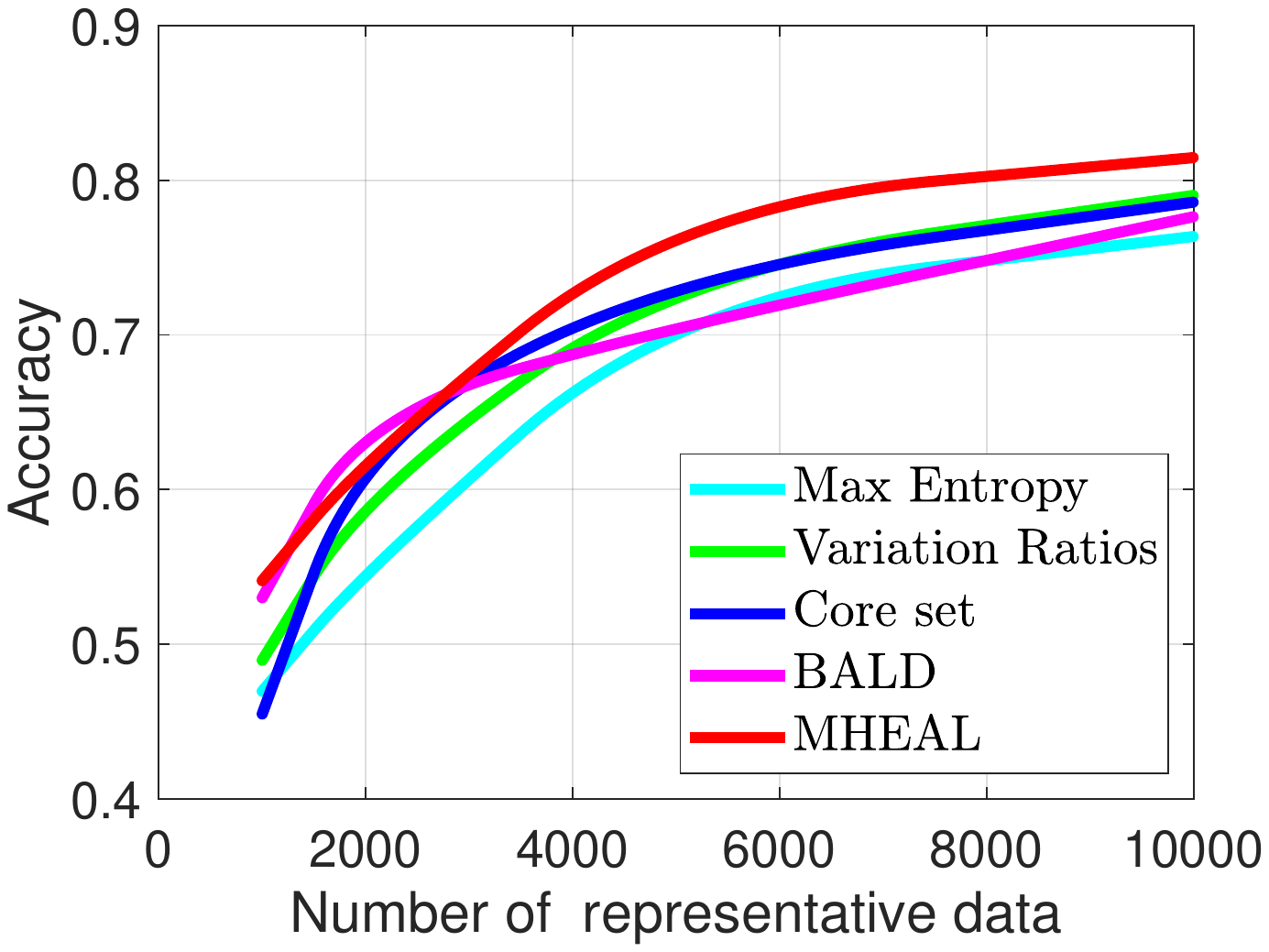}
\end{minipage}
} 
\subfloat[Deep learning vs. MHEAL]{
\begin{minipage}[t]{0.33\textwidth}
\centering
\includegraphics[width=1.8in,height=1.52in]{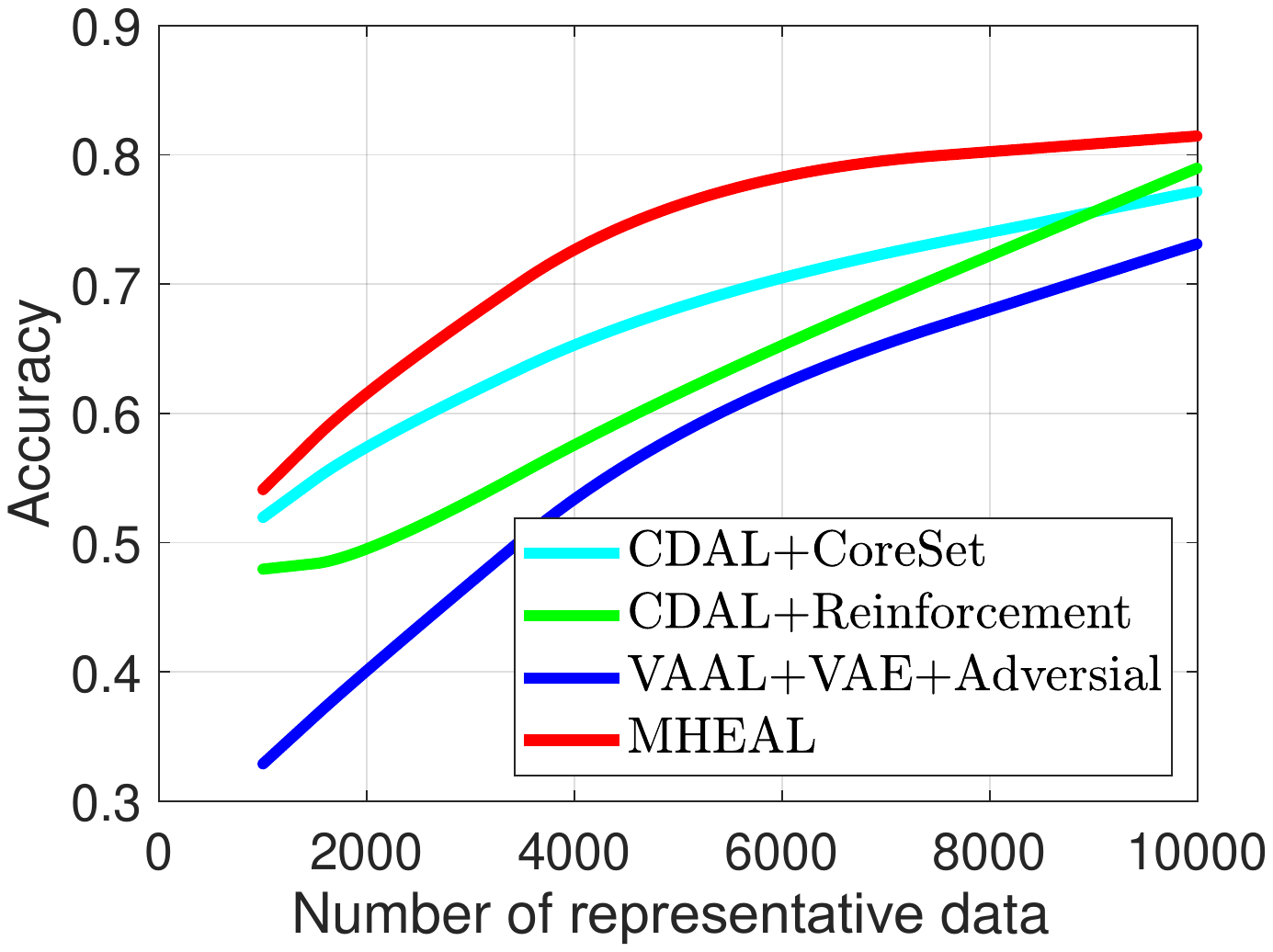}
\end{minipage}
} 
\caption{Data-efficient learning  using  representative data from  CIFAR10. (a)Unsupervised learning vs. MHEAL. (b)Supervised learning vs. MHEAL. (c)Deep learning vs. MHEAL.
 }  
\end{figure*}

\begin{figure*} 
\subfloat[Unsupervised learning vs. MHEAL]{
\begin{minipage}[t]{0.33\textwidth}
\centering
\includegraphics[width=1.8in,height=1.52in]{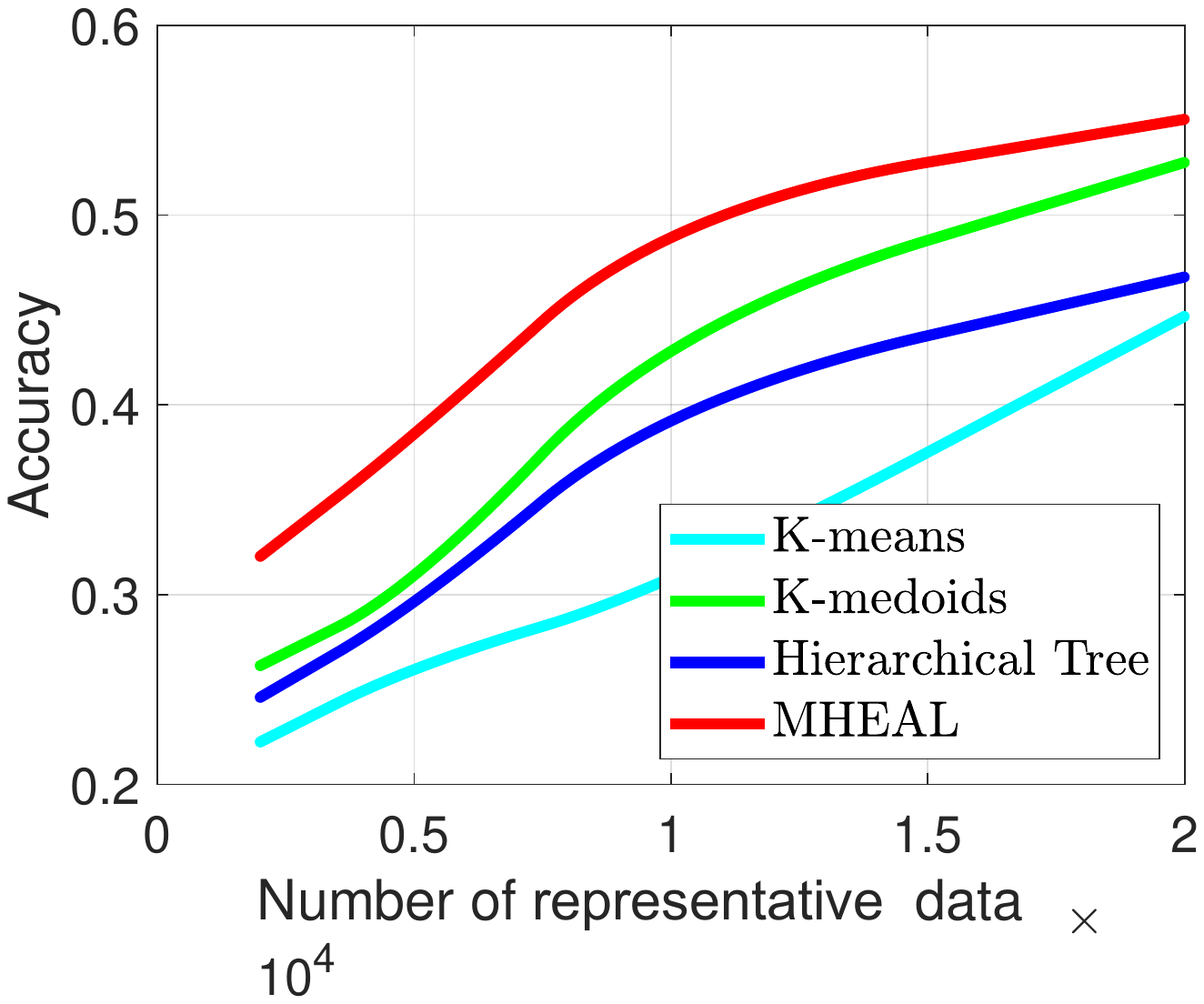}
\end{minipage}
}
\subfloat[Supervised learning vs. MHEAL]{
\begin{minipage}[t]{0.33\textwidth}
\centering
\includegraphics[width=1.8in,height=1.52in]{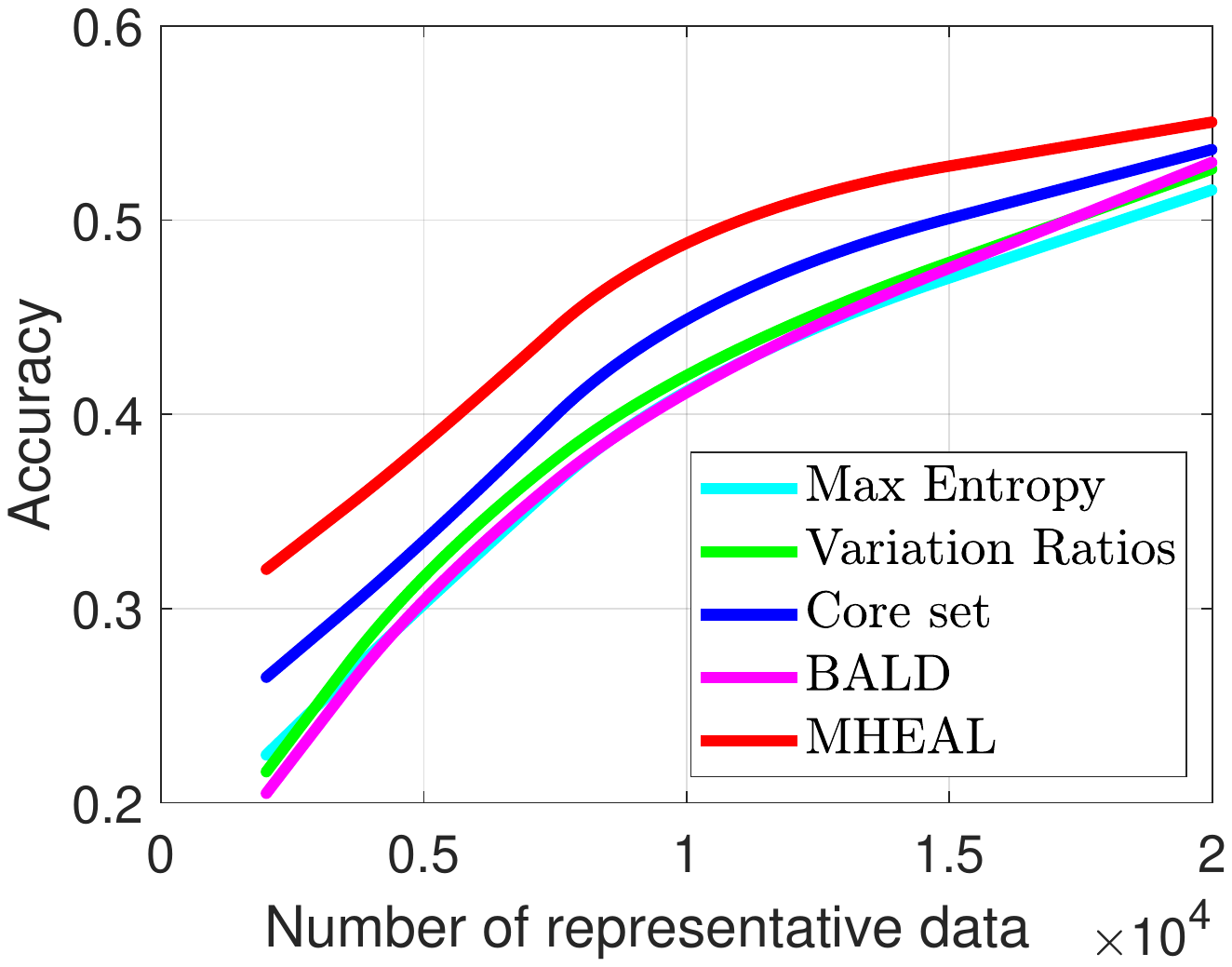}
\end{minipage}
} 
\subfloat[Deep learning vs. MHEAL]{
\begin{minipage}[t]{0.33\textwidth}
\centering
\includegraphics[width=1.8in,height=1.52in]{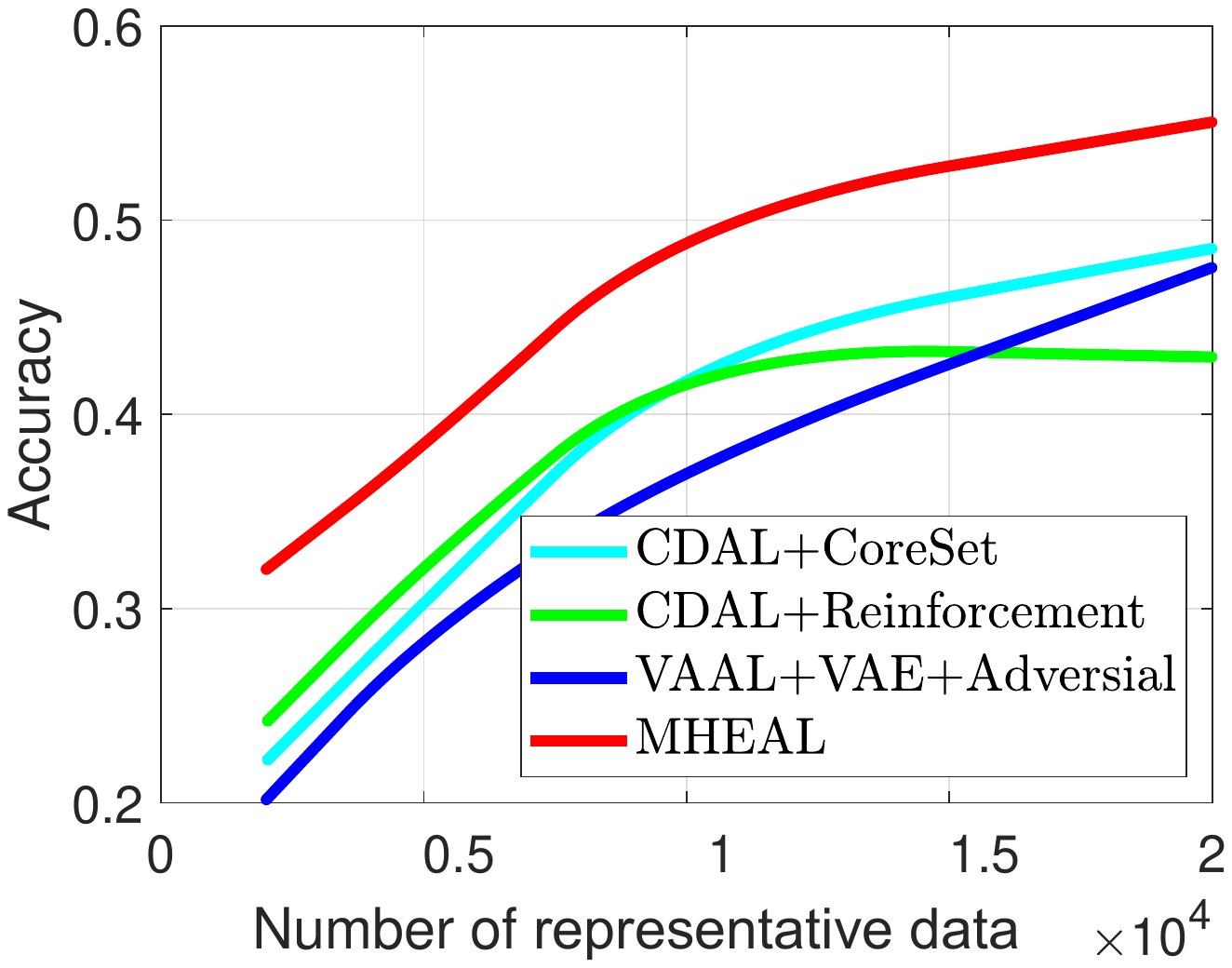}
\end{minipage}
} 
\caption{Data-efficient learning  using  representative data from  CIFAR100. (a)Unsupervised learning vs. MHEAL. (b)Supervised learning vs. MHEAL. (c)Deep learning vs. MHEAL.
 }  
\end{figure*}

\subsection{Data-Efficient Classification}
The effectiveness of learning on   representative data is evaluated by active sampling from input distribution. The selected sampling baselines are Margin\cite{tong2001support}, TED \cite{yu2006active}, GEN \cite{du2015exploring}, SPAL \cite{tang2019self}, ALDR+\cite{gu2020efficient}, and our MHEAL.  

 {Details of the above baselines are as follows. 1) Margin selects those unlabeled data which  derive the minimum margin distances. 2)  Transductive experimental design (TED)  selects data
that represent   the   unlabeled data  using kernel incremental tricks.   3)  General AL framework (GEN)   queries the samples with both representativeness  and diversity.   4) Self-paced AL (SPAL)   takes easiness, informativeness, and representativeness into an uniform  optimization.  5) ALDR+ selects  those discriminative and representative samples. }

 {The selected experimental data are typical UCI datasets, including Phishing, Adult, Satimage, and MNIST. The used classifier tool is LIBSVM following \cite{gu2020efficient}. To present fair comparison, we    use SphericalK-means without AutoEncoder  and DEC to perform MHEAL, and use a single query version of ALDR+. Note that Gu \emph{et al.}  \cite{gu2020efficient} presented a bath query strategy for AL.  Learning curves of these baselines with average accuracies  over 100 times of running are presented in Figure~\ref{Learning_by_small_data_from_Phishing}, where the initial number of labeled data are all the class numbers of datasets.  The results show that the unsupervised SphericalK-means of MHEAL has natural advantages for few   labeled data. The other supervised baselines which need  sufficient labels to estimate subsequent sampling, show poor performance against sufficient  numbers of  representative data.  With the increase of the number of  representative data, the performance disagreements of those baselines are reduced.  }

 {As we can observe from the above analysis, MHEAL adopts an unsupervised manner  to learn  on  representative data, which shows expressive representation performance than those supervised methods. This helps  MHEAL to actively spend  tighter label complexity  to converge into a desired generalization error, verifying the theoretical results of Theorem~1.}

 {\subsection{Data-Efficient Deep Classification}}
 As we stated in the introduction, the typical classification learning by  representative data can be implemented by AL.
We thus study the classification ability with regard on  representative data  derived by our MHEAL and the state-of-the-art deep AL baselines. Table~\ref{table_Accuracy_statistics_small_data} presents   the accuracies of  classification on  representative data of ten baselines on the datasets of  MNIST, CIFAR-10, and CIFAR-100, respectively. For the typical AL baselines, e.g., the first seven baselines and MHEAL,  training MNIST involves the  Bayesian CNN of   \cite{gal2017deep},  and  training CIFAR-10 and CIFAR-10 involve  the ResNet20 of \cite{ResNet20}. (Our experience  shows that  CNN has expressive modeling using few labels on MNIST.)
Besides them, CDAL+CoreSet\cite{agarwal2020contextual}, CDAL+Reinforcement \cite{agarwal2020contextual}, and   VAAL+VAE+Adversial \cite{sinha2019variational} are set as the     network architectures-based deep AL algorithms.   Their details are presented as follows.

 {Details of network architectures-based deep AL baselines.}
 (1) VAAL+VAE+Adversarial (VAAL: Variational adversarial active learning) \cite{sinha2019variational}.  It is a deep AL framework, which adopts a variational autoencoder (VAE) and an adversarial network to learn the latent space.  \footnote{Code: \url{https://github.com/sinhasam/vaal} }
 (2) CDAL+CoreSet (CDAL: Contextual Diversity for Active Learning) \cite{agarwal2020contextual}.   This deep AL approach introduces the contextual diversity  hinges to observe   the probability vectors predicted by a CNN, where the vector regions of interest typically contain  diverse label information. This approach cooperates with  a core-set to evaluate the contextual  diversity.\footnote{Code: \url{https://github.com/sharat29ag/CDAL}} 
(3) CDAL+Reinforcement  (CDAL: Contextual Diversity for Active Learning) \cite{agarwal2020contextual}.   Different from CDAL+CoreSet, this deep AL approach cooperates with  a
reinforcement learning  policy. \footnote{Code:\url{https://github.com/sharat29ag/CDAL}}

We follow \cite{cao2020shattering} to select 20 random samples to initialize the hypothesis  of AL. As the shown in Table~\ref{table_Accuracy_statistics_small_data}, 1) the unsupervised baselines such as K-means,  K-medoids, Hierarchical  Tree, and  Core-set  perform stably, varying the number of desired  representative data, 2) the typical supervised baselines such as Max Entropy, Variation Ratios, and BALD  perform not well if annotating few data; only sufficient  labeled data can  rapidly improve their accuracies, 3) the   network architectures-based deep AL, \eg, CDAL+CoreSet and CDAL+Reinforcement, and   VAAL+VAE+Adversial need large number of data and labels can show expressive modeling, while show very general accuracies among insufficient annotation budgets, 4) MHEAL employs deep clustering to extract structural clusters from the inputs, showing the state-of-the-art accuracies than  other baselines on MNIST. Technically, our locally MHE optimization in each  cluster yields expressive distribution matching, which  derives  comprehensive covering on the input features.  Better accuracies    then are naturally presented.

  {Figures~13-15 present  the learning curves for Table~3. This visualizes the  performance disagreement of those baselines in a different view. Compared to the experimental results of Section~7.4, those deep learning baselines need more labeled data to strength the advantages of their network configurations. However, unsupervised deep clustering does not need labels, and can obtain more useful information for subsequent MHE optimization, presenting positive guarantees. }
 \begin{table*} 
 \caption{Accuracy statistics of classification on  repeated  representative data via different deep AL baselines. }
  \renewcommand{\arraystretch}{1.2}
 \setlength{\tabcolsep}{1.5pt}{
\begin{center}
\scalebox{1.26}{
 \begin{tabular}{c|c  c c c  |c  c cc |c  c ccc  } 
\hline
\multirow{2}{*}{Algorithms}  &    \multicolumn{4}{c|}{MNIST (CNN)}  &         \multicolumn{4}{c|}{CIFAR-10 (ResNet20)}&   \multicolumn{4}{c }{CIFAR-100  (ResNet20)} \\
    &     100  &     200 &     300  &  500     &   1K    &   2K  &5K &10K &   2K    &   5K  &10K  &20K \\
\hline
K-means \cite{cao2020shattering}&   0.7102   & 0.8687       &  0.9312  &   0.9309 &0.3198& 0.3920   &0.4989  &   0.7210     & 0.2201&    0.2603    &    0.3001    &    0.4401  \\
K-medoids \cite{cao2020shattering}&  0.8345  &  0.9217       &  0.9330  &   0.9401        & 0.4702   &0.5602&0.7102 &   0.7670   &  0.2526  &    0.3021&  0.4405& 0.5201 \\
Hierarchical  Tree \cite{dasgupta2008hierarchical} &0.7925 & 0.8698   &0.8875  & 0.9103        &       0.5012 & 0.5623  &0.7109    & 0.7587  &       0.2454&  0.2901 &    0.4058  &0.4623 \\
Max Entropy \cite{gal2017deep}  &   0.6846  &  0.8952       & 0.9105  & 0.9567          &  0.4598 &0.5412&  0.7201   & 	 0.7712    & 0.2201  & 0.2989& 0.4108 &0.5058\\

Variation Ratios  \cite{gal2017deep}& 0.6608   & 0.8436       &  0.9106  & 0.9219 & 0.4802  & 0.5875 &  0.7398   &  0.7874     & 0.2048   &0.3130&0.4257&  0.5147  \\
 Core-set \cite{sener2018active}& 0.6293   &    0.8386     &  0.8956  &  0.9103           &  0.4501    &0.6315   &  0.7309    &  0.7717    &0.2587  & 0.3212&  0.4536&  0.5287 \\
BALD \cite{houlsby2011bayesian}&  0.7542   & 0.8798       &  0.9288  & 0.9541      &  0.5102    & \textbf{0.6426}& 0.6923&  0.7654    & 0.1987   &0.3026 &   0.4156 &0.5185  \\
CDAL+CoreSet \cite{agarwal2020contextual}&     0.8502 & 0.8913   &0.92986        & 0.9510      & 0.5065& 0.5687&   0.6823        & 0.7689& 0.2103   &  0.2987 &    0.4256&0.4745  \\
CDAL+Reinforcement  \cite{agarwal2020contextual}&  0.8298 & 0.8768   &0.9201 & 0.9398   &  0.4687         &0.4770   &0.6103&  0.7792&    0.2365   &  0.3108 &    0.4243&0.4158 \\
VAAL+VAE+Adversial \cite{sinha2019variational}&0.8023 & 0.8409  &0.8956       &  0.9269      &       0.3128    & 0.3956  &  0.6012& 0.7226&  0.1948&   0.2745&   0.3657&0.4635\\
MHEAL&   \textbf{0.8600}   &   \textbf{0.9302}     & \textbf{0.9369}  & \textbf{0.9701}    & \textbf{0.5378} &  {0.6156}  &  \textbf{0.7800}    &\textbf{0.8042}   & \textbf{0.3187} &   \textbf{0.3745}   & \textbf{0.4989} & \textbf{0.5459} \\
\hline
\end{tabular}}
\end{center}}
\end{table*}
 
 {\subsection{Data-Efficient Deep Classification with Repeated  Data}}
 {Sections~7.4 and 7.5  have presented the   classification results of  learning    representative data by employing several state-of-the-art baselines. However, some of them depend on incremental updates of classification hypothesis, in which those updates may   prefer similar or repeated samples  due to the unbalanced class labels in initial data \cite{kirsch2019batchbald}. We thus consider a special scenario: data-efficient deep classification with repeated data following  \cite{kirsch2019batchbald}. The goal is to test whether the baseline will be mislead by repeated data.}

 {Following the settings of Section~7.5, we randomly select 10,000 training data  and add them into the original training sets of   MNIST, CIFAR10, and CIFAR100, respectively. We restart the experiments of Table~3 and report the results in Table~4. The learning curves are then represented in Figures~16 to 18 (see supplementary material). The results show that those baselines who invoke the incremental updates of  hypothesis  may continuously select those scarce repeated samples. However, the unsupervised baselines aim to find the representation samples, which provide fair or sufficient data for each class. It is thus that their biases are lower than those baselines using   incremental updates of hypothesis.}

 \subsection{Data-Efficient Deep Classification with Noisy  Data} 

 {Learning with repeated data is inevitable. However, the tested datasets are clean data; hypothesis updates usually will be misguided by noisy labels due to their larger incremental estimations.  We thus consider a more practical  scenario: learning from noisy  representative data. The goal is to observe  the responses of those compared baselines.}

 {Following the settings of Section~7.5, we randomly select 10,000 noisy data  and add them into the original training sets of MNIST, CIFAR10, and CIFAR100, respectively. We restart the experiments of Table~3 and report   the results in Table~5 (see supplementary material). The learning curves are then represented in Figures~19 to 21 (see supplementary material). The results show that those baselines who invoke the hypothesis incremental updates may be more easily misleaded by noises than clean samples. However, the unsupervised baselines use a global manner to  find those representation samples, which  fairly glances the samples of each class. Therefore, noisy biases of unsupervised methods   are lower than supervised baselines since  noises may return large perturbations to incremental updates of hypothesis.}

\section{Discussion}
Data-efficient  learning using  representative data is the future trend for AI, and deep AL is  one of the effective ways to implement this goal. In this work, MHE is utilized to  extract    representative data from the topology of decision boundaries.
Deep clustering  firstly provides a rough characterization for the clusters before the extraction, yielding superior performance than learning  without  boundary interactions.

However, such improvements depend on the clustering structures. Once the input distribution has no inherent clustering prototypes, our idea will degenerate into pure MHE optimizations in splitting regions. For real-world tasks, MHEAL will benefit the weakly-supervised \cite{li2019towards} and semi-supervised \cite{gadde2014active} sampling issues. For noisy supervision \cite{veit2017learning}, adopting our idea may reduce the wrong estimation rate on noises. Moreover, MHE optimization may bring geometric benefits for the outlier recognition and out-of-distribution detection \cite{hsu2020generalized}.
Learning the representative data from the spherical topology, adversarial attacks \cite{dong2018boosting},  and malicious intrusion around the decision boundaries may be easier. Capturing the characterization of decision boundaries may fool the  
network filtering  and mailbox   defense systems.
 
\section{Conclusion}
Learning from scratch using  representative data is helpful for data-efficient AI, and deep AL is  one of the effective ways to implement this goal.  This paper presents a novel idea of hyperspherical $\ell_0$  MHE   from the physical geometry, to actively learn  the  representative data from the homeomorphic tubes of spherical manifolds.   The proposed MHEAL  algorithm  employs  deep spherical clustering, which provides a pre-estimation for the input distribution, characterizing the topology of version space.  Then the max-min optimization for MHE  derives effective decision boundaries to match  each  cluster, resulting in a lower distribution loss than   $\ell_1$ and $\ell_2$ MHE. In-version-space sampling also showed more expressive modeling for classification than  sampling in out-version-space, which manifests the significance of our MHEAL. 
 Generalization  error and  label complexity bounds  prove  that MHEAL can safely converge.   {A series of experiment  support the claimed theoretical results.
In future, semi-supervised theory of MHEAL  may have   potential interests. }

\bibliographystyle{IEEEtran}
\bibliography{Reference}

\begin{IEEEbiography}
[{\includegraphics[width=1in,height=1.25in,clip,keepaspectratio]{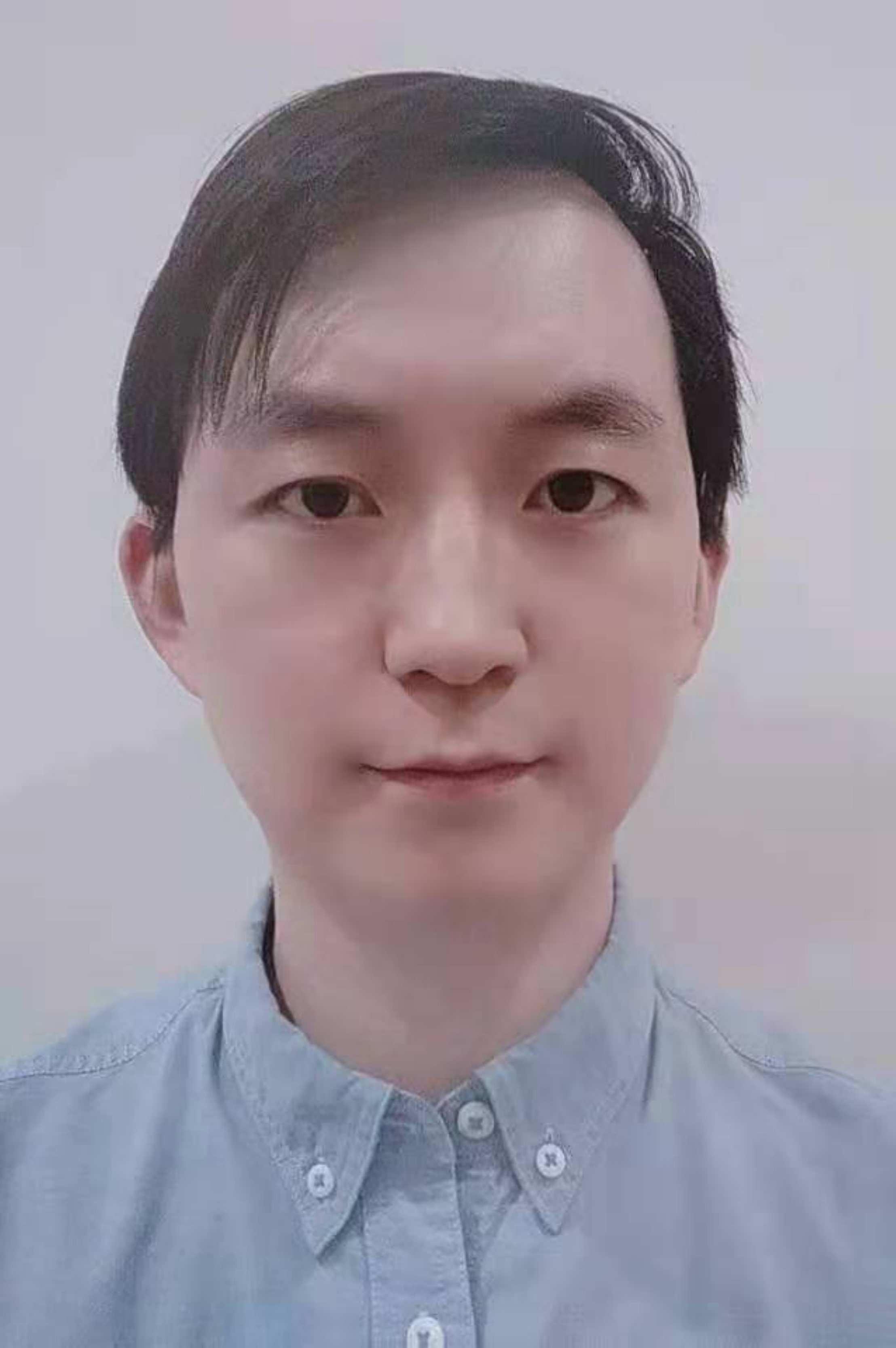}}]{Xiaofeng Cao} received his Ph.D. degree at Australian Artificial Intelligence Institute, University of Technology Sydney. He is currently an Associate
Professor at the School of Artificial Intelligence, Jilin
University, China and leading a Machine Perceptron
Research Group with more than 15 PhD and Master
students. He has published more than 10 technical
papers in top tier journals and conferences, such as
IEEE T-PAMI, IEEE TNNLS, IEEE T-CYB, CVPR,
IJCAI, etc. In 2021, he was nominated as a rising
star in “Lixin Outstanding Young Teacher Training
Program of Jilin University”. In 2022, he was nominated as a candidate in “The
Sixth Jilin Province Youth Science and Technology Talent Support Project”.
His research interests include PAC learning theory, agnostic learning algorithm,
generalization analysis, and hyperbolic geometry.
\end{IEEEbiography}

 \begin{IEEEbiography}
[{\includegraphics[width=1in,height=1.25in,clip,keepaspectratio]{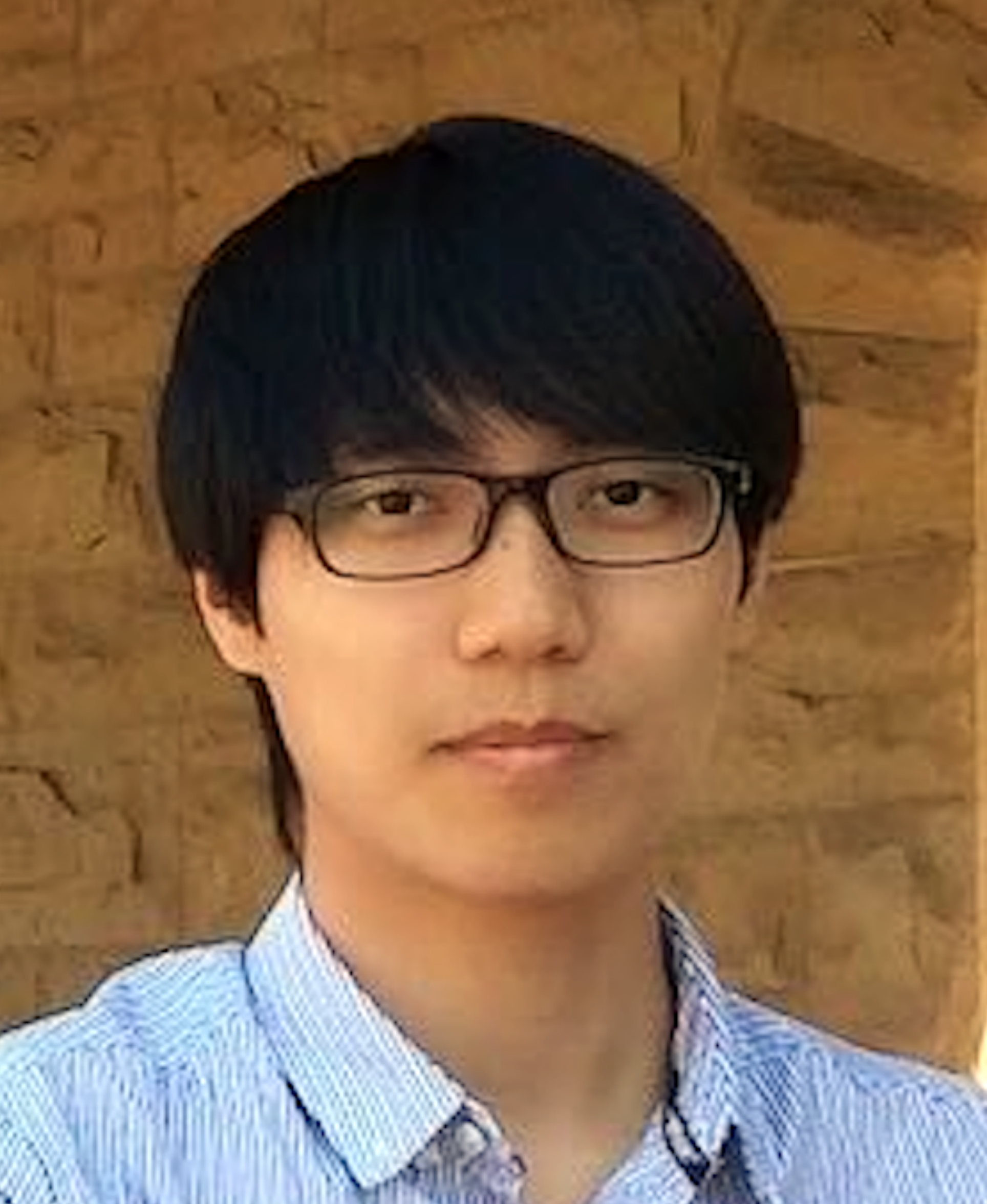}}]{Weiyang Liu} is currently conducting research at the University of Cambridge, UK and the Max Planck Institute for Intelligent Systems, Tübingen, Germany under the Cambridge-Tübingen Fellowship Program. Prior to joining this program, he has been with College of Computing, Georgia Institute of Technology, Atlanta, GA, USA. His research interests broadly lie in deep learning, representation learning, interactive machine learning and causality.
\end{IEEEbiography}

\begin{IEEEbiography}
[{\includegraphics[width=1in,height=1.25in,clip,keepaspectratio]{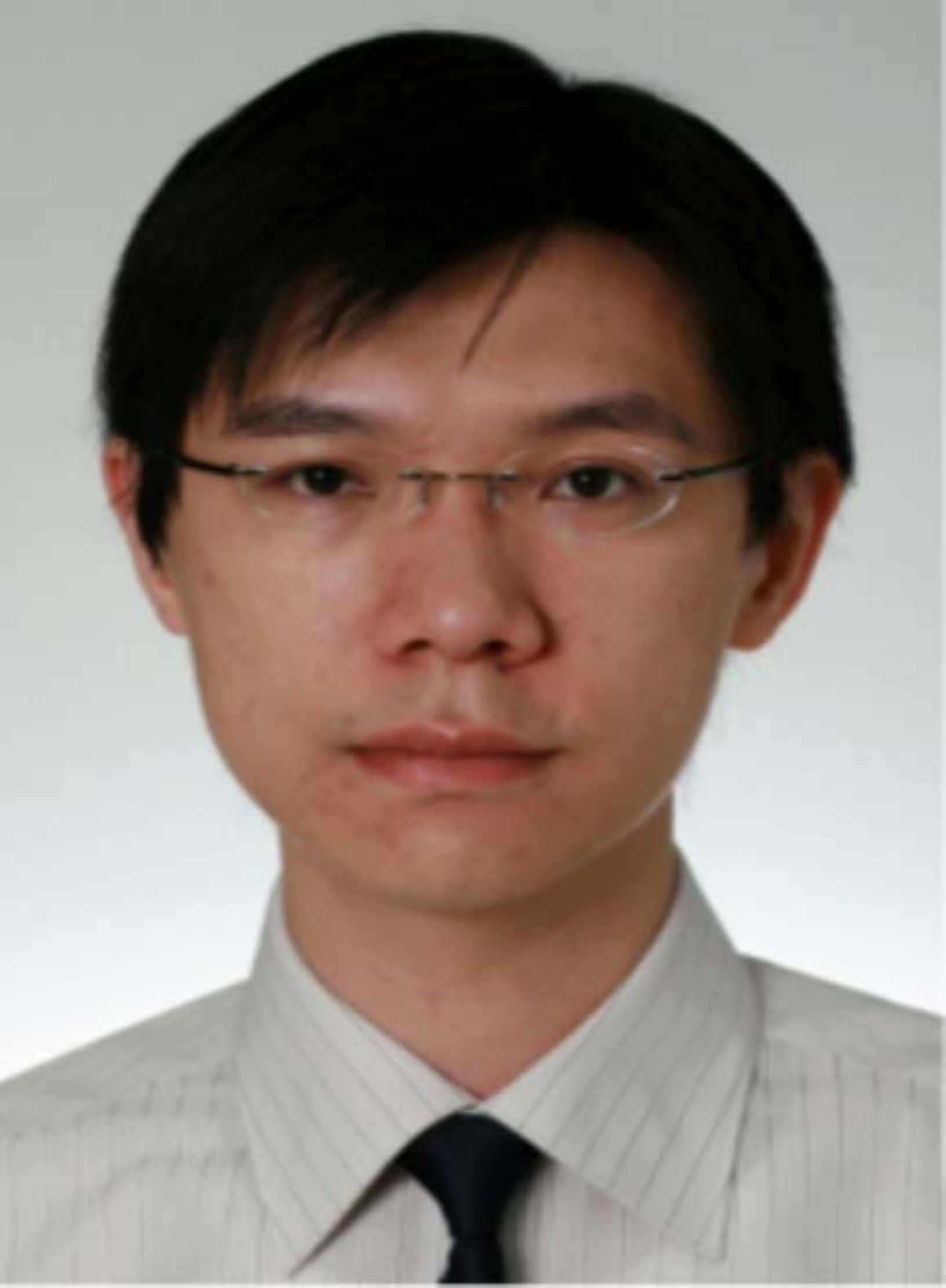}}]{Ivor W. Tsang}  is Professor of Artificial Intelligence, at University of Technology Sydney. He is also the Research Director of the Australian Artificial Intelligence Institute, and an   IEEE Fellow.  In 2019, his paper titled ``Towards ultrahigh dimensional feature selection for big data" received the International Consortium of Chinese Mathematicians Best Paper Award. In 2020, Prof Tsang was recognized as the AI 2000 AAAI/IJCAI Most Influential Scholar in Australia for his outstanding contributions to the field of Artificial Intelligence between 2009 and 2019. His works on transfer learning granted him the Best Student Paper Award at International Conference on Computer Vision and Pattern Recognition 2010 and the 2014 IEEE Transactions on Multimedia Prize Paper Award. In addition, he had received the prestigious IEEE Transactions on Neural Networks Outstanding 2004 Paper Award in 2007.  
\par Prof. Tsang serves as a Senior Area Chair for Neural Information Processing Systems and Area Chair for International Conference on Machine Learning, and the Editorial Board for  Journal Machine Learning Research, Machine Learning, 
Journal of Artificial Intelligence Research, and IEEE Transactions on Pattern Analysis and Machine Intelligence. 
\end{IEEEbiography}

 \newpage
\section*{Appendix}

\subsection*{Proofs}
Proof of Proposition~1.
\begin{proof}
Given $\mathcal{L}_{ \mathbb{E}_{0,d}}( {\hat w}_N)$ be the energy of MHE with an initial   sample $ {\hat w}_N$ where $\mathcal{W}=\{\hat w_1,\hat w_2,..., \hat w_{N-1} \}$, there must exist an inequality of 
\[  \Big(\prod_{ N-1>i>j }  \|\hat w_i-\hat w_j\|\Big) \times  \big ((N-1)\min_{\hat w_j \in \mathcal{W}} \|\hat w_N-\hat w_j\|      \big)       \leq   \mathcal{L}_{ \mathbb{E}_{0,d}}( {\hat w}_N),\]
where       $\min_{\hat w_j \in \mathcal{W}} \|\hat w_N-\hat w_j\| $ denotes the minimal geodesic distance between $\hat w_N$ to the samples of $\mathcal{W}$,    and $\hat w_j \in   \mathcal{W}$ returns $N-1$ samples. Considering that     $\min_{\hat w_j \in \mathcal{W}} \|\hat w_N-\hat w_j\|\leq  \min_{\hat w_i, \hat w_j \in \mathcal{W}}\|\hat w_i-\hat w_j\|$,  we thus obtain an approximation of 
\[    \big (\min_{\hat w_j \in \mathcal{W}} \|\hat w_N-\hat w_j\| \big)^{\frac{N^2-3N+2}{2}} \leq  \Big(\prod_{ N-1>i>j }  \|\hat w_i-\hat w_j\|\Big),  \]  
where $\prod_{N-1>i>j } $ performs $\frac{N^2-3N+2}{2}$ times of multiplication, 
there then exists 
\[\big (\min_{\hat w_j \in \mathcal{W}} \|\hat w_N-\hat w_j\|      \big)^{\frac{N^2-N}{2}}   \leq   \mathcal{L}_{ \mathbb{E}_{0,d}}( {\hat w}_N).\]
In short, given $\hat w_t$, the next observation on $\hat w_{t+1}$ is  then specified by invoking Eq.~(4). It is thus $\hat w_1,\hat w_2,..., \hat w_{N-1}$ is the optimal sequence from $\mathcal{P}$, thereby $\mathcal{L}_{\mathbb{E}_{0,d}}( {\hat w}_N)$ is the optimal hypersphere energy under the sequence of $\hat w_1,\hat w_2,..., \hat w_{N-1}$. On this setting, any energy  under a sub-optima $\hat w_N$ is lower than $\mathcal{L}_{\mathbb{E}_{0,d}}( {\hat w}_N)$. 
\cao{Proposition~1} is then as stated.
\end{proof}

Proof of Proposition~2.
\begin{proof}
Given $\hat w_i', \hat w_j' \in \mathbb{R}^{d+1}$ be the $(d+1)$-dimensional extension vector of  $\hat w_i, \hat w_j \in \mathbb{R}^{d}$,  respectively, 
let $P'$ be the  $(d+1)$-dimensional extension of $P$,  we know $\|\hat w'_i- \hat w'_j  \|=\|\hat w_i- \hat w_j  \|$. It is thus  $\|P\hat w_i- P\hat w_j \|^2=\|P'\hat w'_i- P'\hat w'_j \|^2$ since their $(d+1)$-dimensional projections still keep  zero variables. Given $d'=d+1$, let  $P'_{id'}$, $\hat w'_{id'}$ denote the $(d+1)$-dimension of  $P'_i$ and $\hat w_i'$, respectively, there exists $P'_{id'}w_{id'}=0$. \cao{ With such operation, there   exists $\|\hat w_i'- \hat w_j' \|^2=\|\hat w_i - \hat w_j  \|^2$, $\|P'\hat w_i'- P'\hat w_j' \|^2=\|P\hat w_i- P\hat w_j \|^2$. Besides, $P$ and $P'$ have a same  variance. Then the inequality of  Lemma~1 can be rewritten as}

\begin{equation*}
\begin{split}
&{\rm Pr} \Big \{(1-\varepsilon)\|\hat w_i'- \hat w_j' \|^2k \sigma^2<  \|P'\hat w_i'- P'\hat w_j' \|^2\\
& < (1+\varepsilon)\|\hat w_i'- \hat w_j' \|^2\kappa \sigma^2\Big\}   \geq 1-2{\rm exp} \Big   ( \frac{-\kappa\varepsilon^2}{8}  \Big).  \\
\end{split}
\end{equation*}
Because $P'\hat w_i'- P'\hat w_j' $ actually returns a $d$-dimensional vector, we thus say if $\kappa=d$, Lemma~1 then still holds.
\end{proof}

Proof of Lemma~2.
\begin{proof}
With the same vector extension of proof of Proposition~2, for any geodesic function  $d_\mathcal{M} \in \mathbb{R}^{d}$, it will return a zero operation on its $(d+1)-$dimensional extension.  It is thus the metric of $P\hat w_1, P\hat w_2$  mainly rely on their  operations    from 1 to $d$ dimension. Considering that  $d_\mathcal{M}( g(w_1), g(w_2)):=\nu\|w_1-  w_2\|$,  
$  d_\mathcal{M}(g(\hat w_1),g(\hat w_2))=\nu\Big(\sum_{j=1}^{N}\|\hat w_{1j}-\hat w_{2j}\|^2
\Big)^{1/2} \leq \nu\Big(\sum_{j=1}^{N}\|\hat w_{2j}-\hat w_{3j}\|^2
\Big)^{1/2}$, if $\nu=\nu'$, there holds for 
  $d'_\mathcal{M}( g'(w_1), g'(w_2))=\nu'\Big(\sum_{j=1}^{N}\|\hat w_{1j}-\hat w_{2j}\|^2
\Big)^{1/2} \leq \nu'\Big(\sum_{j=1}^{N}\|\hat w_{2j}-\hat w_{3j}\|^2\Big)$.
  Lemma~2 then holds.
\end{proof}

Proof of Lemma~3.

 \begin{proof}
Following  \cite{lin2020regularizing},  given a constant $e>0$,  a Gaussian random
projection matrix  matrix $P$  leads to \[\theta_\mathcal{M}(\hat w_1,\hat w_2)=\frac{\hat w_1^T\hat w_2}{\|  w_1^T  \|\cdot \|  w_2^T  \|}=\frac{1}{e}\frac{ (P \hat w_1)^T (P \hat w_2)}{\|  (P \hat w_1)^T  \|\cdot \|  (P \hat w_2)^T  \|}.\] 
Then  $\theta_\mathcal{M}(\hat w_1,\hat w_2)=\frac{1}{e} \theta_\mathcal{M}(P\hat w_1,P\hat w_2)    \leq   \theta_\mathcal{M}(\hat w_2,\hat w_3)= \theta_\mathcal{M}(P\hat w_2,P\hat w_3).$

With the same proof skill of Proposition~2, we extend   a $d+1$ dimension vector  with zero-element to $\hat w_1, \hat w_2, \hat w_3, ..., \hat w_N\in \mathbb{R}^{d+1}$ and obtain $\hat w_1', \hat w_2', \hat w_3', ..., \hat w_N'\in \mathbb{R}^{d+1}$. With the Theorem~1 of \cite{lin2020regularizing}, there exists

 \[  \frac{\frac{1}{e}\theta_\mathcal{M}(P\hat w_1',P'\hat w_2')-\varepsilon} {1+\varepsilon} < \theta_\mathcal{M}(P'\hat w_1',P'\hat w_2')< \theta_\mathcal{M}(P\hat w_2',P'\hat w_3')  \]

With a zero operation on the $d+1$ dimension for any $\hat w_i'$, we know  $\theta_\mathcal{M}(P\hat w_1,P\hat w_2)=\theta_\mathcal{M}(P'\hat w_1',P'\hat w_2')$, and  $\theta_\mathcal{M}(P'\hat w_2',P'\hat w_3')=\theta_\mathcal{M}(P'\hat w_2,P'\hat w_3) $,  the  inequality  of Lemma~3  then holds. 
 
 \end{proof}

 Proof of Proposition~3.
 \cao{
\begin{proof}
 Let   $\ell_{LBM}( {\hat w}_N)$ denote the approximation loss of our lower bound approximation on  $ {\hat w}_N$.     With Proposition~2, the approximation loss of Proposition~1 is given by 
\begin{equation*}
\begin{split}
& \ell_{LBM}( {\hat w}_N)=   \mathcal{L}_{ \mathbb{E}_{0,d}}( {\hat w}_N)- \Big (\min_{\hat w_j \in \mathcal{W}} \|\hat w_N-\hat w_j\|      \Big)^{\frac{N^2-N}{2}}\\
  &\leq \Big (\max_{\hat w_j \in \mathcal{W}} \|\hat w_N-\hat w_j\|      \Big)^{\frac{N^2-N}{2}}-\Big (\min_{\hat w_j \in \mathcal{W}} \|\hat w_N-\hat w_j\|      \Big)^{\frac{N^2-N}{2}}.\\
 \end{split}
\end{equation*}
We next analyze this upper bound written as $\ell^{upper}_{LBM}( {\hat w}_N)$. With Lemma~2.3 of \cite{qian2001non}, there  exists $|x-y|^p\leq 2^{p-1} |x^p-y^p|, p\geq 1$. We thus control 
$\ell^{upper}_{LBM}( {\hat w}_N)$ by
\begin{equation*}
\begin{split}
& \frac{1}{\sqrt{2}^{N^2-N-2}}\Big (\max_{\hat w_j \in \mathcal{W}} \|\hat w_N-\hat w_j\| -  \min_{\hat w_j \in \mathcal{W}} \|\hat w_N-\hat w_j\|   \Big)^{\frac{N^2-N}{2}} \\
&  \leq \ell^{upper}_{LBM}( {\hat w}_N)\leq \Big (\max_{\hat w_j \in \mathcal{W}} \|\hat w_N-\hat w_j\|      \Big)^{\frac{N^2-N}{2}}. \\
 \end{split}
\end{equation*}
Let $\ell^{upper}_{LBM}( {\hat w}_{1:N})$ be the upper bound of the approximation loss of acquiring $N$ samples, given $\ell^{\sqrt{upper}}_{LBM}( {\hat w}_{1:N})= \sqrt[\frac{N^2-N}{2}]{\ell^{upper}_{LBM}( {\hat w}_{1:N})}$,  Proposition~3 then holds.
\end{proof}}

Proof of Theorem~1.

 \begin{proof} 
  IWAL denotes a set of observations for its weighted sampling: $\mathcal{F}_t=\{(x_1,y_1,Q_1), (x_2,y_2,Q_2),..., (x_t,y_t,Q_t)\}$. \cao{The key step to prove Theorem~1 is to observe the   martingale difference sequence for any pair $f$ and $g$ in the t-time hypothesis class $\mathcal{H}_t$, that is, $\xi_t=\frac{Q_t}{p_t}\Big(\ell(f(x_t),y_t)-\ell(g(x_t),y_t)\Big)-\Big(R(h)-R(g)\Big)$, where $f, g \in \mathcal{H}_t$.} \cao{By adopting Lemma~3 of \cite{kakade2008generalization}, with $\tau>3$ and $\delta>0$, we firstly know
\begin{equation}
\begin{split}
&{\rm var}[\xi_t|\mathcal{F}_{t-1}]\\
&\leq \textbf{E}_{x_t}\Big[ \frac{Q_t^2}{p_t^2}\Big(\ell(f(x_t),y_t)\!-\!\ell(g(x_t),y_t)\Big)\!-\!\Big(R(h)-R(g)\Big)^2 |\mathcal{F}_{t-1}\Big]\\
& \leq \textbf{E}_{x_t}\Big[ \frac{Q_t^2p_t^2}{p_t^2}  |\mathcal{F}_{t-1}\Big]\\
& = \textbf{E}_{x_t}\Big[ p_t |\mathcal{F}_{t-1}\Big],\\
\end{split}
\label{E_x_t_F}
\end{equation}
and then there exists
\begin{equation}
\begin{split}
&|\sum_{t=1}^T\xi_t| \\
&\leq   \max_{\mathcal{H}_i, i=1,2,...,k} \Bigg\{ 2 \sqrt{\sum_{t=1}^{\tau}\textbf{E}_{x_t}[p_t|\mathcal{F}]_{t-1}}, 6\sqrt{{\rm log}\Big(\frac{8{\rm log}\tau }{\delta}\Big)} \Bigg\} \\ 
&\times \sqrt{{\rm log}\Big(\frac{8{\rm log}\tau}{\delta}\Big)},\\
\end{split}
\label{xi_t}
\end{equation}
where   $\textbf{E}_{x_t}$ denotes the expectation over the operation on $x_t$. By Proposition~2 of  \cite{cesa2008improved}, we know 
\begin{equation}
\begin{split}
&\sum_{t=1}^{\tau}\textbf{E}_{x_t}[p_t|\mathcal{F}_{t-1}]\\
&\leq \Big(\sum_{t=1}^\tau p_t \Big)+36 {\rm log}\Big(\frac{(3+\sum_{t=1}^\tau)\tau^2}{\delta}\Big)\\
&+2\sqrt{\Big(\sum_{t=1}^\tau p_t\Big){\rm log}\Big(\frac{(3+\sum_{t=1}^\tau)\tau^2}{\delta}\Big)}\\
& \leq  \Bigg( \sum_{t=1}^\tau p_t + 6 {\rm log}\Big(\frac{(3+\sum_{t=1}^\tau)\tau^2}{\delta}\Big) \Bigg).   \\
\end{split}
\label{E_x_t}
\end{equation}}
Introducing Eq.~(\ref{E_x_t}) to  Eq.~(\ref{xi_t}), with a probability at least $1-\delta$, we then know

\begin{equation}
\begin{split}
& |\sum_{t=1}^T\xi_t|\\
& \leq  \max_{\mathcal{H}_i, i=1,2,...,k} \Bigg\{\frac{2}{\tau}  \Bigg[\sqrt{\sum_{t=1}^{\tau}p_t}+6\sqrt{{\rm log}\Big[\frac{2(3+\tau)\tau^2}{\delta}\Big] } \Bigg]\\ 
&\times \sqrt{{\rm log}\Big[\frac{16\tau^2|\mathcal{H}_i|^2 {\rm log}\tau}{\delta}\Big]}\Bigg\},
\end{split}
\end{equation}
\cao{For any $\mathcal{B}_i$, the final hypothesis converges into $h_\tau$, which usually holds a tighter disagreement to the optimal $h^*$. It is thus $R(h_\tau)-R(h^*) \leq |\sum_{t=1}^\tau\xi_t|$.
 then the  error disagreement bound of Theorem~1 hold.} We next prove the label complexity bound of MHEAL. Following Theorem~1, there exists
\begin{equation}
\begin{split}
& \textbf{E}_{x_t}[p_t|\mathcal{F}_{t-1}]\\
& \leq 4\theta_{\rm IWAL}K_\ell  \times \Bigg(R^*+\sqrt{(\frac{2}{t-1}){\rm log}(2t(t-1)|)  \frac{|\mathcal{H}|^2  }{\delta})}\Bigg), \\
\end{split}
\end{equation}
where  $R^*$ denotes $R(h^*)$. Let  
$\sqrt{(\frac{2}{t-1}){\rm log}(2t(t-1)|)  \frac{|\mathcal{H}|^2  }{\delta})} \propto O\Big({\rm log}(\frac{\tau|\mathcal{H}|}{\delta})\Big)$, with the proof of Lemma~4 of \cite{cortes2020region}, Eq.~(17) then can be approximated  as
\begin{equation}
\begin{split}
& \textbf{E}_{x_t}[p_t|\mathcal{F}_{t-1}] \\
& \leq 4\theta_{\rm IWAL}  K_\ell\Bigg(\tau R^*+O\Big(R(h^*)\tau {\rm log}(\frac{t|\mathcal{H}|}{\delta})   \Big )  + O\Big({\rm log}^3(\frac{t|\mathcal{H}|}{\delta})\Big) \Bigg), \\
\end{split}
\end{equation}
for any cluster $\mathcal{B}_i$, by adopting the proof of Theorem~2 of \cite{cortes2020region}, we know

\begin{equation}
\begin{split}
& \textbf{E}_{x_t}[p_t|\mathcal{F}_{t-1}]\\  
&\leq   8 K_\ell\Bigg\{\Big[\sum_{j=1}^{N_i} \theta_{\rm MHEAL} R_j^*\tau p_j\Big] 
 +\sum_{j=1}^{N_i} O\Bigg(\sqrt{R_j^*\tau p_j{\rm log}\Big[\frac{\tau|\mathcal{H}_i|N_i}{\delta} \Big]}\Bigg)\!\\
  &+\!O\Bigg(N_i {\rm log}^3\Big(\frac{\tau|\mathcal{H}_i|N_i}{\delta}\Big)\Bigg)\Bigg\}. \\
\end{split}
\end{equation}
Since $\mathcal{Q}=k\times \max_{\mathcal{H}_i}\textbf{E}_{x_t}[p_t|\mathcal{F}_{t-1}], {\rm s.t.}, i=1,2,...,k$, our analysis of Theorem~1 on $\mathcal{Q}$ holds.
 \end{proof}

 \subsection*{Experiments}

\begin{figure*} 
\subfloat[Unsupervised learning vs. MHEAL]{
\begin{minipage}[t]{0.33\textwidth}
\centering
\includegraphics[width=1.8in,height=1.52in]{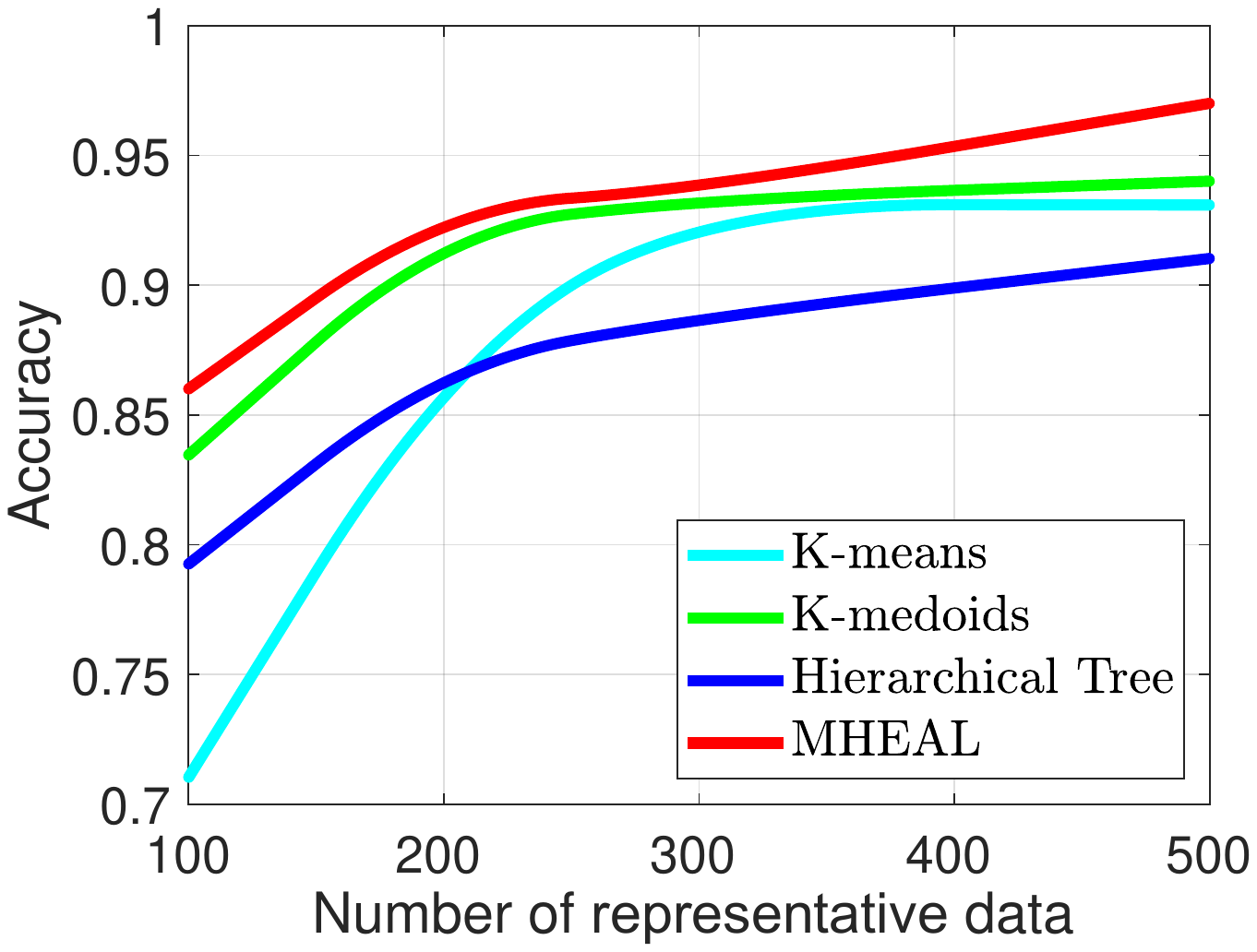}
\end{minipage}
}
\subfloat[Supervised learning vs. MHEAL]{
\begin{minipage}[t]{0.33\textwidth}
\centering
\includegraphics[width=1.8in,height=1.52in]{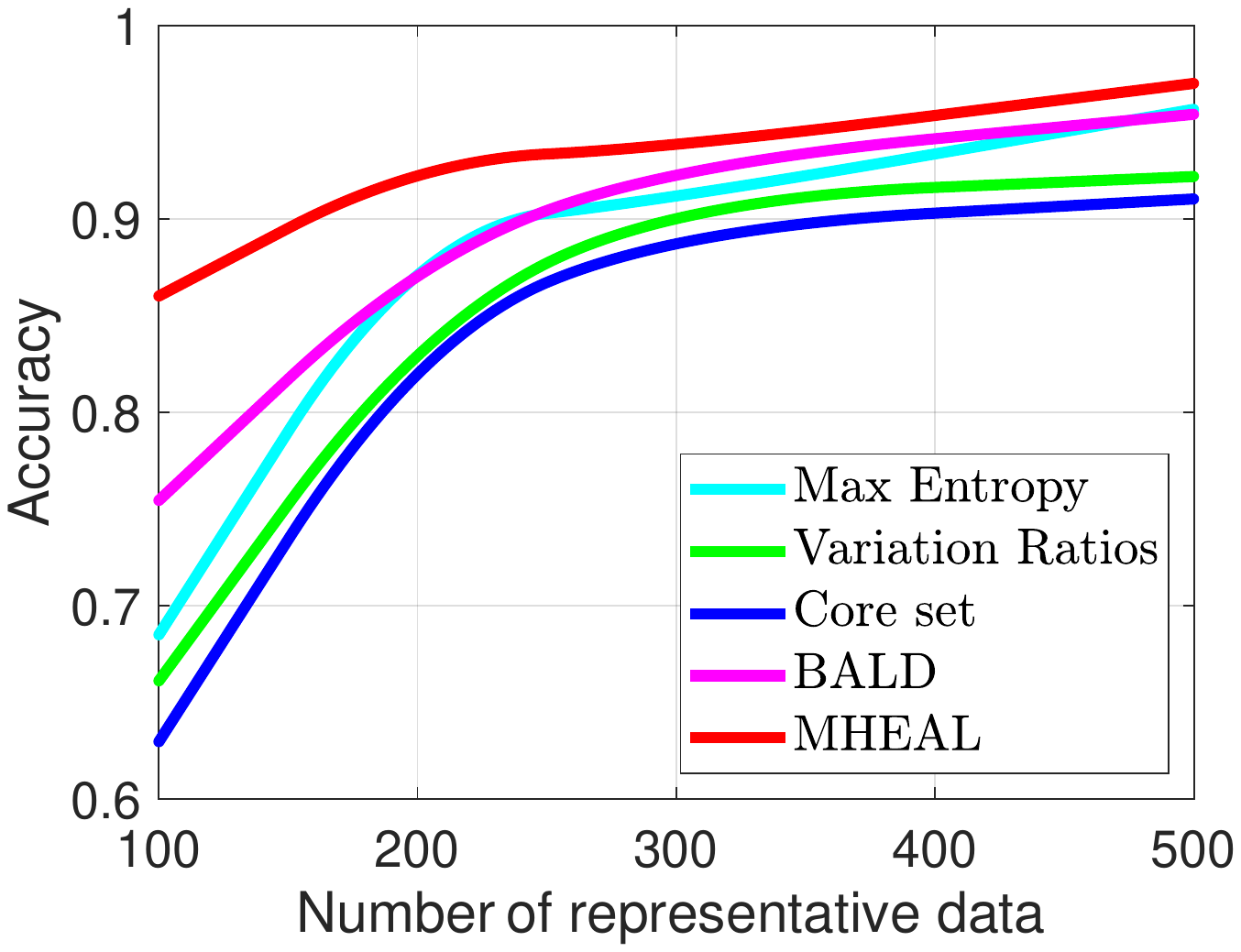}
\end{minipage}
} 
\subfloat[Deep learning vs. MHEAL]{
\begin{minipage}[t]{0.33\textwidth}
\centering
\includegraphics[width=1.8in,height=1.52in]{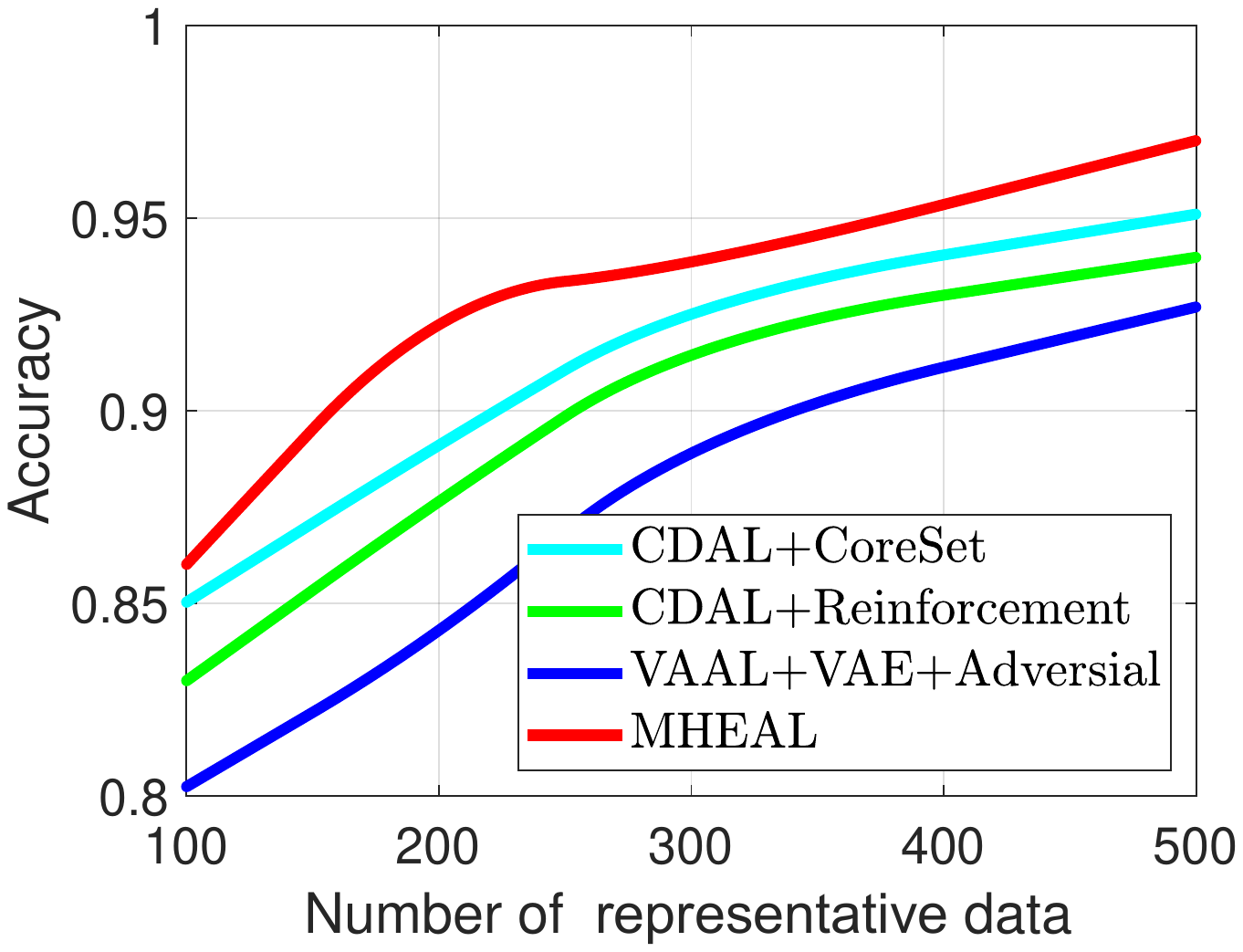}
\end{minipage}
} 
\caption{Data-efficient learning  using  representative data from  repeated  MNIST. (a)Unsupervised learning vs. MHEAL. (b) Supervised learning vs. MHEAL. (c)Deep learning vs. MHEAL.}  
\end{figure*}

\begin{figure*} 
\subfloat[Unsupervised learning vs. MHEAL]{
\begin{minipage}[t]{0.33\textwidth}
\centering
\includegraphics[width=1.8in,height=1.52in]{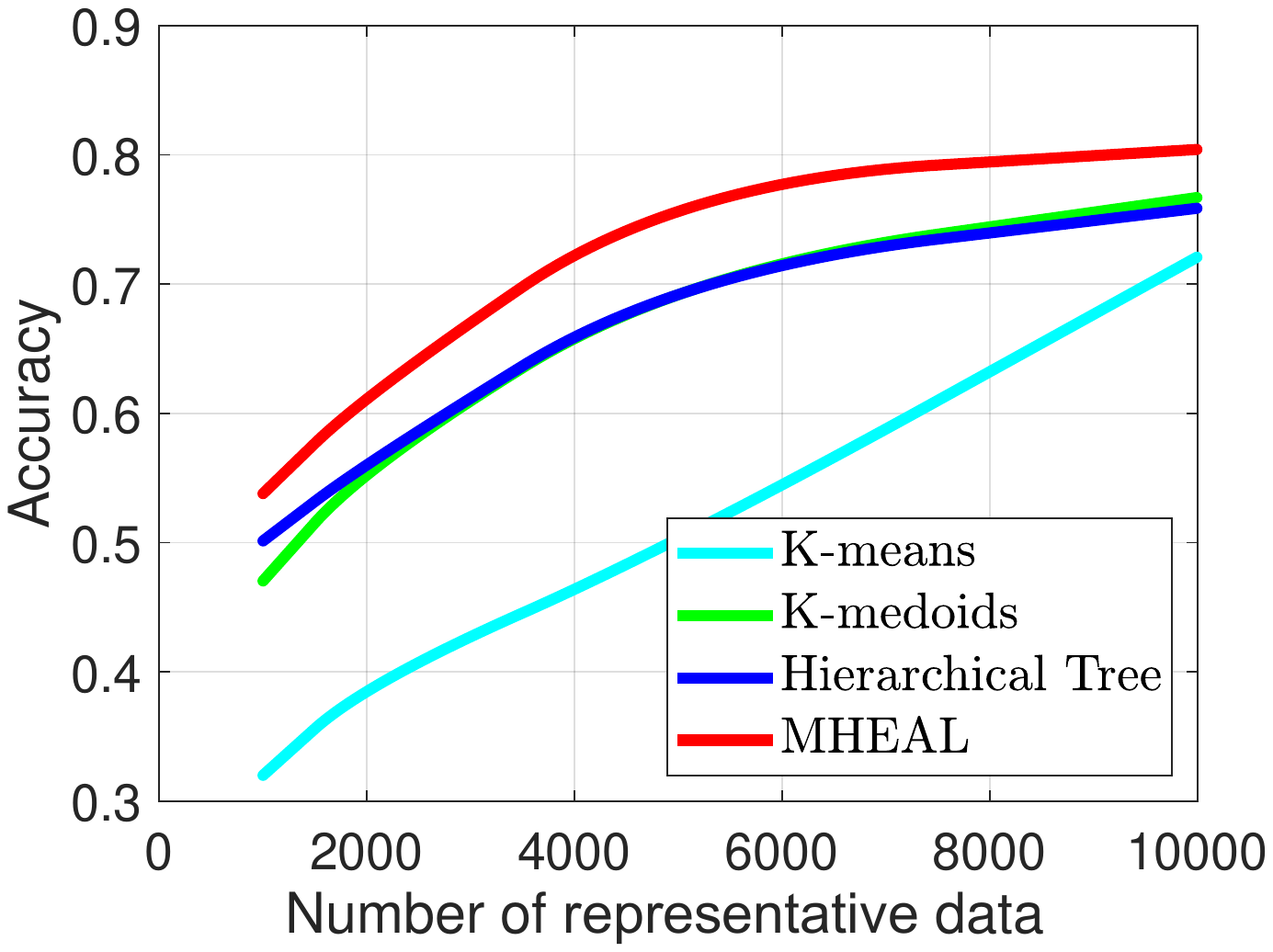}
\end{minipage}
}
\subfloat[Supervised learning vs. MHEAL]{
\begin{minipage}[t]{0.33\textwidth}
\centering
\includegraphics[width=1.8in,height=1.52in]{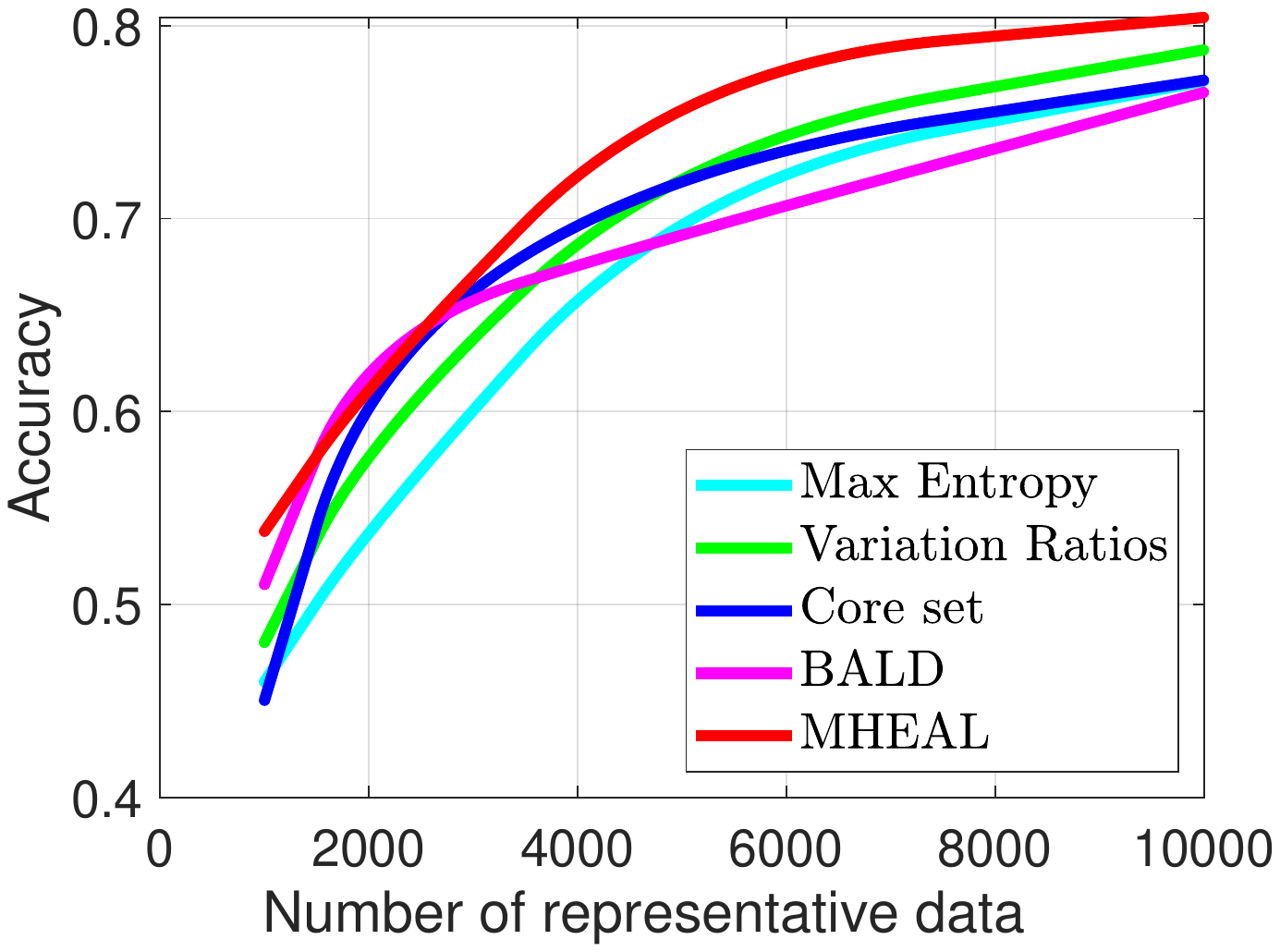}
\end{minipage}
} 
\subfloat[Deep learning vs. MHEAL]{
\begin{minipage}[t]{0.33\textwidth}
\centering
\includegraphics[width=1.8in,height=1.52in]{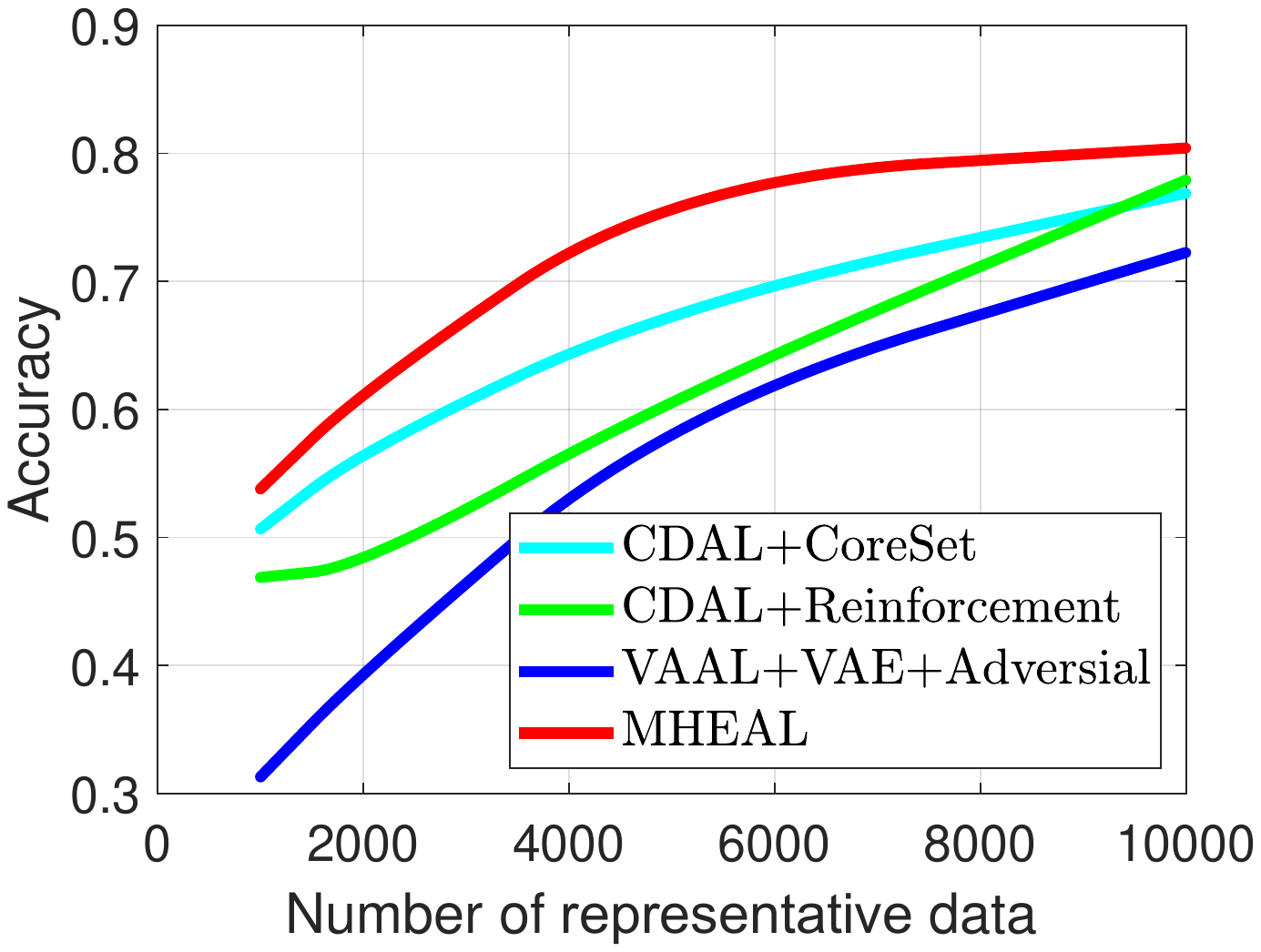}
\end{minipage}
} 
\caption{Data-efficient learning  using  representative data from  repeated  CIFAR-10. (a)Unsupervised learning vs. MHEAL. (b)Supervised learning vs. MHEAL. (c)Deep learning vs. MHEAL.
 }  
\end{figure*}

\begin{figure*} 
\subfloat[Unsupervised learning vs. MHEAL]{
\begin{minipage}[t]{0.33\textwidth}
\centering
\includegraphics[width=1.8in,height=1.52in]{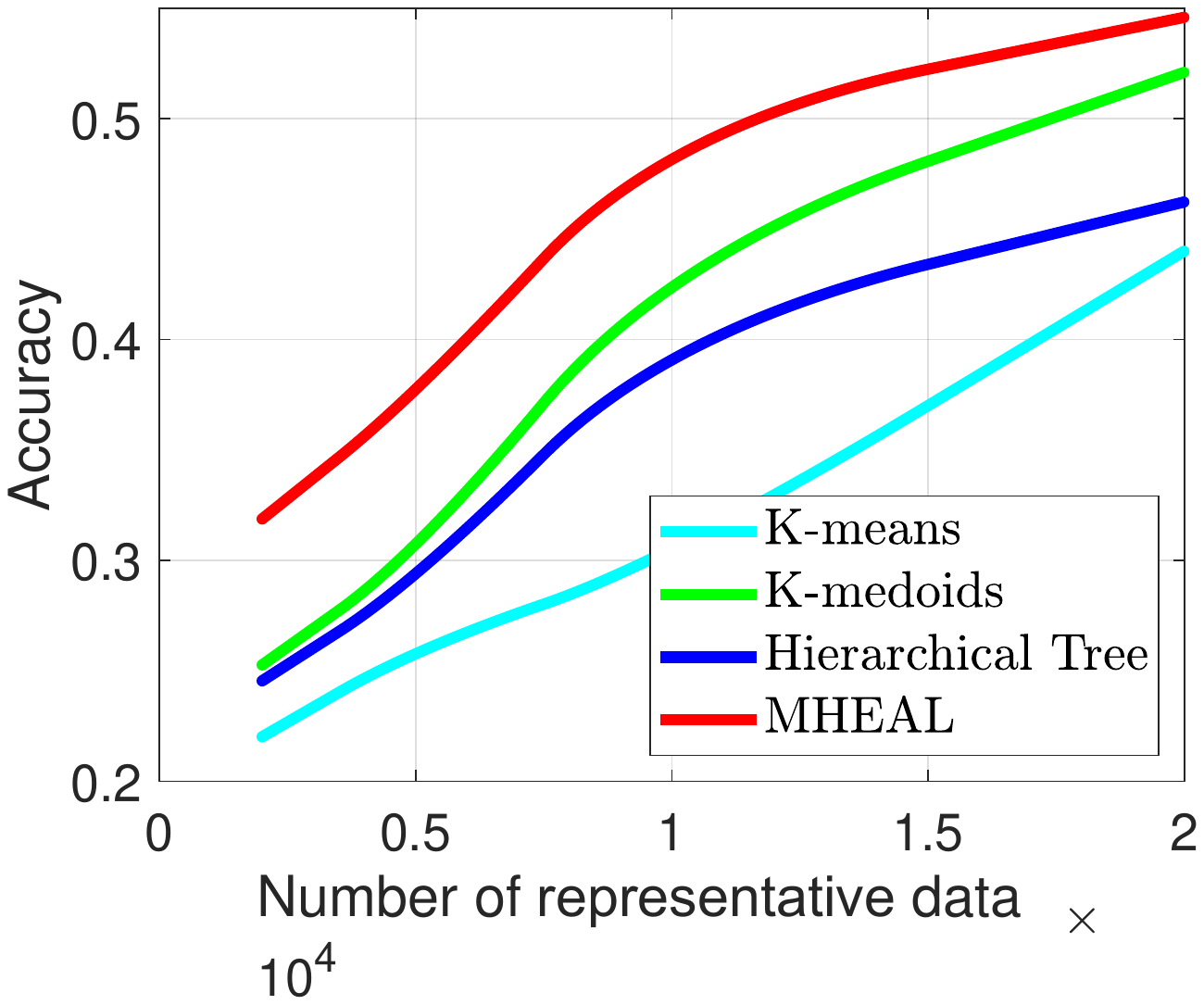}
\end{minipage}
}
\subfloat[Supervised learning vs. MHEAL]{
\begin{minipage}[t]{0.33\textwidth}
\centering
\includegraphics[width=1.8in,height=1.52in]{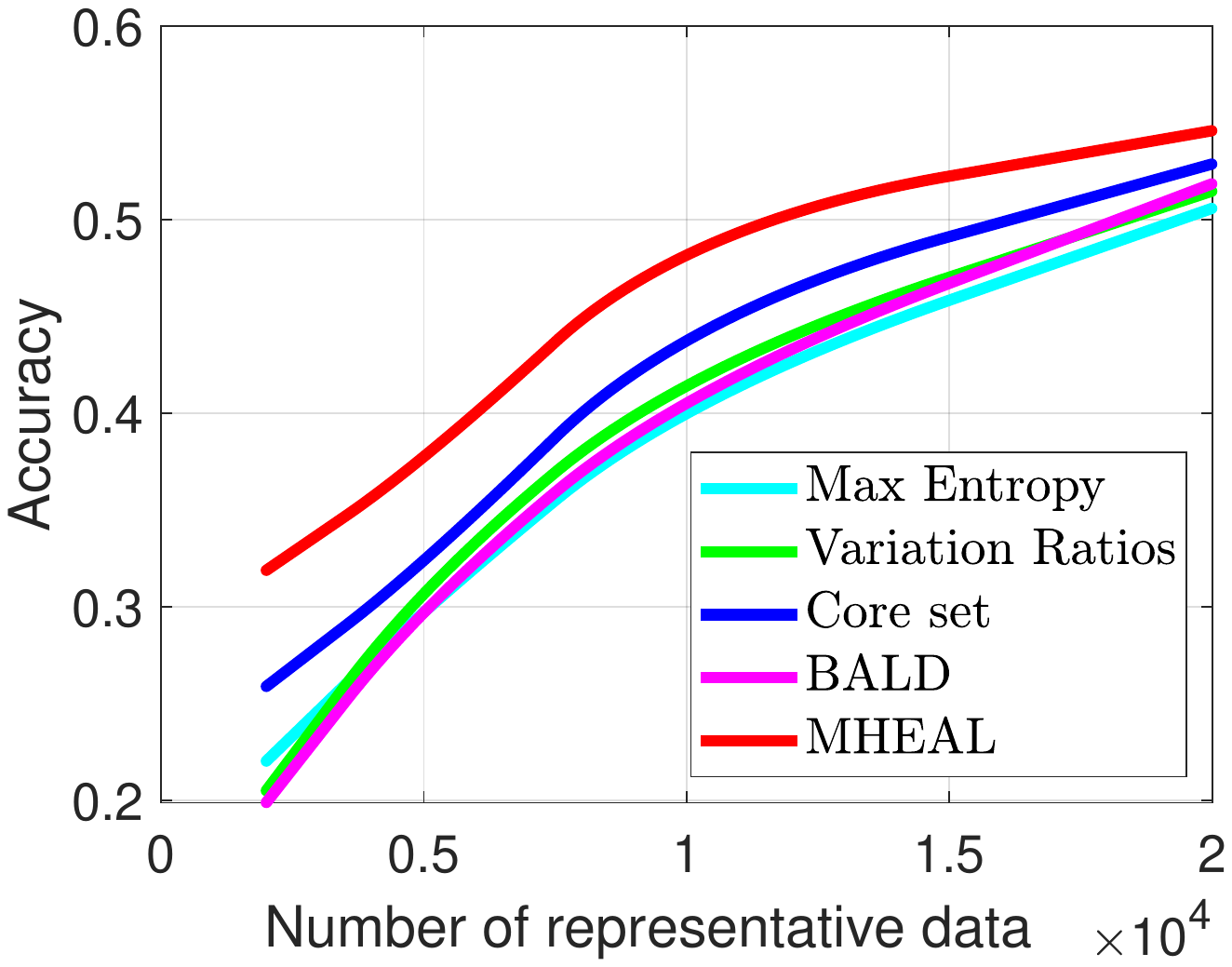}
\end{minipage}
} 
\subfloat[Deep learning vs. MHEAL]{
\begin{minipage}[t]{0.33\textwidth}
\centering
\includegraphics[width=1.8in,height=1.52in]{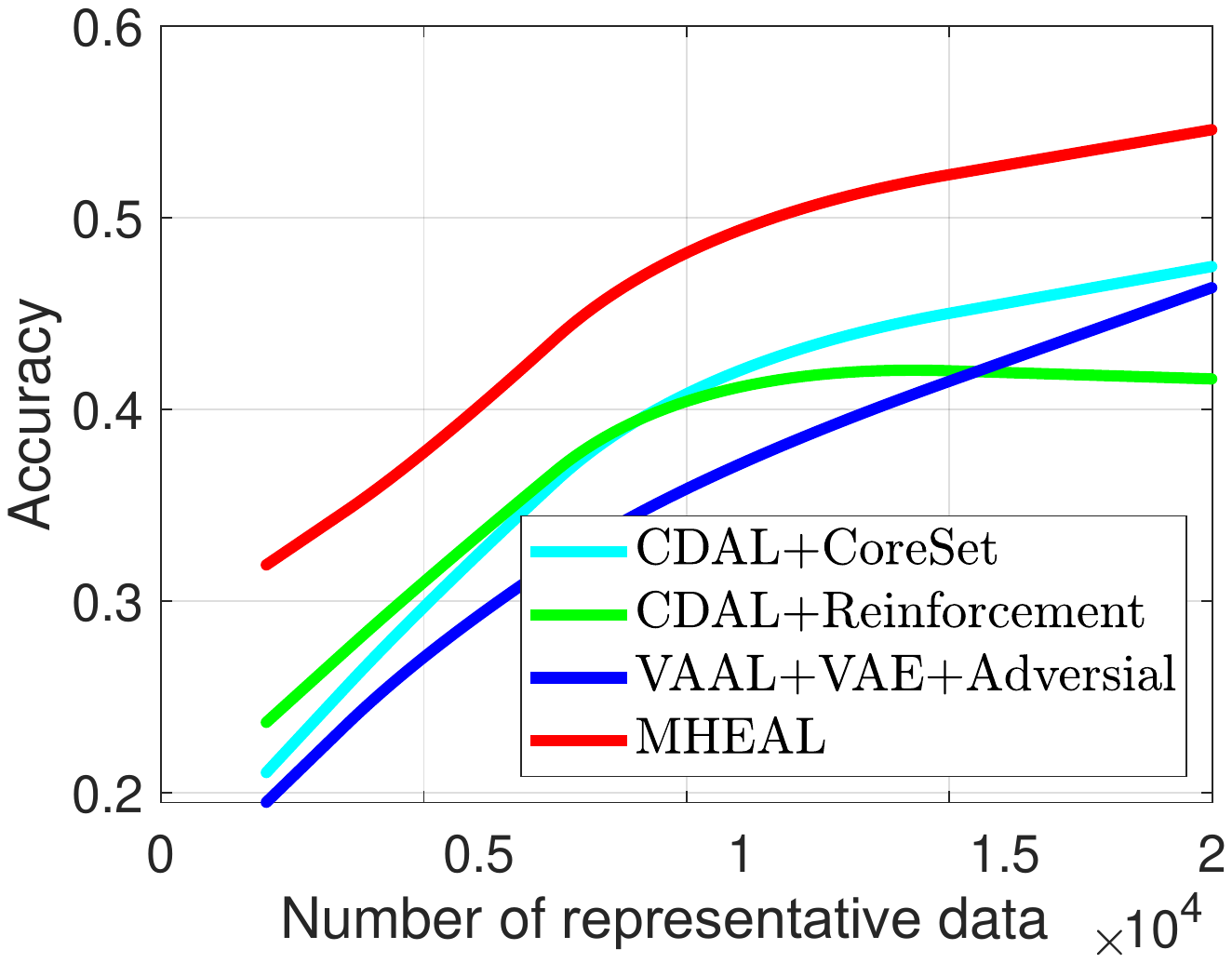}
\end{minipage}
} 
\caption{Data-efficient learning    from  repeated  CIFAR-100. (a) Unsupervised learning vs. MHEAL. (b) Supervised learning vs. MHEAL. (c) Deep learning vs. MHEAL.
 }  
\end{figure*}

 \begin{table*} 
 \caption{Accuracy statistics of classification on noisy  representative data via different deep AL baselines. }
  \renewcommand{\arraystretch}{1.2}
 \setlength{\tabcolsep}{1.5pt}{
\begin{center}
\scalebox{1.29}{
 \begin{tabular}{c|c  c c c  |c  c cc |c  c ccc  } 
\hline
\multirow{2}{*}{Algorithms}  &    \multicolumn{4}{c|}{MNIST (CNN)}  &         \multicolumn{4}{c|}{CIFAR-10 (ResNet20)}&   \multicolumn{4}{c }{CIFAR-100  (ResNet20)} \\
    &     100  &     200 &     300  &  500     &   1K    &   2K  &5K &10K &   2K    &   5K  &10K  &20K \\
\hline
K-means \cite{cao2020shattering}&   0.7021  &  0.8542     & 0.9210 &  0.9254 &0.3102&0.3902   &0.4878  &    0.7156     &   0.2102&    0.2430    &    0.2814    &    0.4325    \\
K-medoids \cite{cao2020shattering}&  0.8265  & 0.9102      &  0.9263 &  0.9361 & 0.4654   & 0.5523&0.7012 &  0.7562   &  0.2315  &    0.2816&  0.4321& 0.5019  \\
Hierarchical  Tree \cite{dasgupta2008hierarchical} &0.7842 & 0.8531   &0.8763 &  0.9026    &       0.4958 &0.5543  &0.7012     & 0.7456 &      0.2251&  0.2743 &    0.3852  &0.4536  \\
Max Entropy \cite{gal2017deep}  &  0.6745   & 0.8756       & 0.9026  & 0.9401         &  0.4412 &0.5320& 0.7019  & 	0.7589    & 0.2015  & 0.2656& 0.3987 &0.4875\\
Variation Ratios  \cite{gal2017deep}& 0.6510  &0.8302      &   0.9064  & 0.9108 & 0.4657  & 0.5658 &   0.7198   &   0.7687   & 0.1846   &0.2965&0.4016&  0.5021   \\
 Core-set \cite{sener2018active}& 0.6184   &   0.8256     & 0.8874  & 0.9010        &  0.4402  &0.6218  &  0.7215     &  0.7650   &0.2321  & 0.3025&  0.4326&  0.5016 \\
BALD \cite{houlsby2011bayesian}&  0.7412  &0.8654       & 0.9152  & 0.9401       &  0.5003    & \textbf{0.6287}& 0.6781&0.7463    & 0.1852   &0.2965 &   0.3978 &0.5016  \\
CDAL+CoreSet \cite{agarwal2020contextual}&    0.8410 & 0.8745   & 0.9103      & 0.9421     & 0.4985& 0.5523&   0.6714        &   0.7563&0.1998 &    0.2785&0.4135&  0.4685   \\
CDAL+Reinforcement  \cite{agarwal2020contextual}&  0.8103& 0.8654   &0.9108 &  0.9256  & 0.4530    & 0.4612     & 0.5814& 0.7563&         0.2216   &  0.2965 &    0.4018&0.4158  \\
VAAL+VAE+Adversial \cite{sinha2019variational}&0.7945 &0.8325  & 0.8847      &  0.9106        &     0.3062    & 0.3785&  0.5874&0.7113 &   0.1746&   0.2625   & 0.3457&   0.4523\\
MHEAL&   \textbf{0.8548}   &   \textbf{0.9210}     & \textbf{0.9263}  & \textbf{0.9654}    & \textbf{0.5214} &  {0.6050}  &  \textbf{0.7712}    &\textbf{0.7885}   & \textbf{0.3056} &   \textbf{0.3578}   & \textbf{0.4716} & \textbf{0.5246} \\
\hline
\end{tabular}}
\end{center}}
\end{table*}

\begin{figure*} 
\subfloat[Unsupervised learning vs. MHEAL]{
\begin{minipage}[t]{0.33\textwidth}
\centering
\includegraphics[width=1.8in,height=1.52in]{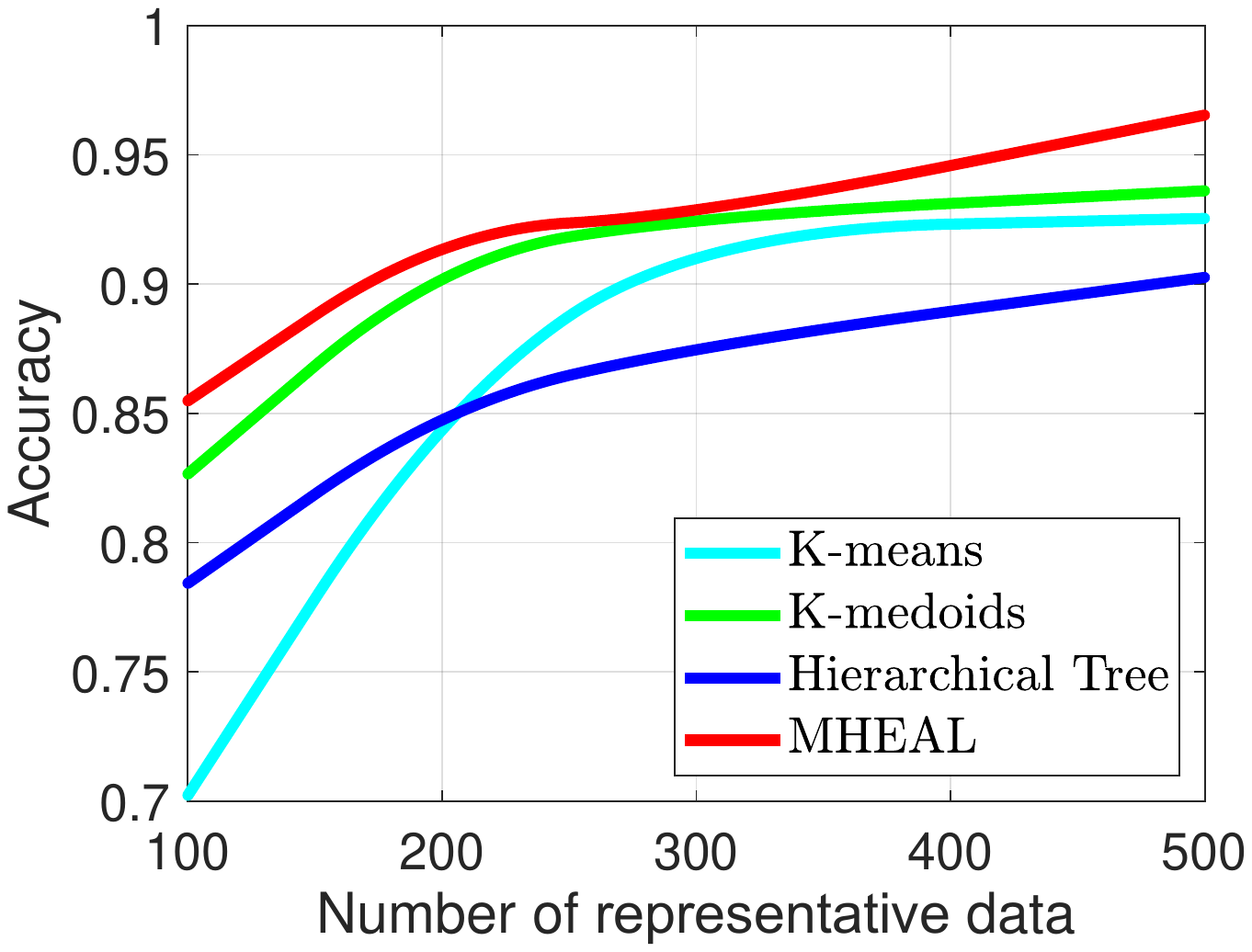}
\end{minipage}
}
\subfloat[Supervised learning vs. MHEAL]{
\begin{minipage}[t]{0.33\textwidth}
\centering
\includegraphics[width=1.8in,height=1.52in]{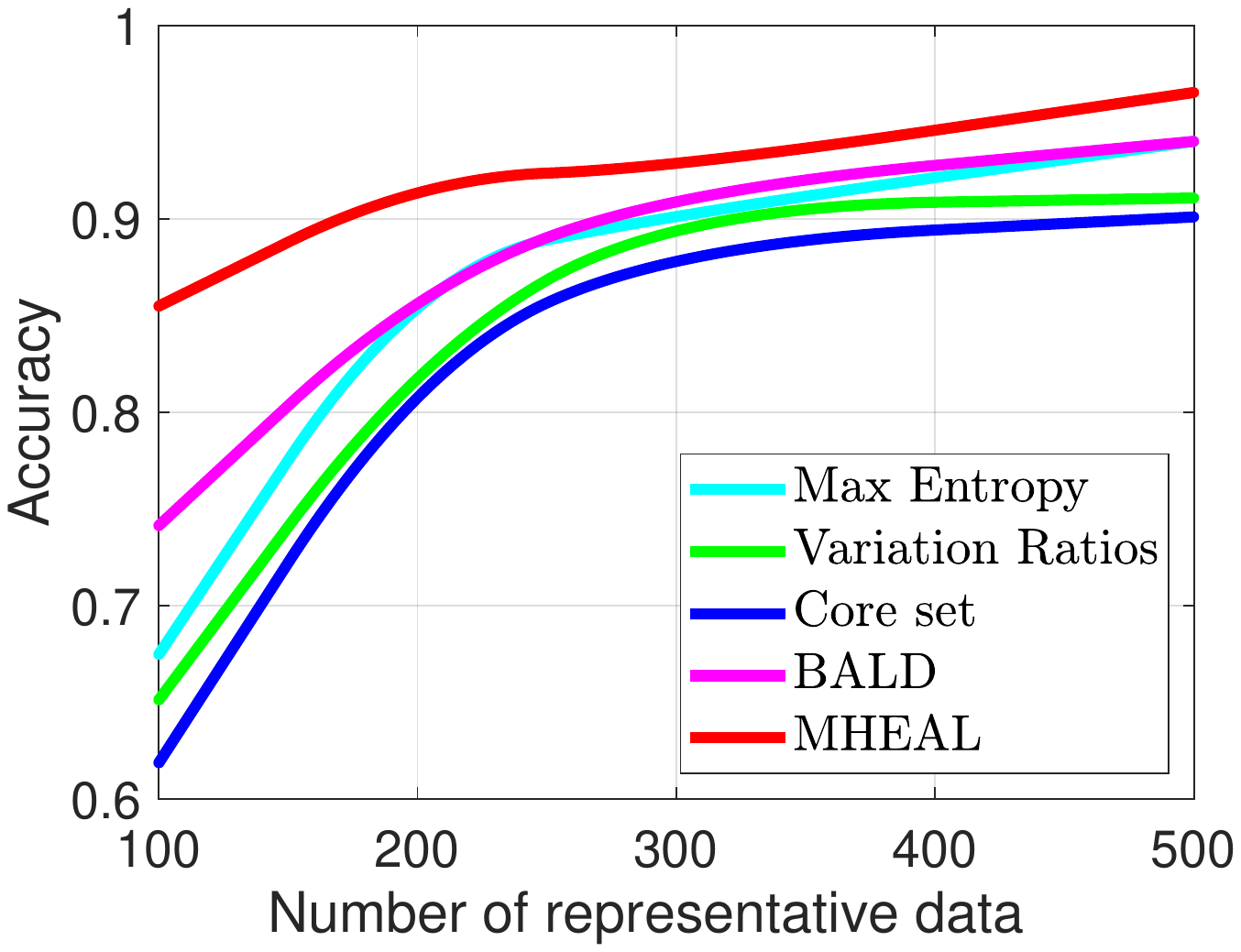}
\end{minipage}
} 
\subfloat[Deep learning vs. MHEAL]{
\begin{minipage}[t]{0.33\textwidth}
\centering
\includegraphics[width=1.8in,height=1.52in]{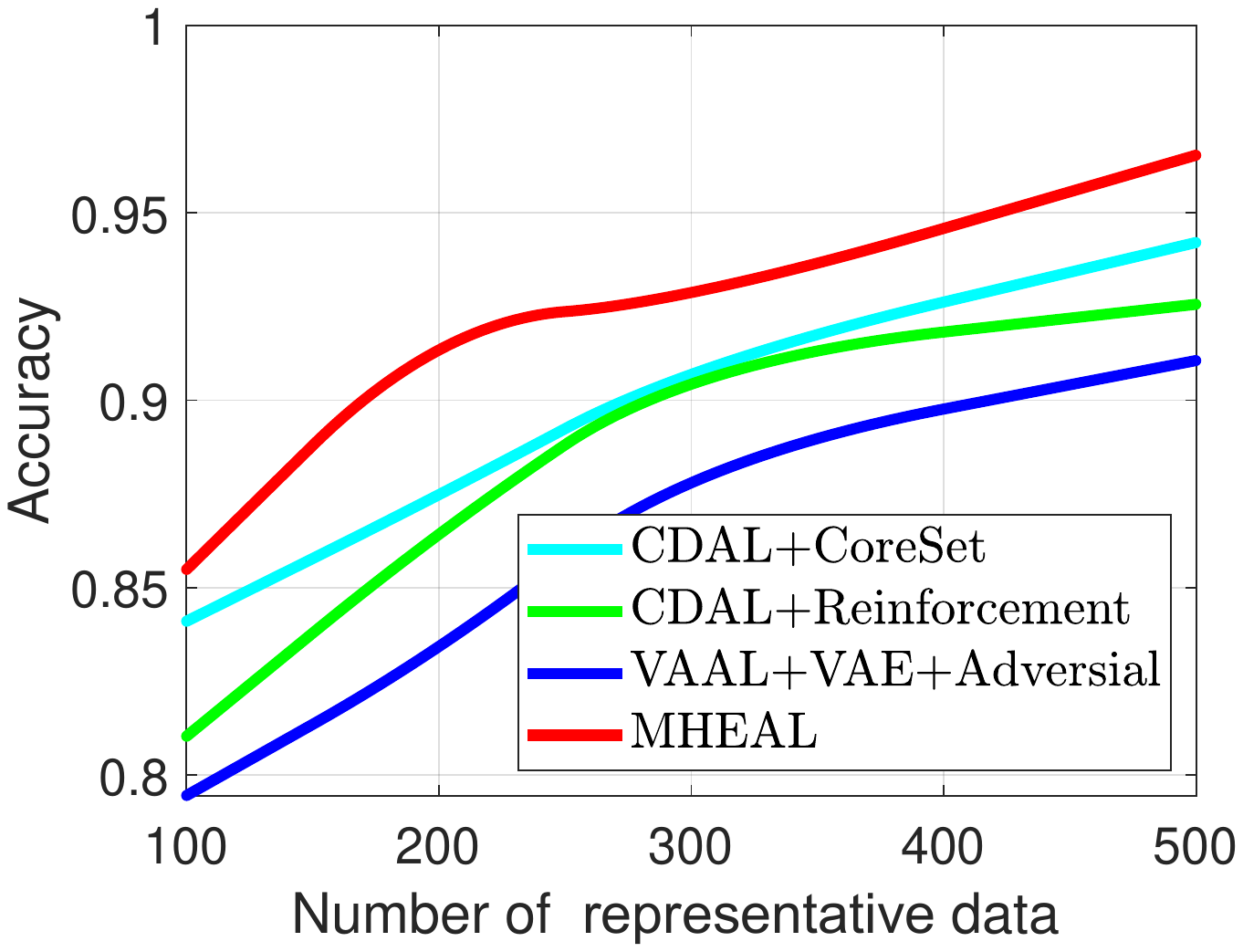}
\end{minipage}
} 
\caption{Data-efficient learning  using  representative data from  noisy  MNIST. (a)Unsupervised learning vs. MHEAL. (b)Supervised learning vs. MHEAL. (c)Deep learning vs. MHEAL.
 }  
\end{figure*}

\begin{figure*} 
\subfloat[Unsupervised learning vs. MHEAL]{
\begin{minipage}[t]{0.33\textwidth}
\centering
\includegraphics[width=1.8in,height=1.52in]{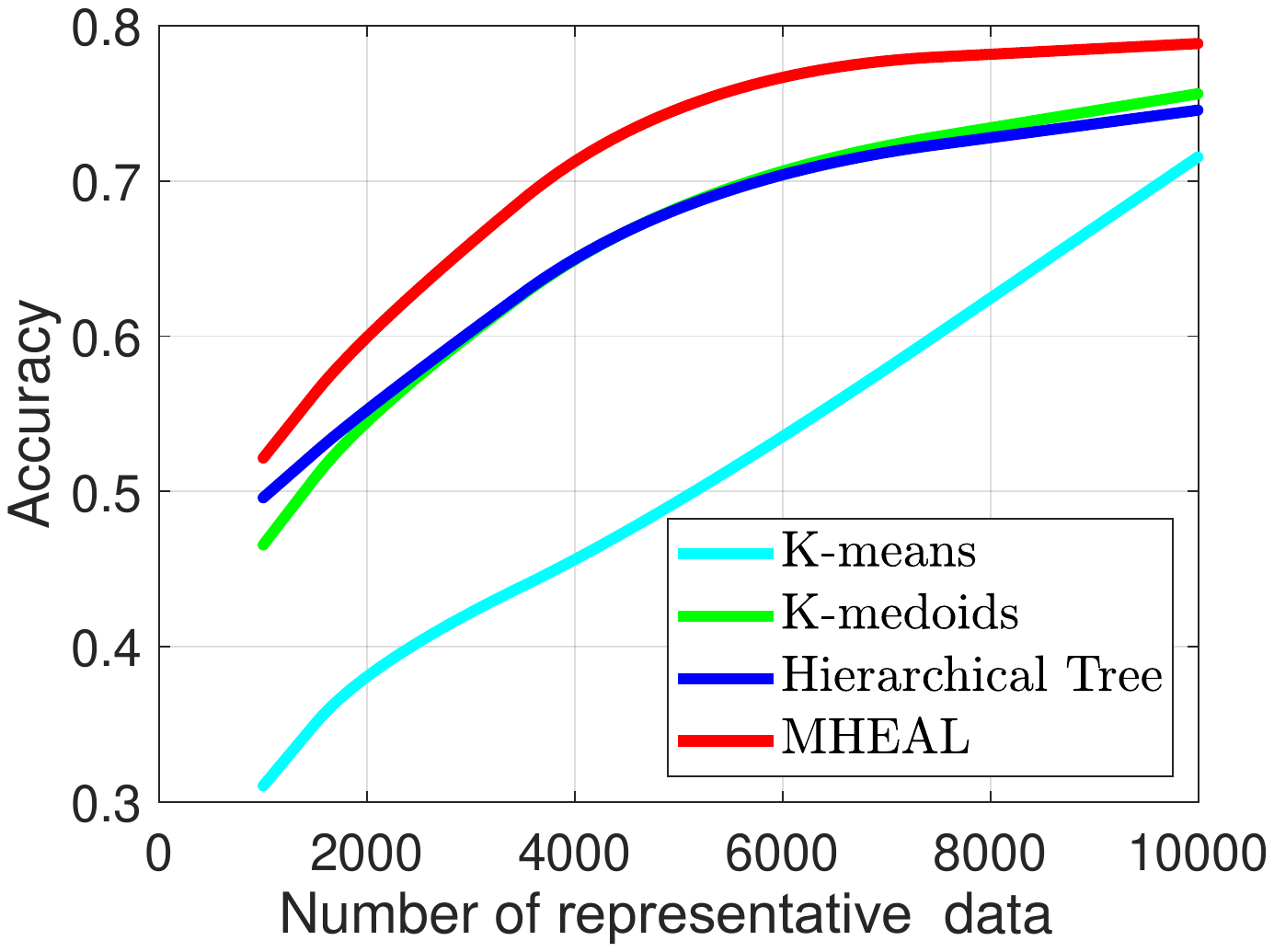}
\end{minipage}
}
\subfloat[Supervised learning vs. MHEAL]{
\begin{minipage}[t]{0.33\textwidth}
\centering
\includegraphics[width=1.8in,height=1.52in]{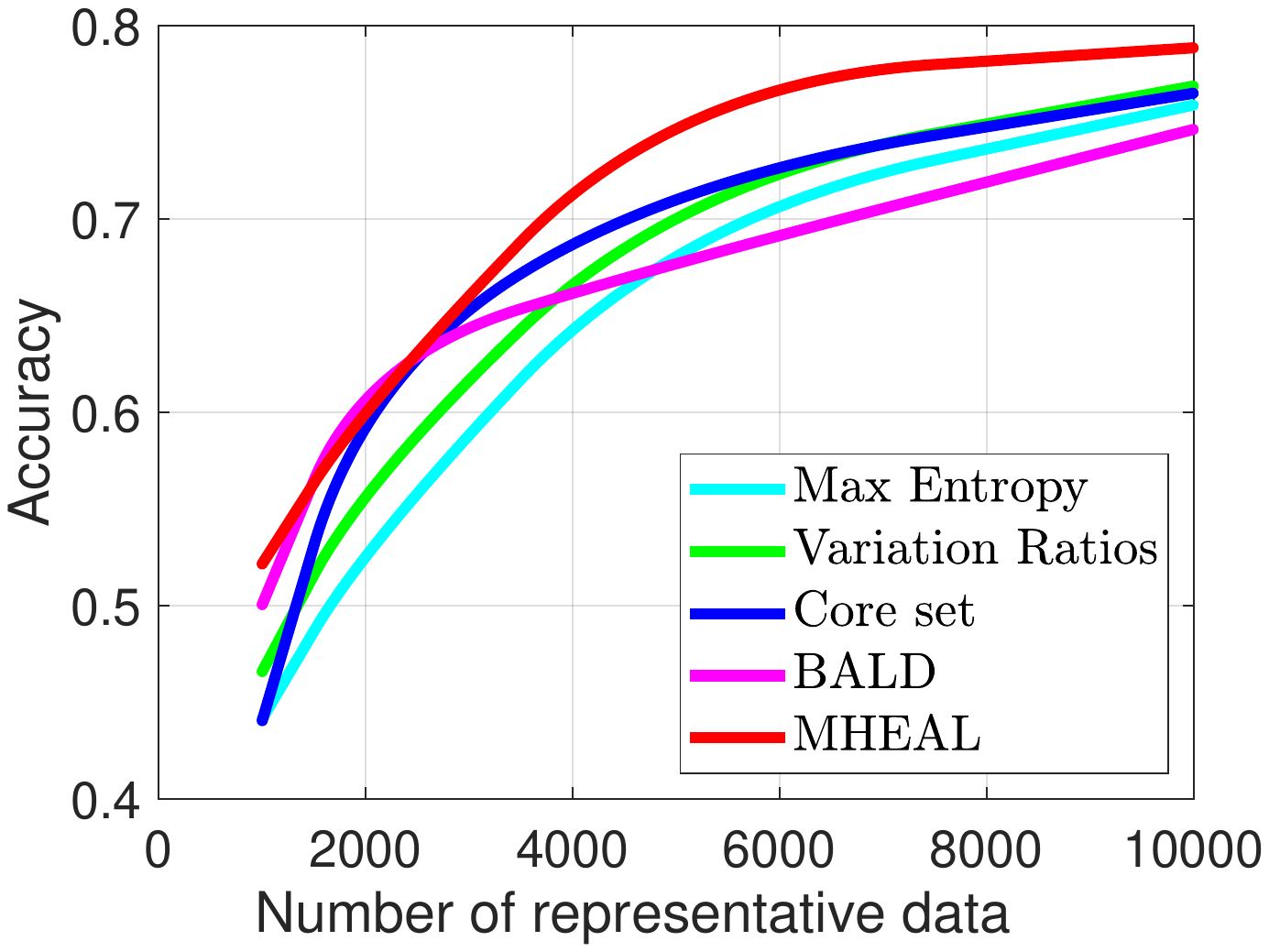}
\end{minipage}
} 
\subfloat[Deep learning vs. MHEAL]{
\begin{minipage}[t]{0.33\textwidth}
\centering
\includegraphics[width=1.8in,height=1.52in]{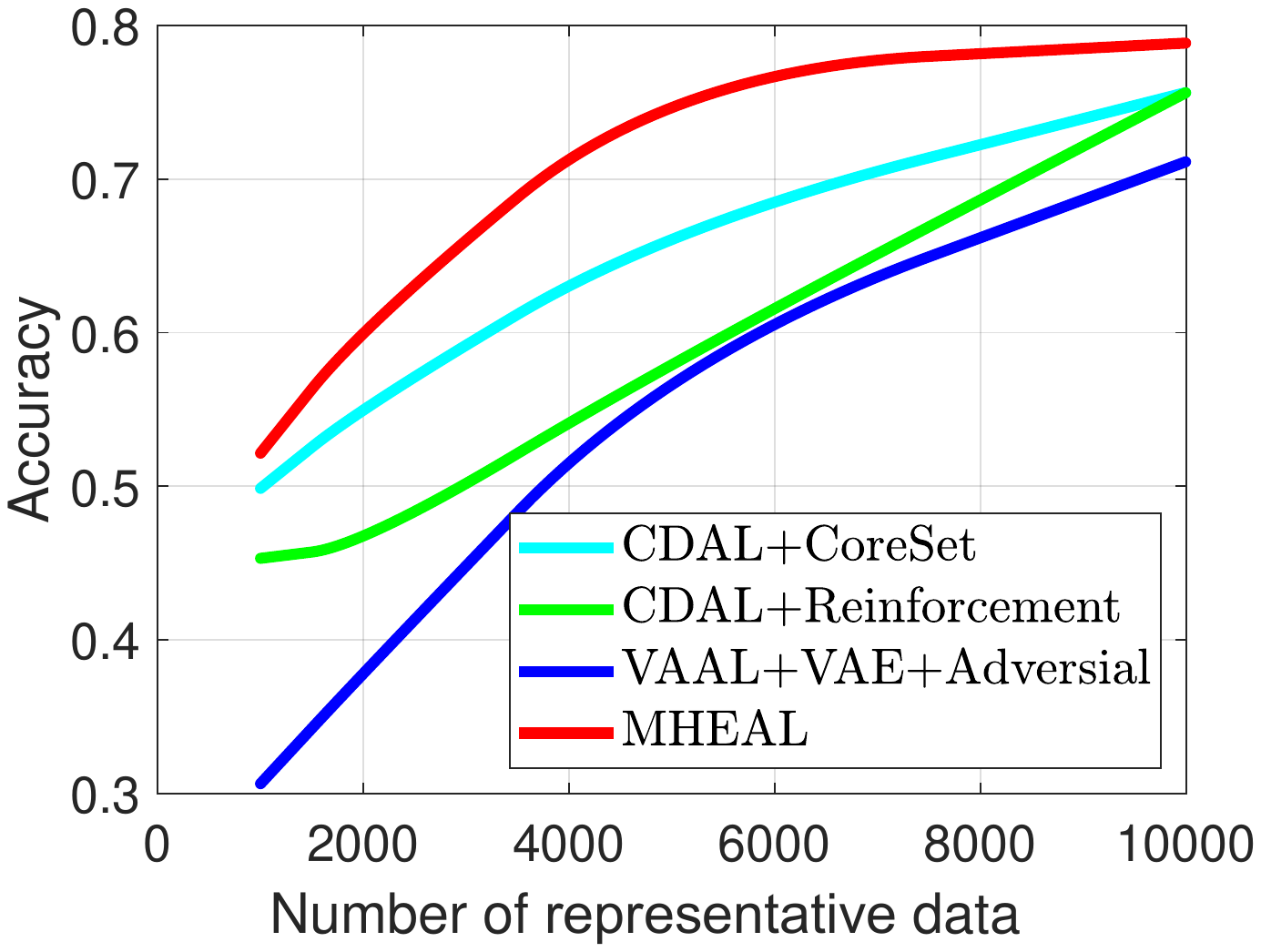}
\end{minipage}
} 
\caption{Data-efficient learning  using  representative data from  noisy  CIFAR-10. (a)Unsupervised learning vs. MHEAL. (b)Supervised learning vs. MHEAL. (c)Deep learning vs. MHEAL.
 }  
\end{figure*}

\begin{figure*} 
\subfloat[Unsupervised learning vs. MHEAL]{
\begin{minipage}[t]{0.33\textwidth}
\centering
\includegraphics[width=1.8in,height=1.52in]{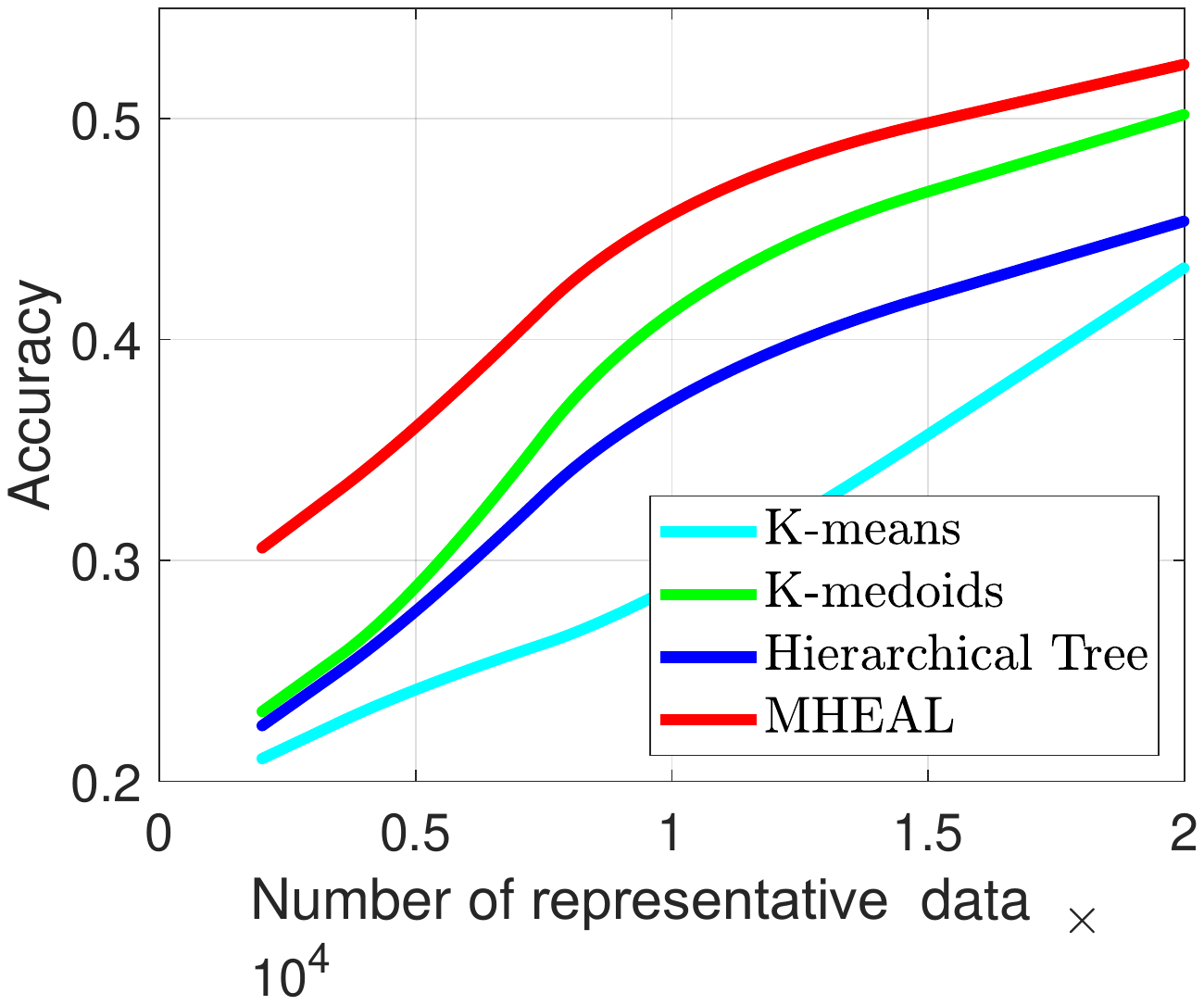}
\end{minipage}
}
\subfloat[Supervised learning vs. MHEAL]{
\begin{minipage}[t]{0.33\textwidth}
\centering
\includegraphics[width=1.8in,height=1.52in]{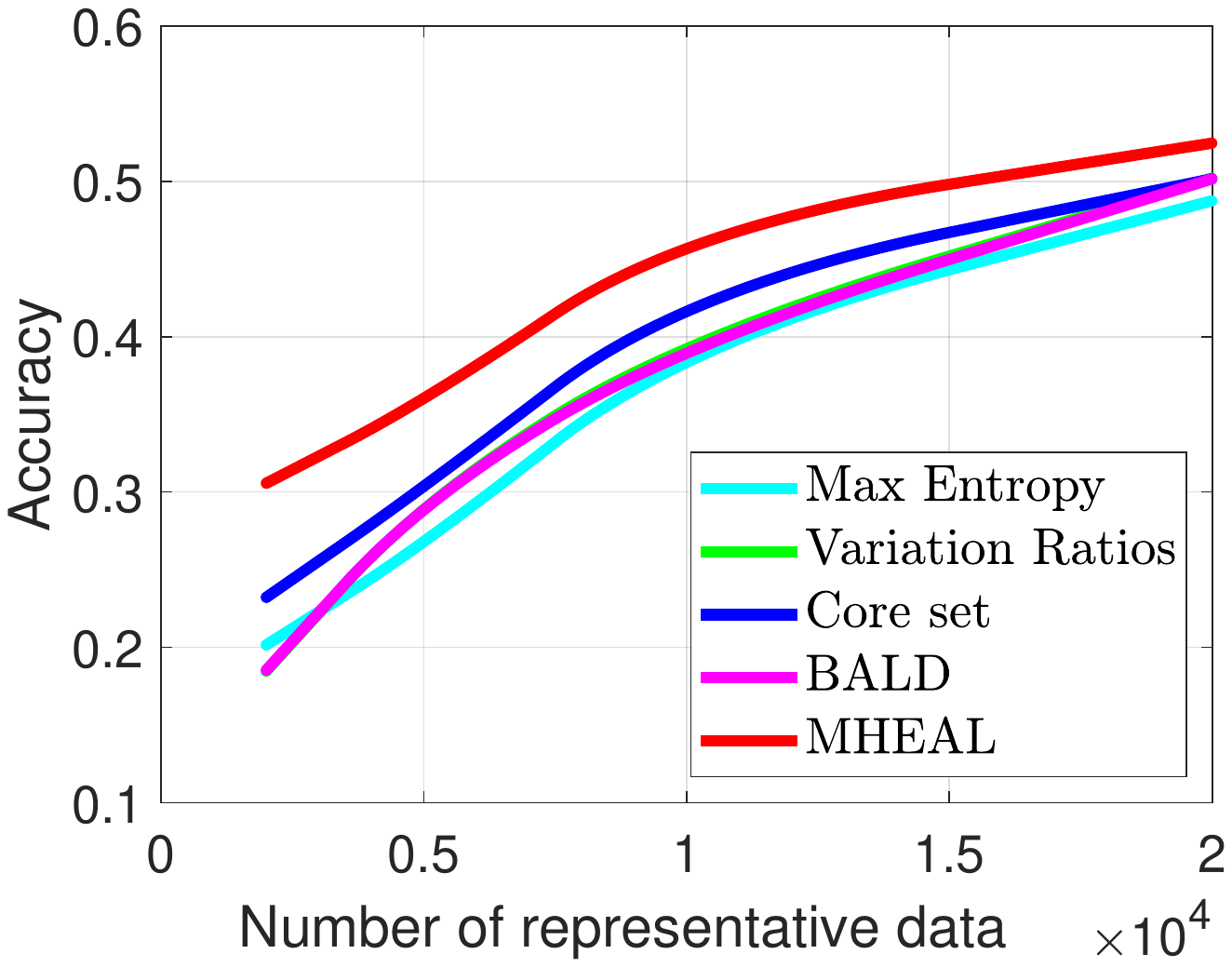}
\end{minipage}
} 
\subfloat[Deep learning vs. MHEAL]{
\begin{minipage}[t]{0.33\textwidth}
\centering
\includegraphics[width=1.8in,height=1.52in]{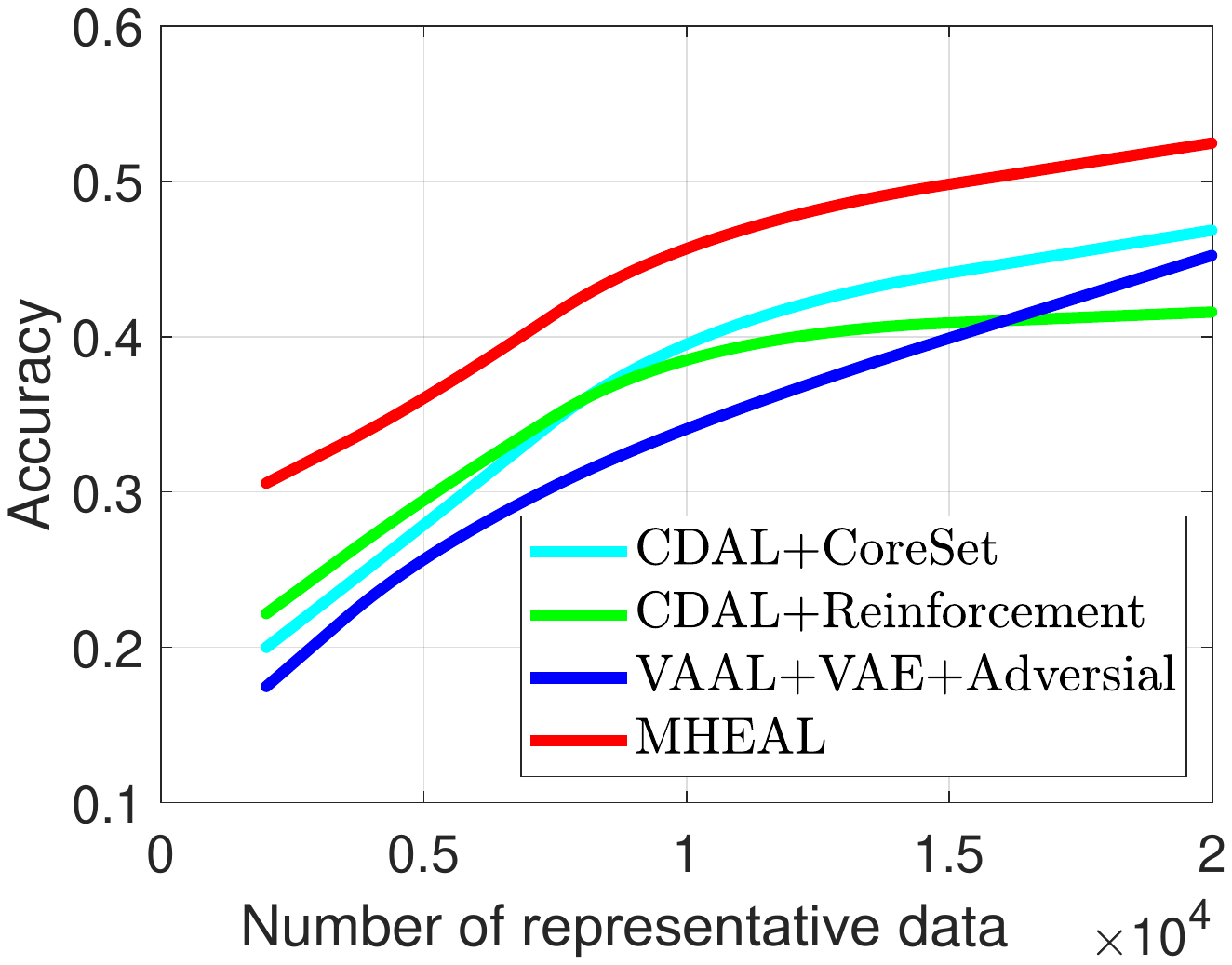}
\end{minipage}
} 
\caption{Data-efficient learning  using  representative data from  noisy CIFAR-100. (a)Unsupervised learning vs. MHEAL. (b)Supervised learning vs. MHEAL. (c)Deep learning vs. MHEAL.
 }  
\end{figure*}
\end{document}